\documentclass[a4paper,10pt]{article}
\usepackage{ifthen}
\newboolean{review}
\setboolean{review}{false}

\usepackage[breaklinks,colorlinks]{hyperref}
\usepackage{doi} 
\newcommand{\arXiv}[1]{\href{http://arxiv.org/abs/#1}{arXiv:#1}} 
\usepackage[round,colon,authoryear,sort&compress]{natbib}

\usepackage{hypernat} 

\usepackage{color}
\definecolor{blue}{rgb}{0.0,0.0,0.6}
\hypersetup{colorlinks=true,citecolor=blue,urlcolor=blue,linkcolor=blue,menucolor=blue}

\usepackage[capitalize]{cleveref}
\usepackage{afterpage} 

\usepackage{graphicx} 
\graphicspath{{./}}
\usepackage{subfigure} 
\usepackage[bf]{caption} 
\usepackage{multirow}
\usepackage{rotating}

\usepackage[a4paper,portrait,margin=2cm]{geometry}
\usepackage{times}

\usepackage[nodayofweek]{datetime} 

\usepackage{lineno}

\newcommand{\dson}{\renewcommand{\baselinestretch}{1.5}\large\normalsize}

\usepackage{enumitem} 
\setlist[itemize]{topsep=0pt,parsep=0pt,itemsep=0pt,leftmargin=1.8em}
\setlist[enumerate]{topsep=0pt,parsep=0pt,itemsep=0pt,leftmargin=2.4em}

\usepackage{xspace}
\newcommand{\etc}{{\it etc.}\xspace}
\newcommand{\ie}{{\it i.e.},\xspace}
\newcommand{\eg}{{\it e.g.},\xspace}

\newcommand{\hide}[1]{}

\usepackage{amssymb}
\usepackage{pifont}
\newcommand{\cmark}{\xspace\mbox{\color[rgb]{0,0.5,0}\ding{51}}\xspace\normalcolor}%
\newcommand{\xmark}{\xspace\mbox{\color[rgb]{0.5,0,0}\ding{55}}\xspace\normalcolor}%
\newcommand{\cmarkw}{\xspace\mbox{\color[rgb]{1,1,1}\ding{51}}\xspace\normalcolor}%
\newcommand{\xmarkw}{\xspace\mbox{\color[rgb]{1,1,1}\ding{55}}\xspace\normalcolor}%
\newlength{\eqcolwidthA}
\newlength{\eqcolwidthB}
\newlength{\eqcolwidthC}
\usepackage{siunitx,array,booktabs,ragged2e}
\newrobustcmd\B{\DeclareFontSeriesDefault[rm]{bf}{b}\bfseries}
\newcolumntype{R}[1]{>{\RaggedLeft\arraybackslash}p{#1}}
\newcolumntype{L}[1]{>{\RaggedRight\arraybackslash}p{#1}}
\newcolumntype{C}[1]{>{\centering\arraybackslash}p{#1}}
\newcommand{\correct}[1]{\textcolor[rgb]{0,0.5,0}{#1}}
\newcommand{\wrong}[1]{\textcolor[rgb]{0.5,0,0}{#1}}

\newcommand{\TP}{\mathrm{TP}}
\newcommand{\FN}{\mathrm{FN}}
\newcommand{\FP}{\mathrm{FP}}
\newcommand{\TN}{\mathrm{TN}}
\newcommand{\TPR}{\mathrm{TPR}}
\newcommand{\FPR}{\mathrm{FPR}}

\newcommand{\alg}[1]{\texttt{#1}}
\newcommand{\none}{\alg{baseline}\xspace}
\newcommand{\noise}{\alg{noise}\xspace}
\newcommand{\AT}{\alg{AT}\xspace}

\newcommand{\pixmix}{\alg{pixmix}\xspace}
\newcommand{\regmixup}{\alg{regmixup}\xspace}
\newcommand{\PGD}{PGD$_{l_\infty}^{10}$\xspace}

\makeatletter
\ifreview
\renewcommand\maketitle{\hfill \titlebibdata\\\par\noindent {\Large \bf \@title}\\\par \noindent {\large \bf \@author}
  \newline\paragraph{Address:}\parbox[t]{6cm}{\addressfull}
  \newline\paragraph{Email:} {\tt \email}
  \newline\paragraph{ORCID ID:} 0000-0001-9531-2813
  \newline\paragraph{Submitted:} \today\newpage}
\else
\renewcommand\maketitle{\hfill \titlebibdata\\\par\noindent {\Large \bf \@title}\\\par \noindent {\large \bf \@author}\\\address\hfill {\tt \email}}
\fi
\makeatother

\begin{document}

\title{A Comprehensive Assessment Benchmark for Rigorously Evaluating Deep Learning Image Classifiers}
\author{Michael William Spratling}
\newcommand{\address}{King's College London, Department of Informatics, London. UK. and University of Luxembourg, 
Department of Behavioural and Cognitive Sciences, L-4366 Esch-sur-Alzette, Luxembourg}
\newcommand{\addressfull}{Department of Behavioural and Cognitive Sciences\\University of Luxembourg\\Maison Sciences Humaines\\11 Porte des Sciences\\L-4366 Esch-sur-Alzette\\Luxembourg}
\newcommand{\email}{michael.spratling@uni.lu}
\newcommand{\titlebibdata}{} 

\ifreview
\linenumbers
\fi

\maketitle
\ifreview \dson \fi


\begin{abstract}
  Reliable and robust evaluation methods are a necessary first step towards developing machine learning models that are themselves robust and reliable. Unfortunately, current evaluation protocols typically used to assess classifiers fail to comprehensively evaluate performance as they tend to rely on limited types of test data, and ignore others. 
  For example, using the standard test data fails to evaluate the predictions made by the classifier to samples from classes it was not trained on. On the other hand, testing with data containing samples from unknown classes fails to evaluate how well the classifier can predict the labels for known classes. 
  This article advocates benchmarking performance using a wide range of different types of data and using a single metric that can be applied to all such data types to produce a consistent evaluation of performance.
  Using the proposed benchmark it is found that current deep neural networks, including those trained with methods that are believed to produce state-of-the-art robustness, are vulnerable to making mistakes on certain types of data. This means that such models will be unreliable in real-world scenarios where  they may encounter data from many different domains, and that they are insecure as they can be easily fooled into making the wrong decisions.
  It is hoped that these results will motivate the wider adoption of more comprehensive testing methods that will, in turn, lead to the development of more robust machine learning methods in the future.
  
  
\end{abstract}

\paragraph{Keywords:} 
deep learning, neural networks, classification, generalisation, robustness, out-of-distribution, open-set, common corruptions, adversarial training, data augmentation

\paragraph{Code:} \url{https://codeberg.org/mwspratling/RobustnessEvaluation}

\ifreview\newpage\fi

\section{Introduction}
\label{sec-introduction}

\subsection{Need for Robustness}
\label{sec-need}

The tremendous success of deep learning (DL) makes it easy to overlook the fact that such models are often extremely brittle and insufficiently reliable for deployment in real-world scenarios, particularly in domains where security, safety, or trust are concerns \citep{Amodei_etal16,Heaven19,Serre19,YuilleLiu21,Bowers_etal23,Marcus20}. This lack of robustness is illustrated in \cref{fig-DL_failures}, using the domain of image classification as an example. Essentially DL, and more generally machine learning (ML) and artificial intelligence (AI), is poor at correctly generalising to novel data/situations. However, to be dependable it is essential that such methods be able to make accurate predictions on data that was not used for training, \ie for the trained model to generalise to novel, unseen, data.

\subsection{Types of Robustness}
\label{sec-types}

\begin{figure}[tbp]
  \begin{center}
    \begin{minipage}{0.1\textwidth}
      \vspace*{53mm}
        True class= \\
        Prediction= \\
    \end{minipage}
    \subfigure[]{\begin{minipage}{0.16\textwidth}\centering
        IID\\ Generalisation\\(clean accuracy)\\[2pt]
      \includegraphics[width=0.48\textwidth]{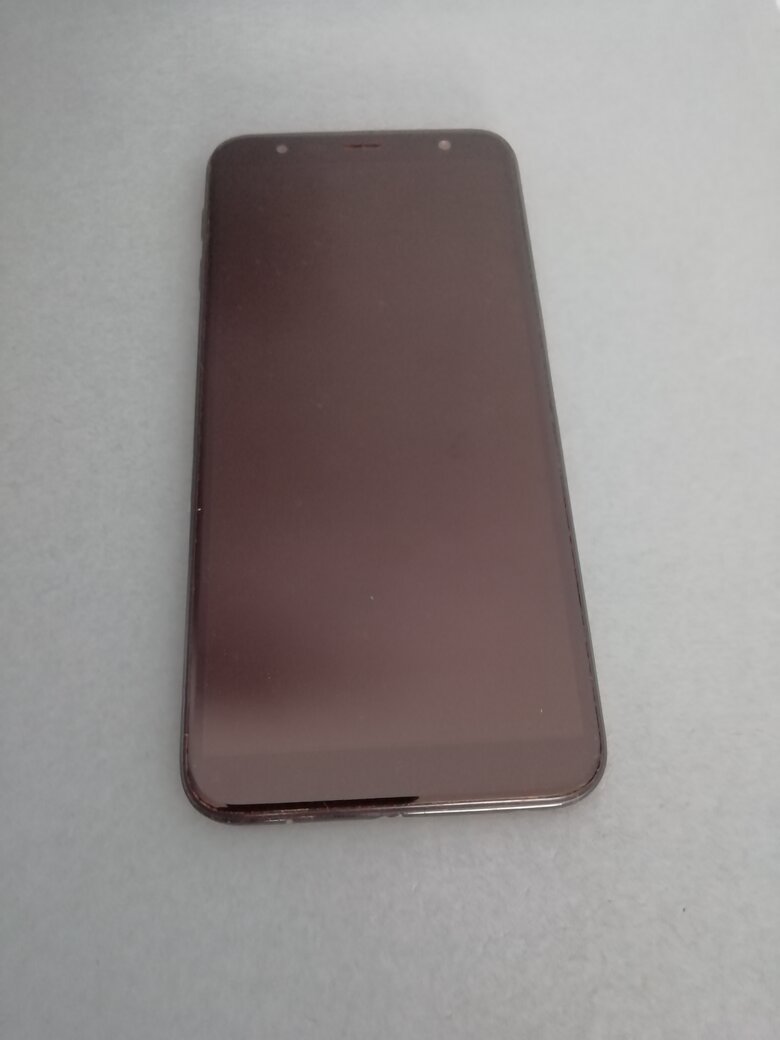}
      \includegraphics[width=0.48\textwidth]{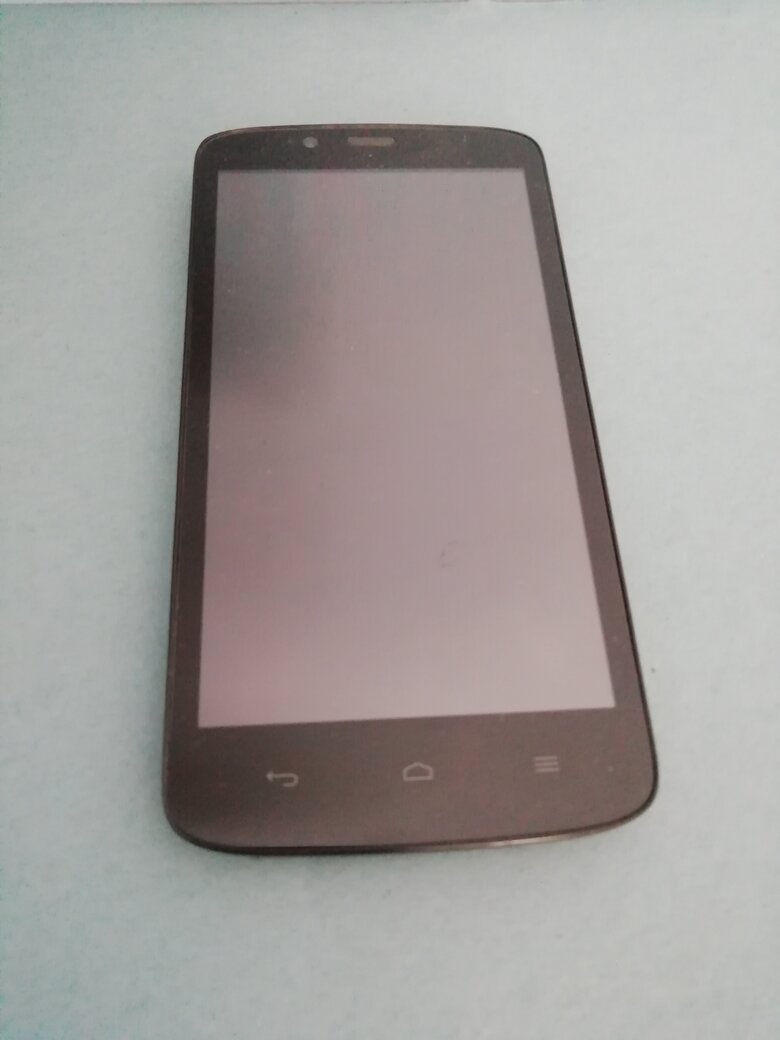}\\
      \includegraphics[width=0.48\textwidth]{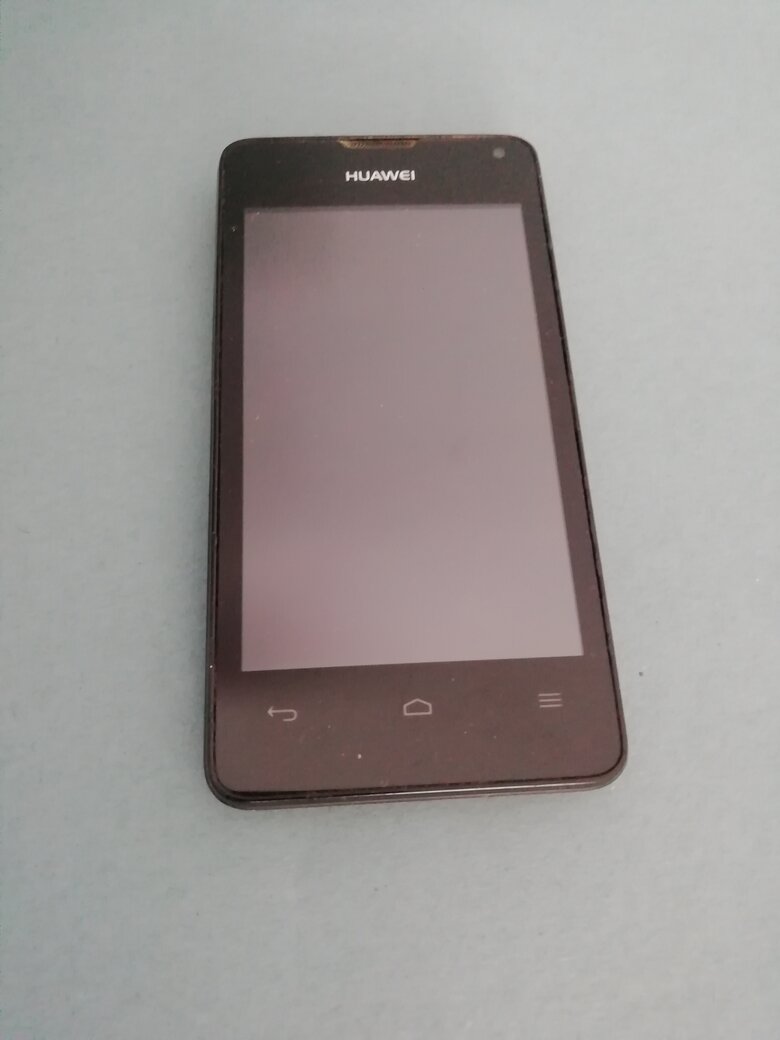}
      \includegraphics[width=0.48\textwidth]{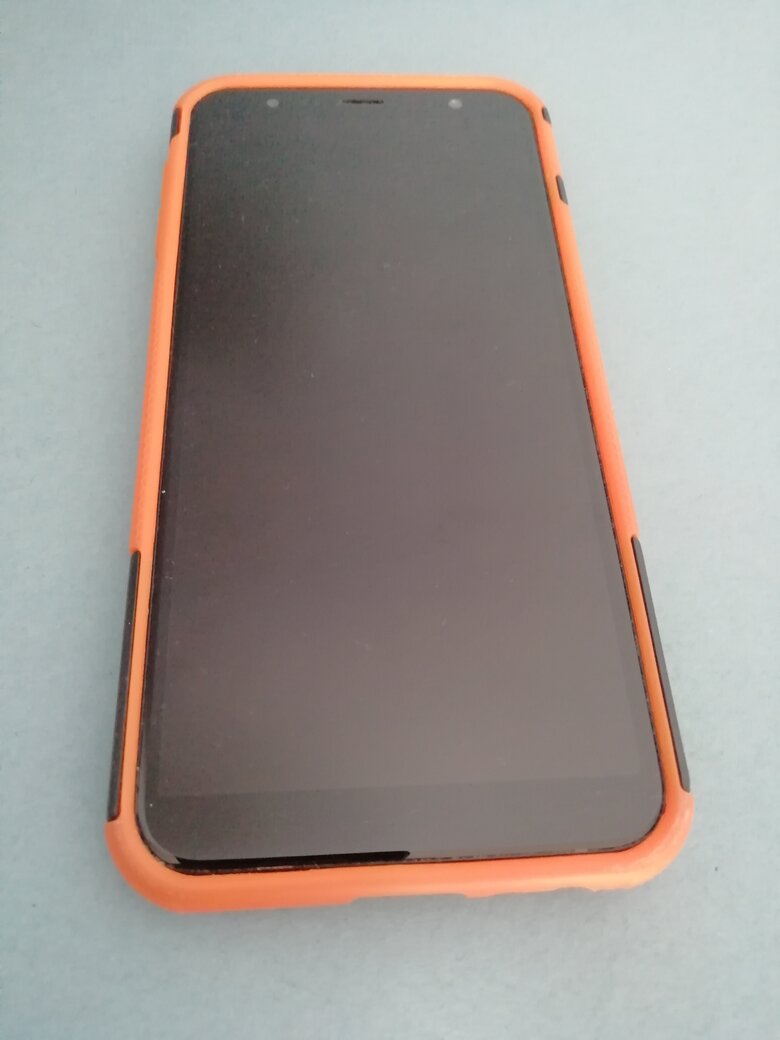}\\
      phone\\\cmarkw phone\cmark
    \end{minipage}}\hfill
    \subfigure[]{\begin{minipage}{0.16\textwidth}\centering
        OOD\\ Generalisation\\(corrupt accuracy)\\[2pt]
      \includegraphics[width=0.48\textwidth]{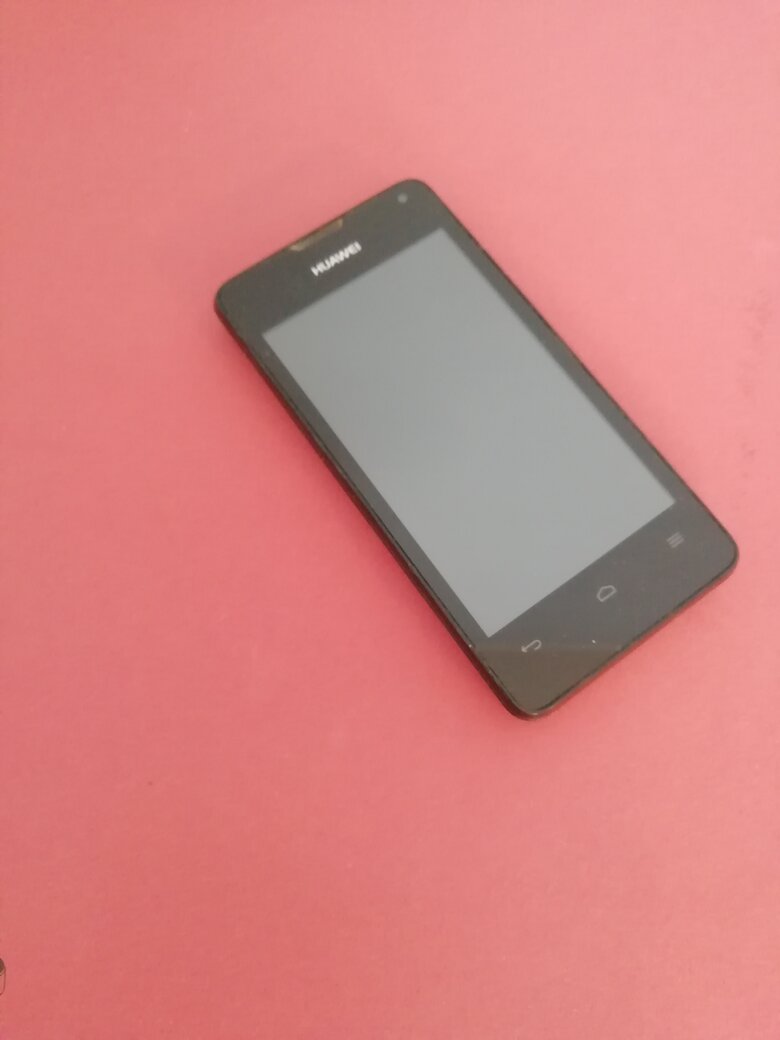} 
      \includegraphics[width=0.48\textwidth]{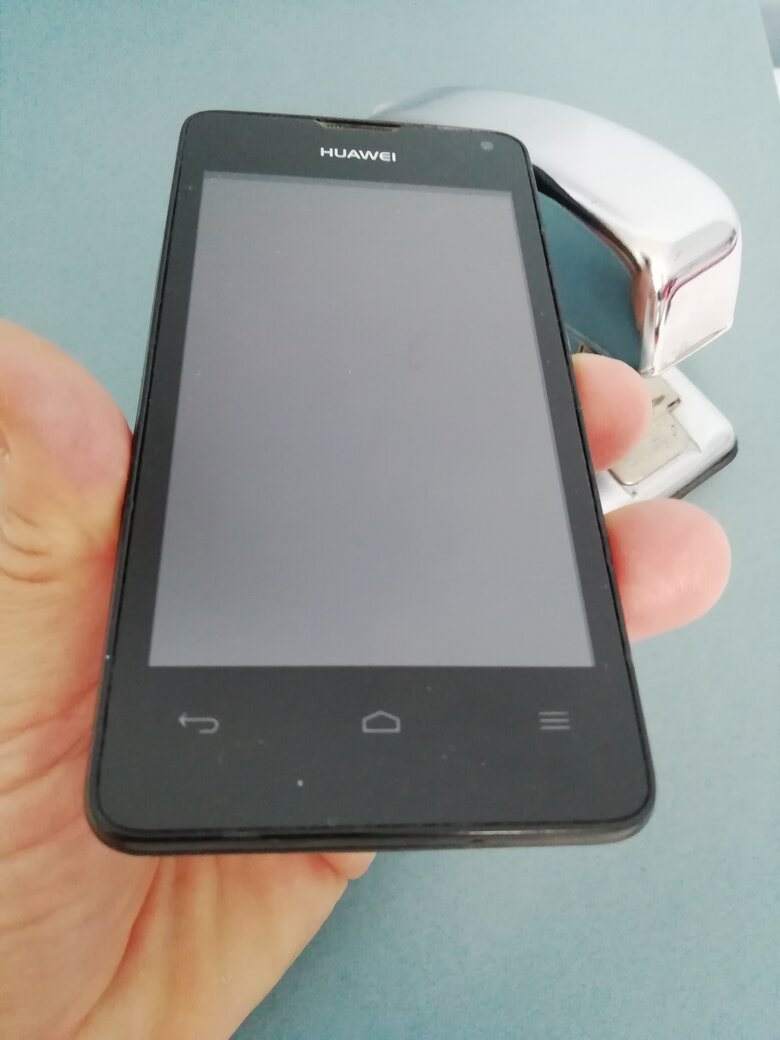}\\ 
      \includegraphics[width=0.48\textwidth]{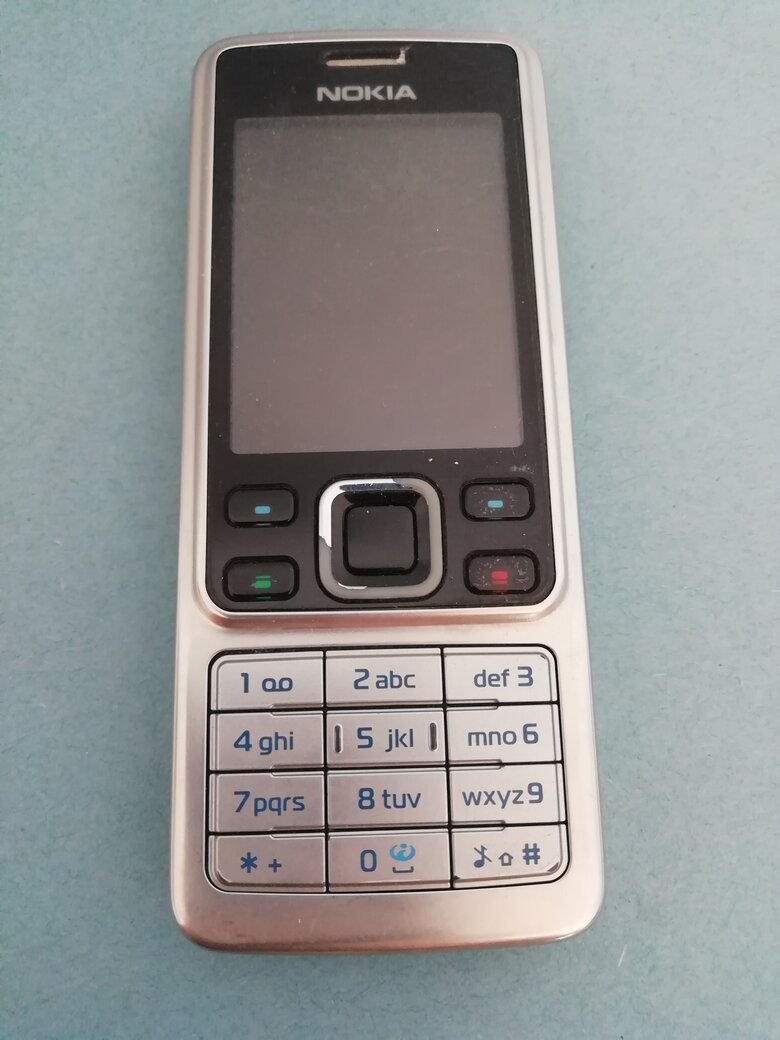} 
      \includegraphics[width=0.48\textwidth]{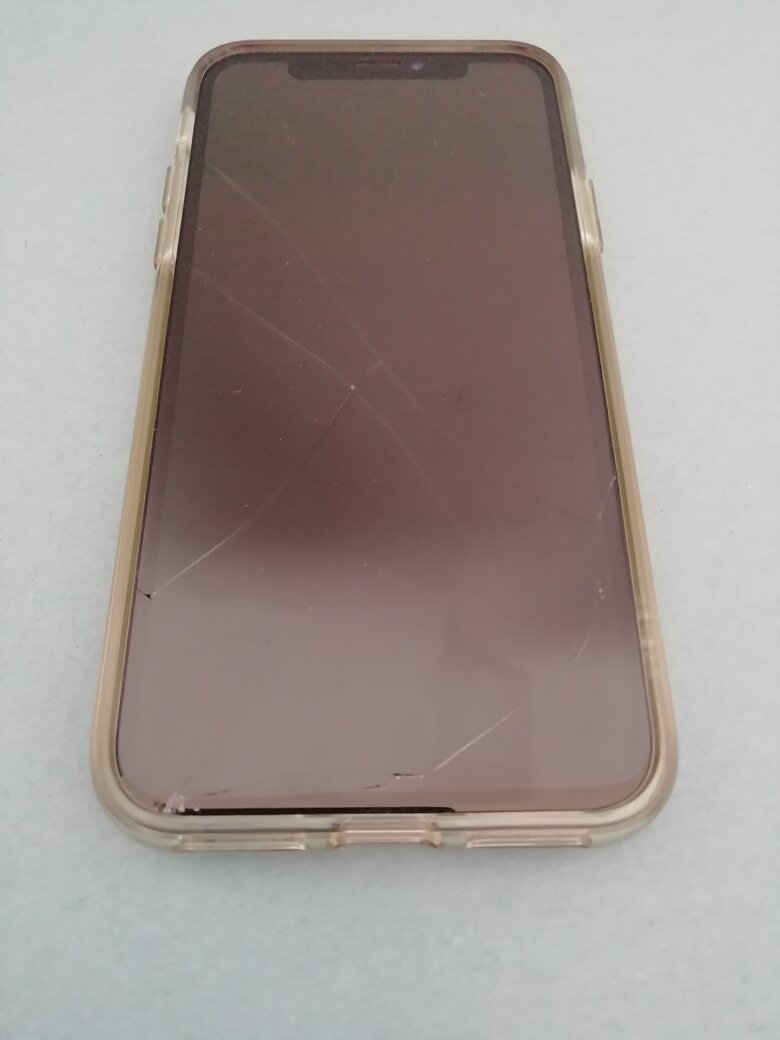}\\ 
      phone\\\xmarkw radio\xmark
    \end{minipage}}\hfill
    \subfigure[]{\begin{minipage}{0.16\textwidth}\centering
        Adversarial Robustness\\(adversarial acc.)\\[2pt]
      \includegraphics[width=0.48\textwidth]{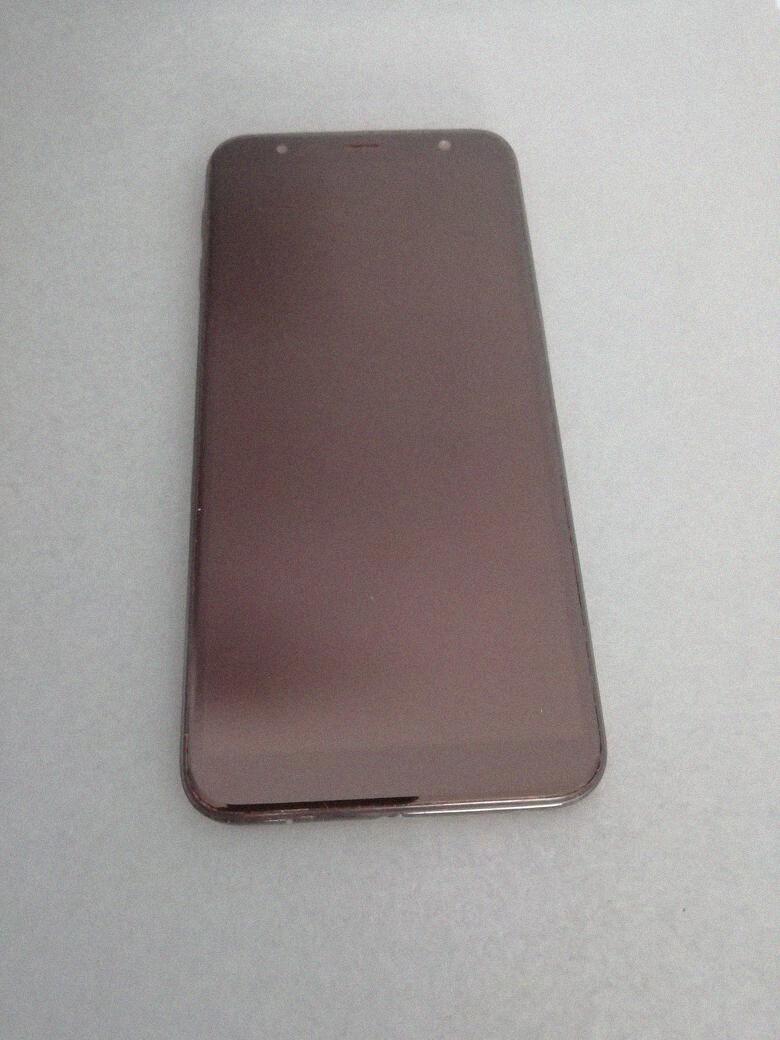}
      \includegraphics[width=0.48\textwidth]{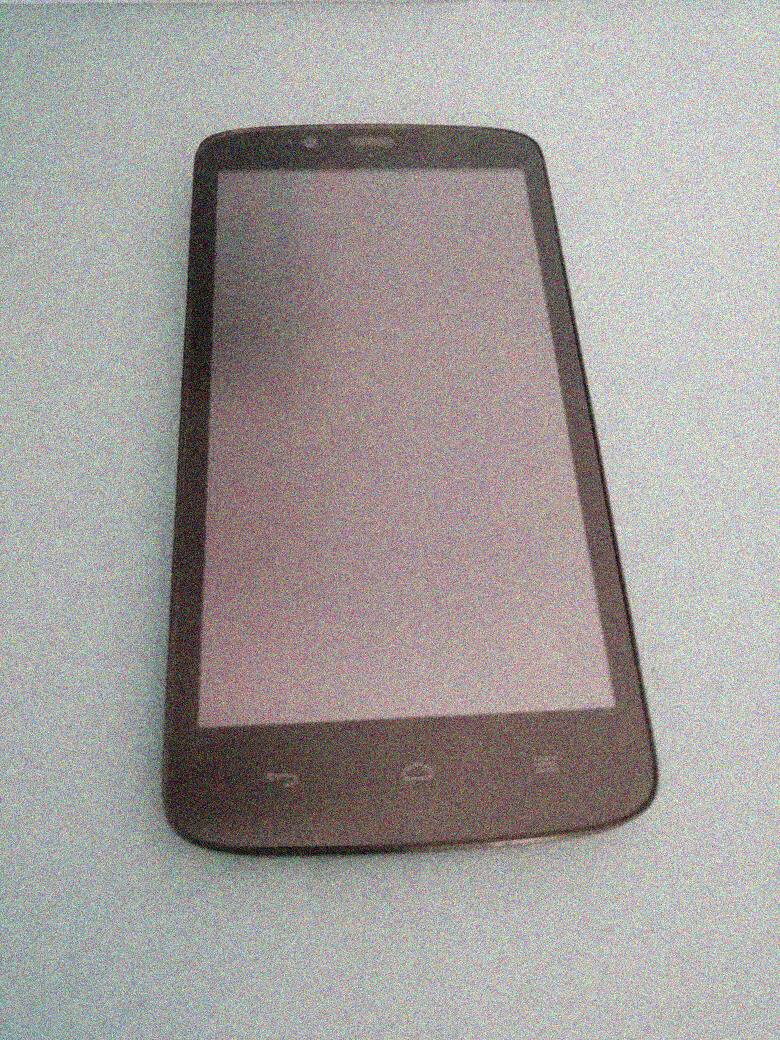}\\
      \includegraphics[width=0.48\textwidth]{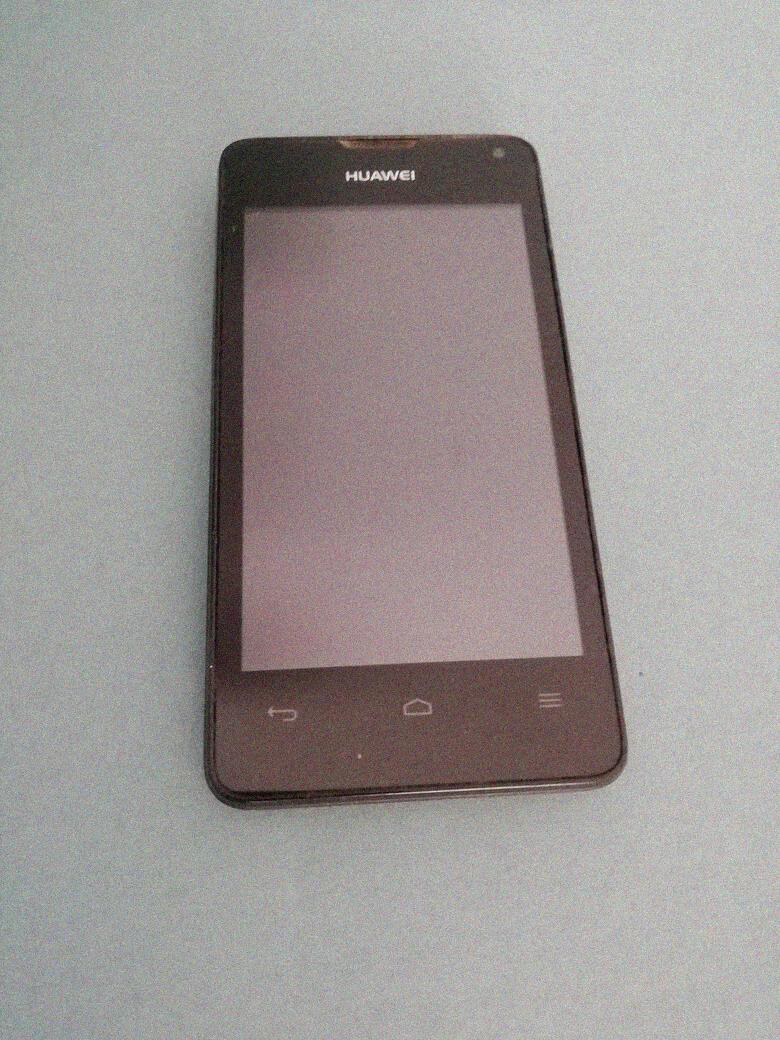}
      \includegraphics[width=0.48\textwidth]{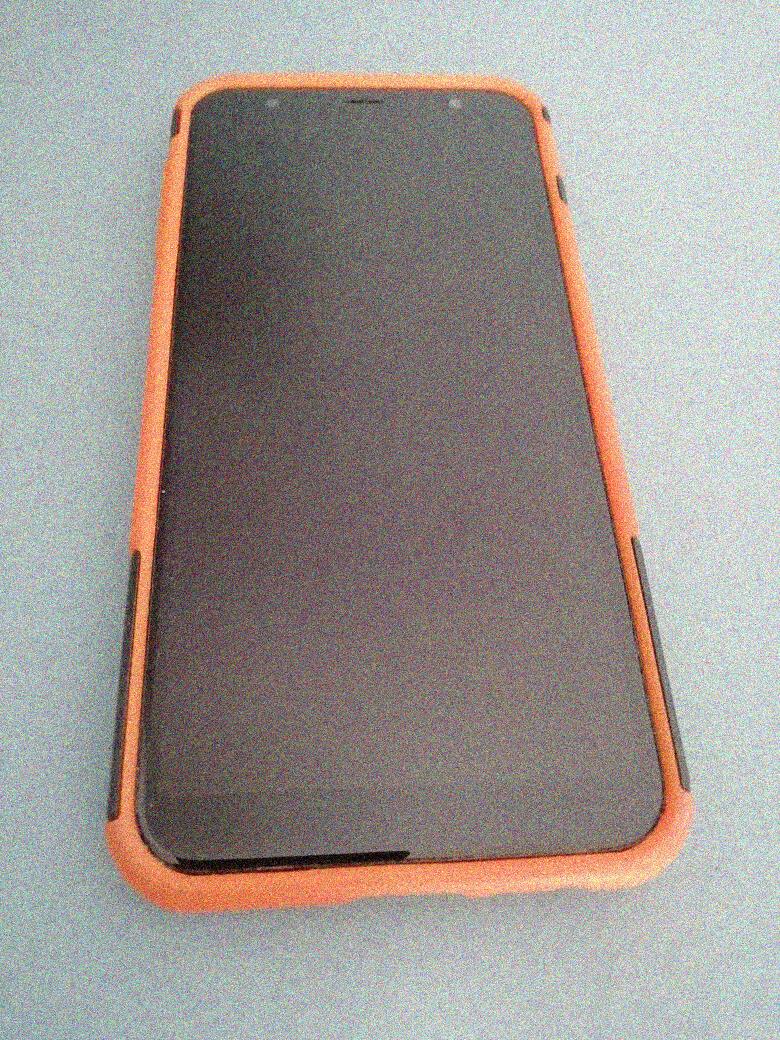}\\
      phone\\\xmarkw radio\xmark
    \end{minipage}}\hfill
    \subfigure[]{\begin{minipage}{0.16\textwidth}\centering
        OOD Detection\\(novel class rejection)\\[2pt]
      \includegraphics[width=0.48\textwidth]{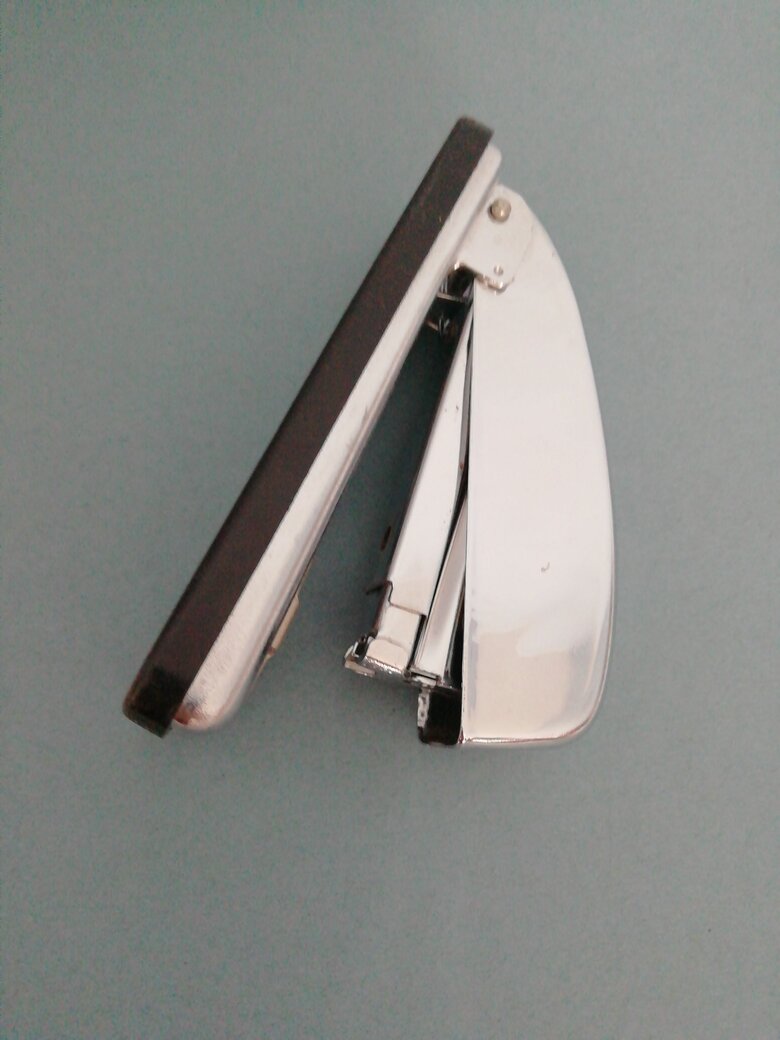} 
      \includegraphics[width=0.48\textwidth]{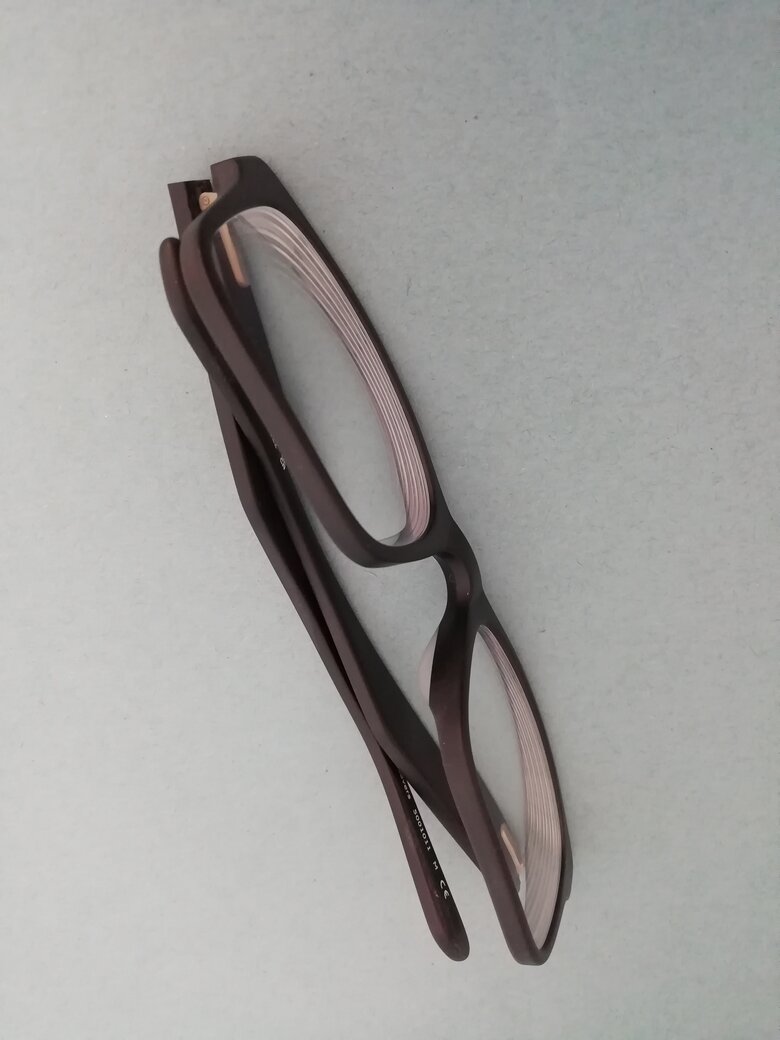}\\ 
      \includegraphics[width=0.48\textwidth]{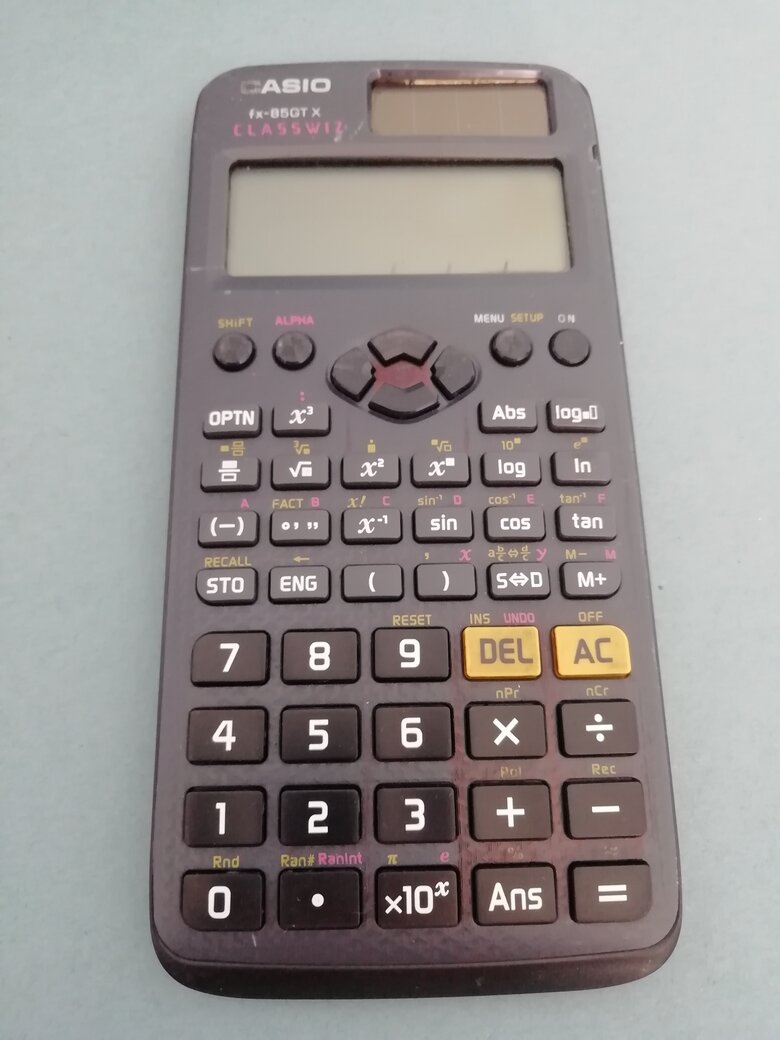} 
      \includegraphics[width=0.48\textwidth]{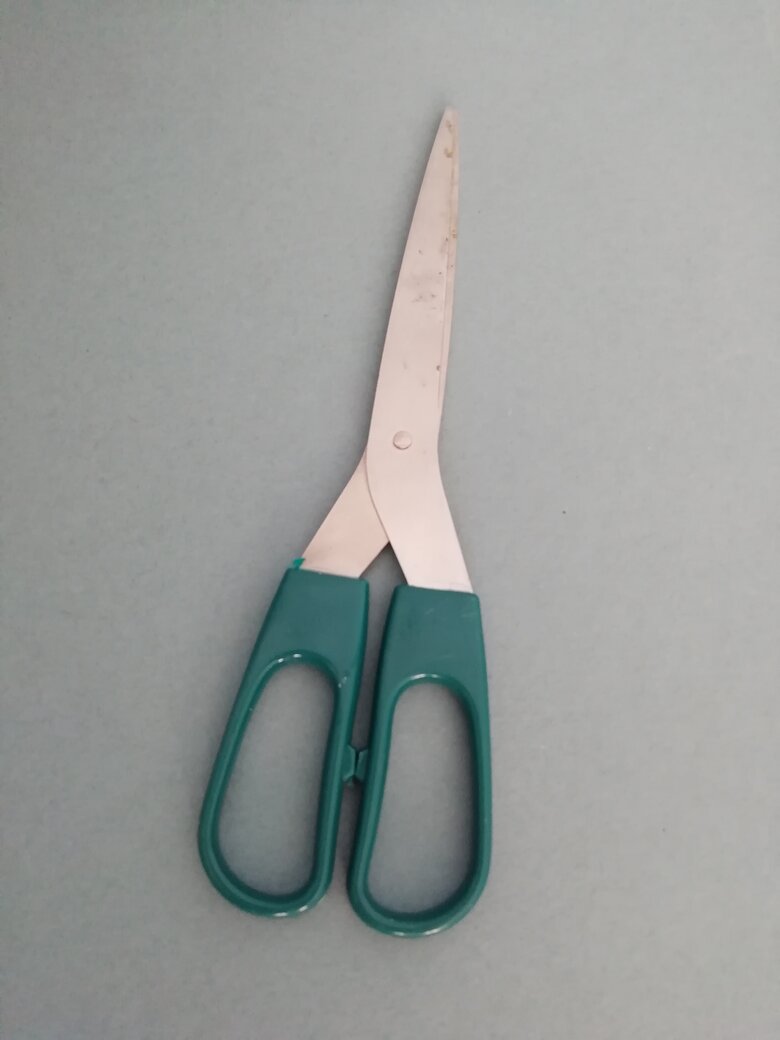}\\ 
      unknown\\\xmarkw phone\xmark
    \end{minipage}}\hfill
    \subfigure[]{\begin{minipage}{0.16\textwidth}\centering
        OOD Detection\\(unrecognisable image rejection)\\[2pt]
      \includegraphics[width=0.48\textwidth]{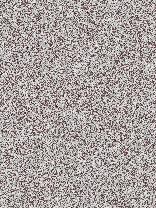}
      \includegraphics[width=0.48\textwidth]{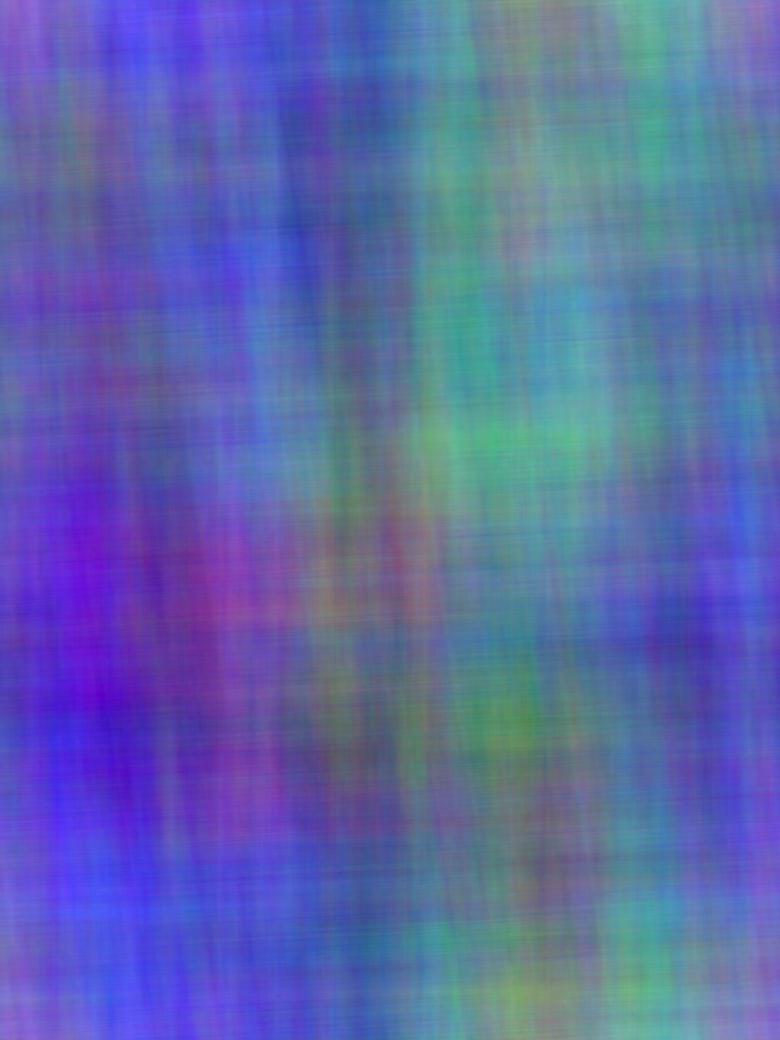}\\
      \includegraphics[width=0.48\textwidth]{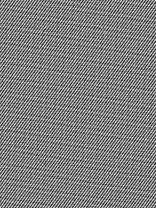}
      \includegraphics[width=0.48\textwidth]{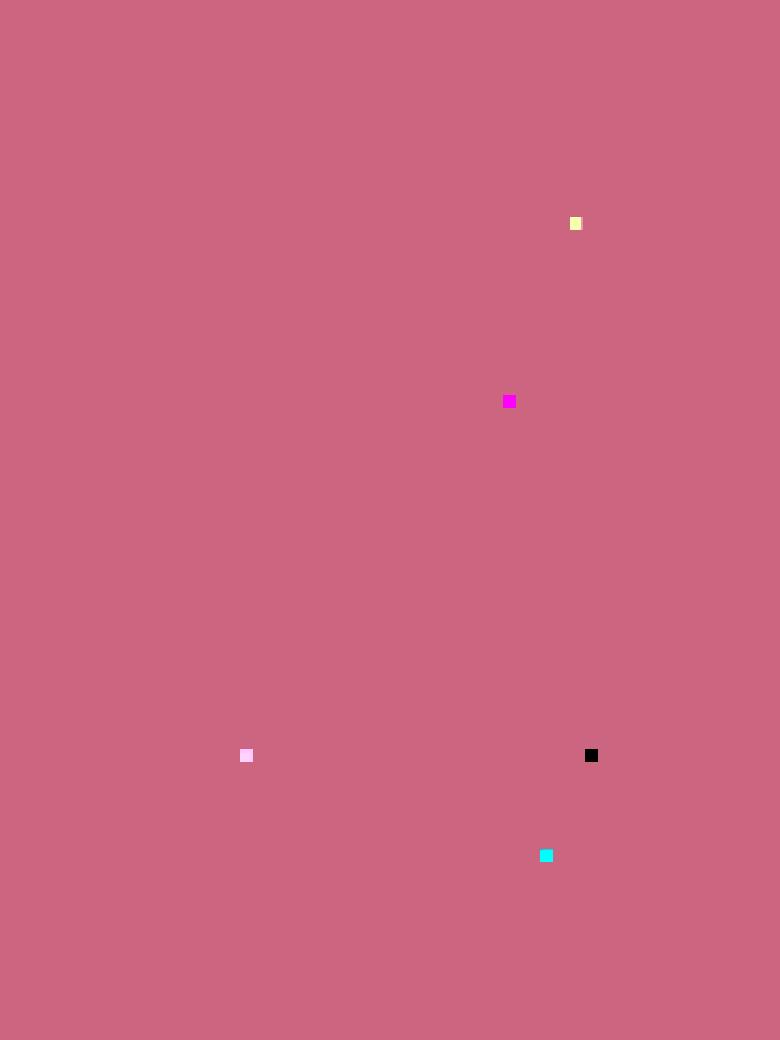}\\
      unknown\\\xmarkw phone\xmark
    \end{minipage}}

    \caption{A synthetic example of the sort of predictions made by a deep learning classifier when tested in different ways after being trained to recognise images of objects from categories including ``phone'' and ``radio''. Each sub-figure illustrates a different method of testing the accuracy of the classifier, it shows four sample images and the ground-truth and predicted classes for those images. (a) Samples from the test/validation data-set (this data looks very similar to the training data). (b) Samples used to assess generalisation. (c) Adversarially perturbed samples (the modifications to the images are virtually imperceptible but cause significant change to the classifier's prediction). (d) Samples showing objects from categories not included in the training data. (e) Samples that differ from images of natural objects or that have been synthetically generated to fool the classifier into making incorrect, high confidence, predictions. \citep[Figure inspired by Fig.~1 of][]{Kumano_etal22}.}
\label{fig-DL_failures}
\end{center}
\end{figure}

The ability to generalise can be assessed in multiple ways. Firstly, generalisation performance can be assessed using a test or validation set (see \cref{fig-DL_failures}a). Such data is typically similar to the training data as it is drawn from the same distribution in the input feature-space. This assessment method is thus said to measure in-distribution generalisation, or generalisation to independent and identically distributed (IID) samples. Such data is routinely used to assess the performance of classifiers.

A more challenging form of generalisation is out-of-distribution (OOD) generalisation (see \cref{fig-DL_failures}b). This can be assessed using test data that has different characteristics than the training data, but the same class labels. For example, in the domain of image categorisation it might be necessary for the classifier to cope with changes in appearance caused by images being captured under different conditions to those used to capture the training images. This can result in variations in orientation, scale, viewpoint, background, clutter, lighting, blur, jpg compression, image resolution, \etc. Furthermore, to succeed at OOD generalisation the classifier should ideally be tolerant to changes in appearance due to within-class variations, and for certain types of object (such as human bodies) non-rigid deformations. For image classification tasks, OOD generalisation can be assessed using clean samples that have been deliberately chosen because they differ from the training data \citep{Zhang_etal23,Hendrycks_etal21b,HendrycksDietterich19,Li_etal17} or using IID test images that have been synthetically modified, or corrupted, to simulate changes in image capture conditions \citep{Michaelis_etal19,Hendrycks_etal21,MuGilmer19,HendrycksDietterich19,JanssonLindeberg20}.

The limits placed on such generalisation may be task-dependent. For example, the image on the bottom-right of \cref{fig-DL_failures}b shows a phone with a smashed screen. If the task were classification then this image should be identified as a phone, even if no training image showed an exemplar with a broken screen. However, if the task being performed was anomaly detection (\ie finding defects in objects) then this exemplar would need to be identified as an anomaly.
OOD generalisation is closely related to the task of domain generalisation (DG), although DG typically employs multiple training data-sets from different domains to train a method to generalise to an unseen target domain. In turn DG is closely related to domain adaptation (DA), although DA allows access to (a limited amount of) training data from the target domain to improve model transfer \citep{Wang_etal22,RiazSmeaton23}.

A third method of assessing generalisation performance is through the use of adversarial exemplars (see \cref{fig-DL_failures}c). These are clean, in-distribution, samples that have been modified (or ``perturbed'') in such a way as to cause a change in the predicted class even when the class of the perturbed input appears unchanged to a human observer \citep{AkhtarMian18,Szegedy_etal14,Goodfellow_etal15,Kurakin_etal17,Eykholt_etal17,BiggioRoli18}. Adversarial perturbations have the appearance of low-amplitude random noise, and can be generated using a wide range of different algorithms \citep{Liang_etal22,Ren_etal20,Xu_etal20} which attempt to find small changes in the inputs that cause large changes in the predictions of the classifier.

A fourth requirement for robust ML is the ability to reject samples from categories that were not seen during training (see \cref{fig-DL_failures}d). Current DL models are susceptible to producing high-confidence, but incorrect, predictions to samples that are far from the training data distribution \citep{Amodei_etal16,HendrycksGimpel17}. The susceptibility of a classifier to making such errors can be evaluated simply by measuring its response to samples taken from data-sets containing classes distinct from those used for training. The task of creating classifiers that are less susceptible to making such errors is called open-set recognition \citep[OSR;][]{Vaze_etal22,Yang_etal22} or OOD detection/rejection \citep{Mohseni_etal20,HendrycksGimpel17,Bitterwolf_etal22,ZhangRanganath23,Hendrycks_etal22}. This task is also related to anomaly detection: as aim is to identify (and reject) novel or unexpected samples.

Finally, another situation in which an image classifier may produce high-confidence, but incorrect, predictions is when tested with exemplars that are unlike images of any object. Typically, synthetic images that occupy regions in feature-space that are far from those occupied by natural images, and hence, appear to human observers to have no similarity to the class label predicted by the classifier (see \cref{fig-DL_failures}e). Such images can be created using stochastic processes, or 
like adversarial exemplars, can be algorithmically generated to deliberately fool the classifier. This latter category, known as ``fooling images'', may look like noise \citep{HendrycksGimpel17}, or textures
\citep{Nguyen_etal15}, or images that are uniform except for a few differently coloured pixels \citep{Kumano_etal22}.


The two images at the bottom-left of \cref{fig-DL_failures}b and d, illustrate a fundamental difficulty of accurate classification. It is necessary to have tolerance to within-class variation, and other changes in image capture conditions, so that visually dissimilar images can be allocated to the same class, while simultaneously being able to make fine-grained distinctions so that similar looking images from different classes can be distinguished \citep{Bar04}.

\subsection{Terminology}
\label{sec-terminolgy}



OOD generalisation is concerned with classifying data where the distribution of the input (\ie the characteristics of the images) has changed, but the distribution of output (\ie the range of class labels associated with the images) is constant. In contrast OOD rejection is concerned with identifying data where the distribution of both the input and output has changed. The term ``distribution'' is therefore being used to refer to different distributions in each case. For OOD generalisation it is the distribution of the input, while for OOD rejection it is primarily the distribution of the output (the change in the input distribution is simply a side-effect of the change to new class labels). To avoid any possible confusion, OOD detection/rejection will hence-forth be referred to as ``Unknown Class Rejection''. The data used to assess unknown class rejection will be subdivided (as in \cref{fig-DL_failures}) into ``novel'' classes (images containing objects from classes unseen during training) and ``unrecognisable'' images (those containing no sematically meaningful information).

The term ``distribution'' when applied to the input is also poorly defined and, arguably, irrelevant when discussing IID/OOD generalisation.
All data other than that included in the training set is OOD for the classifier \citep{Guerin_etal22}, so the distinction between IID and OOD data is qualitative rather than quantitative, and the boundary between in- and out-of distribution is necessarily arbitrary. Furthermore, the shift in distribution in image-space often has little correlation with the difficulty in correctly classifying an image. For example, two images of the same object taken from different viewpoints may have very little similarity in terms of corresponding pixel values, but the images may be easily identified as containing the same object. In contrast, it may be very difficult for the classifier to assign the same label to two images that are almost identical in image-space (\eg a clean image and an adversarially perturbed version of that same image). Most importantly, how far an image is from the training distribution is irrelevant to the task of the classifier. Ideally, the classifier should be able to successfully recognise novel exemplars of the objects it was trained on regardless of how similar or dissimilar they are in appearance to the images used in training. In contrast an ideal classifier should reject images containing unknown objects even if such images are close to the training distribution. Hence, rather than referring to samples as IID and OOD, the remaining text will refer to such data as being from ``known'' classes. The broad class of known data will be sub-divided (as in \cref{fig-DL_failures}) into different types of test data-set: ``clean'' data from the standard test set, ``corrupt'' for samples from the standard test set that have been manipulated to test for generalisation to changes in viewing conditions, and ``adversarial'' for samples that have been adversarially-perturbed.



The terms ``generalisation'' and ``robustness'' both refer to a classifier's ability to produce the correct predictions. These two terms will, therefore, be used interchangeably. They will be used not only to refer a classifier's ability to predict correct class labels for samples from known categories, but also a classifier's ability to correctly reject samples from unknown classes.

\subsection{Issues with Current Assessment Methods}
\label{sec-practice}

While there has been a great deal of previous work carried out on assessing and improving all forms of robustness described in \cref{sec-types}, almost all of this work has considered one form of generalisation in isolation. For example, generalisation to clean test data has been the sole pre-occupation of most research in the history of ML so far. This has resulted in methods that produce exceptional performance on test/validation data, but that are poor at generalisation to changes in appearance \citep{Recht_etal18,Recht_etal19,YadavBottou19,Geirhos_etal18,Sa-CoutoWichert21}, that are highly susceptible to adversarial attacks \citep{Ilyas_etal19,Papernot_etal16,AkhtarMian18}, that are prone to finding short-cuts to improve clean accuracy at the expense of learning more generally useful information \citep{Geirhos_etal20,Amodei_etal16,Lapuschkin_etal19,Malhotra_etal20,Geirhos_etal18}, and that produce high confidence predictions to samples that do not belong to any of the known categories \citep{Amodei_etal16,HendrycksGimpel17,Kumano_etal22,Nguyen_etal15}.
Similarly, most previous work on unknown class rejection considers this task in isolation, and does not evaluate models on clean test data let alone assess generalisation to corrupted or adversarial known classes \citep{Yang_etal21}.

Some work has concurrently considered different types of robustness. For example, most work on adversarial robustness also evaluates clean accuracy. In addition, the RobustBench benchmark also considers generalisation to corupt data \citep{Croce_etal21}, but only considers robustness to one type of data when ranking models. Other work has considered: clean accuracy and generalisation to corruptions \citep{Vasconcelos_etal21}; generalisation to input distribution shifts and unknown class rejection \citep{Khazaie_etal22,Yang_etal23b}; clean accuracy, generalisation to corruptions and unknown class rejection \citep{Pinto_etal23,Zhang_etal23b,AverlyChao23}; adversarial robustness and unknown class rejection \citep{Song_etal20,Feng_etal22}; clean accuracy, adversarial robustness and unknown class rejection \citep{Feng_etal22,Shao_etal20}; generalisation to corruptions, adversarial robustness, and unknown class rejection \citep{Guerin_etal23,KyungpilJoonhyuk23}; or clean accuracy, adversarial robustness and generalisation \citep{Geirhos_etal20,Sun_etal21,Rusak_etal20,Ford_etal19}. \citet{Hendrycks_etal19b} considers clean accuracy, generalisation to corruption, adversarial robustness, and unknown class rejection, but each type of robustness is assessed in isolation for models that have been trained in slightly different ways. Similarly, \citep{Mohseni_etal20} considers both clean accuracy and unknown class rejection, however, the clean accuracy of the classifier is not measured after it is modified to learn to reject unknown classes. \citet{Guille-Escuret_etal23} benchmarks performance using corrupt, adversarial, and unknown class images, but only considers the model's ability to detect input distribution shifts and does not measure the accuracy with which known classes are classified. 


Previous work that has considered multiple types of robustness suggests that there are trade-offs. For example, maximising clean accuracy results in poor adversarial robustness, while increasing adversarial robustness decreases clean accuracy \citep{Zhang_etal19,Tsipras_etal19,Madry_etal18,Kannan_etal18}.
Similarly, many adversarial defences do not improve robustness to image corruptions 
and methods that improve generalisation to corrupt data typically fail to produce a strong defence against adversarial attack \citep{Sun_etal21,Rusak_etal20,Ford_etal19}. Improving adversarial robustness has also been shown to make a classifier less sensitive to changes in object class \citep{Tramer_etal20,Rauter_etal23}, and less able to perform unknown class rejection \citep{Song_etal20}.
While \citet{Vaze_etal22} show that there is generally a correlation between clean accuracy and the ability to detect unknown classes, it is not hard to imagine that methods could be developed to reject samples from unknown classes that will also result in the rejection of known samples (especially ones, such as corrupt images, that differ in appearance from the training samples) \citep{Khazaie_etal22,Aboitiz_etal23}. 
The converse would be a model which has been designed to be tolerant to changes in appearance in order to improve generalisation to corruptions: such a model is likely to classify more images of unknown objects with high certainty.

These trade-offs in performance on different data types raise the concern that efforts to increase robustness in one area may be resulting in a decrease in robustness in another area. However, as models are not comprehensively assessed for all different types of generalisation performance, such issues will not be noticed. Failure to fully assess the performance of models may result in huge efforts being wasted producing models that give a false sense of security by appearing to be robust only because their performance has not been tested on data for which they perform poorly. 

\subsection{Issues with Current Metrics}  
\label{sec-metrics}

As outlined in the previous section, there is a need for a more comprehensive assessment of robustness, using multiple different types of test data-set as illustrated in \cref{fig-DL_failures}. However, metrics that are most commonly used to assess each individual form of robustness are inadequate for producing more comprehensive and rigorous assessment.





\begin{figure}[tbp]
  \begin{center}
    \subfigure[]{\includegraphics[scale=0.55, trim=300 250 300 90,clip]{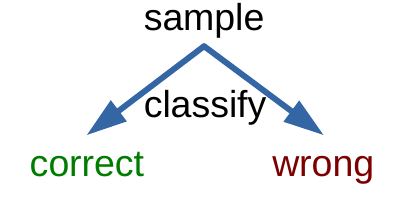}\label{fig-metrics_acc}}\hspace*{0.075\textwidth}
    \subfigure[]{\includegraphics[scale=0.55, trim=300 250 300 90,clip]{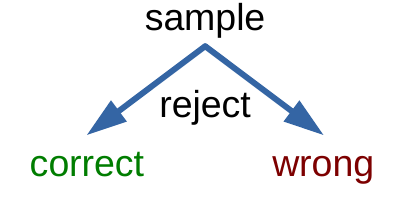}\label{fig-metrics_ood}}\hspace*{0.075\textwidth}
    \subfigure[]{\includegraphics[scale=0.55, trim=250 180 250 90,clip]{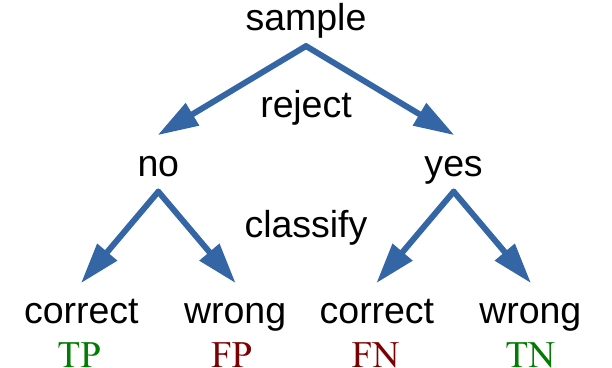}\label{fig-metrics_new}}
    \caption{Decision trees illustrating standard assessment methods for (a) known class data, and (b) unknown class rejection. (c) The method proposed by \citet{Zhu_etal22} for dealing with test data from both known and unknown classes.}
    \label{fig-metrics}
  \end{center}
\end{figure}

Performance on known classes (clean, corrupt, and adversarial samples) is typically evaluated by calculating the classification accuracy (or equivalently the percentage error), see \cref{fig-metrics_acc} and \cref{tab-confusion_matrix_acc}, although many other metrics can also be used \citep{Guo_etal23}. However, if models are to be allowed to reject samples, to deal with those from unknown classes, then the performance on known samples should also be assessed taking into account that some samples may be rejected. On the other hand, models designed to perform unknown class rejection \citep{Kirchheim_etal22,Chen_etal23,Cheng_etal23,Xu-Darme_etal23,Yang_etal22,Yang_etal23,Lee_etal22}, and adversarial defences that reject perturbed samples \citep{Stutz_etal20,Ren_etal22,Lee_etal22,Chen_etal22}, are typically evaluated purely in terms of how accurately samples are rejected and accepted. In other words, the task is considered to be that of binary classification (see \cref{fig-metrics_ood} and \cref{tab-confusion_matrix_ood}). Hence, standard metrics for binary classification can be used, such as the False Positive
Rate when the true positive rate is X\% (FPR@X\%), which assesses performance for a single rejection threshold, or Area Under the Receiver Operating Characteristic curve (AUROC) or 
Area Under the Precision-Recall curve (AUPR) both of which assess performance over varying rejection thresholds.
An issue with FPR@X\% is that separate thresholds are used for distinguishing each unknown data-set from the clean test data. Worse still, some rejection methods have hyper-parameters that are tuned separately on each unknown data-set \citep{Zhang_etal23}. This is unrealistic, as in the real-world a single rejection criteria with fixed hyper-parameters would need to work to reject any unknown data, regardless of its origin \citep{Shafaei_etal18,Feng_etal22}. An issue with all standard unknown class rejection metrics is that the ability of the model to accurately classify the accepted samples is not evaluated. In some publications the clean accuracy of the model is evaluated, but this is done without any samples being rejected, and hence, does not reflect the performance that would be expected in the real-world where the same rejection criteria would be applied to all samples.
  As a result of this issue, a model that could accurately reject unknown classes would score highly with the standard metrics, even if that classifier assigned a random label to all accepted samples making it virtually useless in any practical application.

\begin{table}[tbp]
  \begin{center}
  \caption{Confusion matrices illustrating assessment methods for (a) known class data, (b) unknown class rejection, and (c) the method proposed by \citet{Zhu_etal22}. For (a) performance is typically evaluated by calculating the proportion of correct predictions. For (b) performance is typically evaluated by calculating the $\FPR=\frac{\FP}{\FP+\TN}$ when the rejection threshold has been set to produce a predetermined $\TPR=\frac{\TP}{\TP+\FN}$, where $\TP$, $\FN$, $\FP$, and $\TN$ are the total number of true-positives, false-negatives, false-positives, and true-negatives, respectively. (c) Considers both if the sample has been correctly accepted or rejected, and whether the predicted class label is correct or incorrect.}
  \label{tab-confusion_matrix}
  \subfigure[]{
      \begin{tabular}{ll|c|c|c|c|} 
        & & \multicolumn{4}{c|}{\bf predicted class} \\ 
        & &  1 & 2 & \dots & c \\
        \hline
        \multirow{4}{*}{\rotatebox[origin=r]{90}{\bf true class}}
        & 1     & \cmark & \xmark & & \xmark \\\cline{2-6}
        & 2     & \xmark & \cmark & & \xmark \\\cline{2-6}
        & \vdots& & & & \\\cline{2-6}
        & c     & \xmark & \xmark & & \cmark \\\cline{2-6}
        \hline
      \end{tabular}
      \label{tab-confusion_matrix_acc}
  }\hspace*{0.015\textwidth}
  \subfigure[]{
      \begin{tabular}{ll|C{13mm}|C{13mm}|} 
        & & \multicolumn{2}{c|}{\bf predicted} \\ 
        & &  accept & reject \\
        \hline
        \multirow{2}{*}{\rotatebox[origin=c]{90}{\bf true}}
        & known & \correct{$\TP$} & \wrong{$\FN$} \\\cline{2-4}
        & unknown & \wrong{$\FP$} & \correct{$\TN$} \\
        \hline
      \end{tabular}
      \label{tab-confusion_matrix_ood}
  }\hspace*{0.015\textwidth}
  \subfigure[]{
      \begin{tabular}{llll|C{13mm}|C{13mm}|} 
        & & & & \multicolumn{2}{c|}{\bf predicted class} \\ 
        & & & &  correct & wrong \\
        \hline
        \multirow{4}{*}{\rotatebox[origin=c]{90}{\bf true}}
        & \multirow{2}{*}{known}   & \multirow{4}{*}{\rotatebox[origin=c]{90}{\bf predicted}}& accept & \correct{$\TP$} & \wrong{$\FP$} \\\cline{4-6}
        &                          & & reject & \wrong{$\FN$} & \correct{$\TN$} \\\cline{2-2}\cline{4-6}
        & \multirow{2}{*}{unknown} & & accept & N/A & \wrong{$\FP$} \\\cline{4-6}
        &                          & & reject & N/A & \correct{$\TN$}\\
        \hline
      \end{tabular}
      \label{tab-confusion_matrix_new}
  }
  \end{center}
\end{table}

Some metrics do attempt to evaluate classification performance in the presence of rejection. For example, in the context of evaluating accuracy on adversarial data,
confidence-thresholded robust error \citep{Stutz_etal20} measures the classification error for test examples that are not rejected, when the rejection threshold has been set so that 1\% of correctly classified clean examples can be rejected.
Using the metric robust risk with detection \citep{Tramer21} a classifier is accurate if it correctly classifies
the clean samples, and either rejects or correctly classifies perturbed samples.
However, with these metrics a high score can be obtained by a classifier that rejects a large proportion of the perturbed samples, even though these samples come from a known class and may be only perturbed imperceptibly \citep{Chen_etal22}.
To address this issue, robustness with rejection \citep{Chen_etal22} defines a variable perturbation threshold that can vary from zero to the adversarial attack budget. For test samples with a perturbation less than or equal to the threshold, both incorrect classification and rejection is counted as an error. For test samples with a perturbation greater than the threshold, incorrect classification is considered an error, but rejection is not counted as an error. When the threshold is greater or equal to the attack perturbation budget, robustness with rejection is equivalent to standard robust error. When the threshold is zero this metric is equivalent to robust risk with detection as proposed by \citet{Tramer21}, discussed earlier.


An issue with all the methods discussed in the previous paragraph, is that they define which samples can be rejected in terms of an arbitrary measure of distribution shift in the input space \citep{Guerin_etal23}. This means that these methods would not allow for 
rejection of non-adversarial samples. For example, in the situation where the sample is clean but ambiguous (\eg in an optical character recognition task a letter written in terrible handwriting) rejection would be counted as an error even though the distribution shift for such a sample might be far higher than that of any adversarial sample. These metrics would also be incapable of assessing performance on data-sets containing unknown samples. In this situation the correct action of the classifier is to reject all samples, however, for confidence-thresholded robust error \citep{Stutz_etal20} a score is only calculated for accepted samples, so no score could be calculated for a classifier correctly rejecting all unknown samples. Both robustness with rejection \citep{Chen_etal22} and robust risk with detection \citep{Tramer21} would suffer from the same issue, assuming unknown samples were considered to be sufficiently perturbed to be out-of-distribution. Alternatively, if clean samples from unknown classes were considered to be unperturbed, these metrics would give a score of zero to a classifier that correctly rejected all samples, as any rejection of clean sample is counted as an error for these metrics.

It would be more reasonable to decide whether or not rejection or acceptance of the sample was the correct choice depending on whether or not that sample would have been mis-classified or correctly classified \citep{Pang_etal22,AverlyChao23}, as the goal of sample rejection is to protect the classifier from making predictions that are unreliable, not to identify samples that differ from the training data \citep{Guerin_etal23,Zhu_etal22}.
Taking this approach, \citet{Zhu_etal22} propose a metric that they apply to evaluate the performance of classifiers using both corrupt samples and unknown samples from novel classes. As shown in \cref{fig-metrics_new} and \cref{tab-confusion_matrix_new}, unlike standard metrics, this approach considers both whether or not a sample is rejected, and whether or not the classification label predicted by the classifier is correct. In this scheme, a sample is considered a true-positive ($\TP$) if it is not rejected and assigned the correct class label. A sample is a false-positive ($\FP$) if it is accepted but incorrectly classified. A false-negative ($\FN$) is a sample that is rejected although it would have been assigned the correct class label. Finally, true-negatives ($\TN$) are samples that are rejected and that would have been classified incorrectly. The proposed metric is the detection error rate (DER), which counts the number of false positives and false-negatives as a proportion of all test samples. \citet{Zhu_etal22} apply their metric using two rejection thresholds, that are set such that either 95\% or 99\% of 
 correctly classified clean examples are accepted. These different values for the rejection threshold define the trade-off between accuracy on the clean data (perfect performance on the clean data could only be achieved if 100\% of correctly classified clean examples were accepted) and correct rejection of unknown samples (good performance on unknown class rejection is more likely if more clean samples are also rejected).

\section{Methods: The Proposed Comprehensive Assessment Benchmark}
\label{sec-methods}

\subsection{Assessment Method}
\label{sec-metric}

The proposed benchmark uses a metric to evaluate performance that is an extended and modified version of the evaluation method proposed by \citet{Zhu_etal22}, described in the preceding section. The proposed modifications are as follows:
\begin{enumerate}
\item Measuring performance in terms of accuracy rather than error. The metric defined by \citet{Zhu_etal22}, DER, is a measure of the proportion of errors (expressed as a percentage) made by the classifier. Here, the percentage of correctly processed samples is reported. 
  This simple change means that a higher score correlates with improved performance which is more intuitive, and is more 
  consistent with previously used metrics (such a AUROC and those used in RobustBench \citep{Croce_etal21}) where this is also the case.  Hence, rather than using detection error rate (DER), detection accuracy rate (DAR) is reported.

\item Summarising results by averaging over task, not data-set. \citet{Zhu_etal22} report results for each data-set separately. However, using many separate metrics makes it difficult to evaluate the overall robustness of any model especially when there are trade-offs in performance of different data-sets (meaning that no method of improving robustness is likely to have strong performance across all data-sets). It is therefore convenient to have a single metric to summarise performance. For this purpose, \citet{Zhu_etal22} average the performance across each data-set. However, there is a risk that a more extensive evaluation of robustness using more data-sets might result in this average becoming biased towards one type of robustness. For example, as there are potentially a very large number of data-sets containing images from unknown classes, including more such data-sets would result in the average performance favouring methods better able to reject unknown classes. Alternatively, calculating a single value of the metric using all data from all data-sets would be biased towards favouring methods better able to deal with data-sets containing the most samples.
    To better balance the importance of robustness to each type of data, here a single summary metric is obtained by calculating the metrics separately for each type of data, and then averaging across data type
\footnote{As the sub-sets of data used for each type of data (see \cref{sec-data}) contain similar
numbers of exemplars, the DAR for each type of data is calculated for all the data of that type. However, if the metric was to be used in future with less well-balanced sub-data sets, it would be sensible to calculate the DAR speartely for each sub-set and then average over those results to produce the estimate of DAR for that data type.}.
    \item Extending the evaluation to consider other forms of generalisation. \citet{Zhu_etal22} evaluate performance using three types of test data: clean, corrupt, and novel. In the current work, unknown class rejection is evaluated not only using images from novel classes, as in \citep{Zhu_etal22}, but also using unrecognisable images. Furthermore, performance is also evaluated using adversarial samples. In other words, here performance is evaluated using the five types of data illustrated in \cref{fig-DL_failures}, whereas \citet{Zhu_etal22} use only three. The results reported in \cref{sec-results} highlight the necessity of expanding the range of test data types used. The poorest performance of the tested networks is almost always for either adversarial samples or unrecognisable images, \ie for the data-types that \citet{Zhu_etal22} do not use. Hence, the benchmark proposed by \citet{Zhu_etal22} is more likely to fail to identify robustness issues than the benchmark proposed here.

\end{enumerate}

\subsection{Choice of Test Data}
\label{sec-data}

\newcolumntype{C}[1]{>{\centering\let\newline\\\arraybackslash\hspace{0pt}}m{#1}}
\begin{table}[tp]
  \begin{center}
    \caption{The data used by the proposed comprehensive assessment benchmark to evaluate classifiers trained on different training data-sets.}
    \label{tab-test_data}
    \resizebox{\ifdim\width>\linewidth\linewidth\else\width\fi}{!}{%
      \setlength{\tabcolsep}{3.5pt}{

        \begin{tabular}{l|l|l|ll|lll|llll} \hline
          Training & \multicolumn{11}{c}{Test Data} \\ 
          Data-set & Clean & Corrupt & \multicolumn{2}{c|}{Adversarial} & \multicolumn{3}{c|}{Novel} & \multicolumn{4}{c}{Unrecognisable}\\ 
    \hline
          MNIST & MNIST test-set & MNIST-C              & AA$_{l_\infty}\epsilon=0.3$ & AA$_{l_2}\epsilon=2$             & Omniglot & FMNIST & KMNIST & \multirow{7}{*}{\rotatebox[origin=c]{90}{Phase}} & \multirow{7}{*}{\rotatebox[origin=c]{90}{Scramble}} & \multirow{7}{*}{\rotatebox[origin=c]{90}{Blobs}} & \multirow{7}{*}{\rotatebox[origin=c]{90}{Uniform}}\\
          FMNIST & FMNIST test-set & FMNIST-C           & AA$_{l_\infty}\epsilon=0.3$ & AA$_{l_2}\epsilon=2$             & Textures & MNIST & CIFAR10 \\
          SVHN & SVHN test-set & MNIST--C               & AA$_{l_\infty}\epsilon=\frac{8}{255}$ & AA$_{l_2}\epsilon=0.5$ & Textures & Icons50 & CIFAR10 \\
          CIFAR10 & CIFAR10  test-set & CIFAR10-C       & AA$_{l_\infty}\epsilon=\frac{8}{255}$ & AA$_{l_2}\epsilon=0.5$ & Textures & SVHN & CIFAR100 \\
          CIFAR100 & CIFAR100 test-set & CIFAR100-C     & AA$_{l_\infty}\epsilon=\frac{8}{255}$ & AA$_{l_2}\epsilon=0.5$ & Textures & SVHN & CIFAR10  \\
          TinyIN & TinyIN val.-set & TinyIN-C           & AA$_{l_\infty}\epsilon=\frac{8}{255}$ & AA$_{l_2}\epsilon=0.5$ & Textures & iNaturalist & ImageNetO \\
          ImageNet1k & ImageNet1k val.-set & ImageNet-C & AA$_{l_\infty}\epsilon=\frac{4}{255}$ & AA$_{l_2}\epsilon=0.5$ & Textures & iNaturalist & ImageNetO \\
          \hline
        \end{tabular}
      }
    }
  \end{center}
\end{table}

The benchmark is defined to assess the performance of classifiers trained on seven popular image classification data-sets: MNIST \citep{LeCun_etal98}, FashionMNIST \citep[FMNIST;][]{Xiao_etal17}, SVHN \citep{Netzer_etal11}, CIFAR10 \citep{Krizhevsky09}, CIFAR100 \citep{Krizhevsky09}, TinyImageNet (TinyIN), and ImageNet1k 
\citep[a. k. a. ILSVRC2012;][]{Russakovsky_etal15}.
Performance is evaluated using test-data from each of the five categories illustrated in \cref{fig-DL_failures}.
The specific data-sets used in each category is summarised in \cref{tab-test_data} and described in more detail below.
\begin{description}
\item[Clean.] The standard test-set provided with each data-set.

\item[Corrupt.] Common corruptions data-sets: MNIST-C, FashionMNIST-C (FMNIST-C), CIFAR10-C, CIFAR100-C, Tiny\-ImageNet-C and ImageNet-C \citep{HendrycksDietterich19,MuGilmer19,WeissTonella22}. These data-sets modify the standard test-set using multiple corruption methods including  
  different types of noise, blurring, geometric transformations, superimposed patterns, synthetic weather conditions, and digital corruptions.
  As typical in the literature, performance is evaluated using all the corruptions at all degrees of intensity.
When the test-set resolution differs from that of the training set, the test images is resized (using bi-linear interpolation) to match the image resolution used for training. Similarly, the number of colour channels in the test images is modified, if necessary, to match the training data (\eg grayscale images are converted to colour by making the three colour channels all equal to the original image intensity values).

\item[Adversarial.] Samples are produced using AutoAttack \citep[AA;][]{CroceHein20}, a state-of-the-art ensemble attack method that employs both gradient-based (white-box) and gradient-free (black-box) attacks. AA is implemented using the torchattacks PyTorch library \citep{Kim21torchattacks}.
  Two sets of adversarial samples are used. Each set is created by perturbing the entire standard (clean) test-set\footnote{Do to the extremely high computational demands of performing AA on large ImageNet1k trained networks, only 1000 adversarial samples were used for these networks.}, but with a different method of constraining the magnitude of the perturbation. Specifically, AA is used to apply both $l_\infty$ and $l_2$-norm constrained attacks.
  The perturbation budget ($\epsilon$) used for each attack is the standard value used in the previous literature for each data-set, and is defined in \cref{tab-test_data}.
  
\item[Novel Class.] Three data-sets containing objects of unknown classes are used for each training data-set. Specifically,
  MNIST trained networks are tested using unknown, novel, classes from the test-sets of the Omniglot \citep{Lake_etal15}, FashionMNIST \citep{Xiao_etal17} and KMNIST \citep{Clanuwat_etal18} data-sets.
FashionMNIST trained networks are evaluated with the test-sets of the Textures \citep{Cimpoi_etal14}, MNIST and CIFAR10 data-sets. When trained with SVHN, unknown class rejection is evaluated with the Textures, Icons-50 \citep{HendrycksDietterich19}, and CIFAR10 test data-sets.
  The CIFAR10 and CIFAR100 trained networks are tested using novel classes from the test-sets of the Textures and SVHN data-sets and the CIFAR data-set that was not used for training (\ie a CIFAR10 trained network is evaluated on its ability to reject samples from CIFAR100, and vice versa).
  For TinyImageNet and ImageNet1k trained networks novel class data comes from the Textures test set, the iNaturalist 2021 validation set \citep{VanHorn_etal18}, and the ImageNet-O data-set \citep{Hendrycks_etal19c}.
When the test-set resolution differs from that of the training set, the test images is resized (using bi-linear interpolation) to match the image resolution used for training. Similarly, the number of colour channels in the test images is modified, if necessary, to match the training data (\eg colour images are converted to grayscale).

\item[Unrecognisable.] Data to test unrecognisable image rejection is the same for all training data-sets, and consisted of synthetic images randomly generated using four different methods.
  First, the images of the standard (clean) test set after randomising the phase, in the Fourier domain, of each image (``Phase'').
  Second, the images of the clean test set after a random permutation of all pixels (``Scramble''). 
  Third, as used in \citet{Hendrycks_etal19}, images containing random blobs (``Blobs'').
  Fourth, images in which each pixel intensity value is independently and randomly selected from a uniform distribution (``Uniform''). 
  Each of these four data-sets contain the same number of samples as the standard test-set associated with the training data.
\end{description}

In principal, the amount of test data that could be used is unlimited (as there are countless images containing objects that any one classifier has not been trained to recognise, and due to the stochastic nature of the methods used to create corrupt, adversarial, and unrecognisable images). Hence, the above selection of test data necessarily represents a small, arbitrary, sub-set of all the possible data that could be used. In making this selection the aim was to define a reasonably wide variety of data while ensuring that evaluation could be performed in a reasonable time using reasonable resources. The results in \cref{sec-results} suggest that the chosen data is sufficiently challenging that modern Deep Neural Networks (DNNs) are far from ceiling performance. However, as classification methods improve in the future, the data used for evaluation could be extended to allow a more thorough and challenging assessment in two ways: adding more data in each category, or by adding categories.

Expanding the range of data in each of the existing categories could be done by:
using new sets of clean data, such as CIFAR10.1 \citep{Recht_etal18}, or ImageNet-A \citep{Hendrycks_etal21b};
using  different corruptions (\eg different forms of noise or image transformation), higher magnitude corruptions (\eg a wider range of scales such as used in the MNIST Large Scale data set \citep{JanssonLindeberg20}),
   the different types of corruption used in the OOD-CV-v2 benchmark \citep{Zhao_etal23},
or objects shown against different backgrounds (as is used in the NICO challenge \citep{Zhang_etal23}); using any of numerous alternative algorithms for generating adversarial images \citep{Liang_etal22,Ren_etal20,Xu_etal20}, or different methods of constraining the perturbations, or higher perturbation budgets; using the additional novel class data-sets proposed in the OpenOOD benchmark \citep{Yang_etal22}; using fooling images \citep{Kumano_etal22,Nguyen_etal15} that are unrecognisable images which have been optimised to be difficult to reject. Adversarial attack-type techniques could also be used not only to perturb clean test samples, as is currently the case,  but samples from other data types to make these types of data more challenging to correctly process \citep{Li_etal24}.
New categories of test data that could be used to extend and increase the challenge of the proposed benchmark include: known classes presented in the form of 
drawings, art, or cartoons, such as in ImageNet-R \citep{HendrycksDietterich19} or PACS \citep{Li_etal17}; 
or images where texture has been replaced by that from another image \citep{Geirhos_etal18b}.

\subsection{Rejection Criteria}
\label{sec-rejection}

In order for classifiers to be evaluated using the proposed method, it is necessary to define a criteria that will be used to determine if a sample is accepted for classification, or rejected. Many different methods for unknown class rejection have been proposed \citep{Vaze_etal22,Pang_etal22,Chen_etal22,HendrycksGimpel17,Tajwar_etal21}. The proposed benchmark  uses maximum softmax probability \citep[MSP;][]{HendrycksGimpel17}, which defines the confidence that a sample is of a known class as the maximum response of the network output after application of the softmax activation function.
MSP is used as this is a standard, baseline method, used frequently in the literature on unknown class rejection. Furthermore, a comparison of different rejection methods using the same metric on which the proposed benchmark is based found MSP to be one of the best methods \citep{Zhu_etal22}.

Previous work has shown that relative performance of different unknown class rejection methods vary depending on a number of factors \citep{Szyc_etal23} including the data-sets \citep{Tajwar_etal21} and models \citep{Zhu_etal22} used. Hence, to confirm that MSP is an appropriate choice, results for three other post-hoc methods of unknown class rejection (those that rely on the responses produced by the network) are also reported in \cref{sec-results_validate_benchmark_rejection}:
 (1) Maximum logit score \citep[MLS;][]{Vaze_etal22}, which defines the confidence that a sample is of a known class as the maximum response of the network output before any activation function is applied. This was shown to be the best method of unknown class rejection in a recent comparison of state-of-the-art methods \citep{Vojir_etal23}. (2) Energy score \citep{Liu_etal20}, which defines confidence as the negative logarithm of the denominator of the softmax activation function applied the network output layer. (3) GEN \citep{Liu_etal23} a recently proposed method that claims to be superior to all three other methods considered here. Note, only post-hoc methods are considered, as (A) they can be applied to the same network, allowing for a fair comparison and avoiding the need to re-train the network to assess different methods, (B) because post-hoc methods have performance comparable to, or better than, other methods \citep{Yang_etal22}.

Consistent with the method proposed by \citet{Zhu_etal22}, a rejection threshold is set such that a fixed percentage correctly classified clean examples are accepted. In other words, such that the confidence associated with those samples, is greater than or equal to the threshold. All samples, from any data-set, with confidence lower than the threshold are rejected, while all others are accepted and the class label predicted. \citet{Zhu_etal22} propose setting the threshold such that either 95\% or 99\% of correctly classified clean examples are accepted. For the proposed benchmark 95\% is used, but results for the alternative are also reported in \cref{sec-results_validate_benchmark_threshold}. Instead of using a fixed threshold, it would also be possible to calculate performance over a range of thresholds, as with the AUROC and AUPR metrics. However, in a practical application it would be necessary to define a fixed threshold, so the proposed approach is more likely to accurately reflect real-world performance. Furthermore, the creator of the classifier is likely to want it to perform well on the data it was trained on, so will want to accept a high proportion of the  correctly classified clean samples.

\section{Results}
\label{sec-results}

\cref{sec-results_validate_benchmark} reports results of experiments designed to validate the proposed benchmark and justify some of the choices that have been made in defining it (\cref{sec-results_validate_benchmark_rejection,sec-results_validate_benchmark_threshold}). \cref{sec-results_robustness} reports results designed to determine how well existing DNNs perform when evaluated using the more comprehensive assessment methods being advocated. 
The proposed benchmark can be used to assess the performance of any classifier trained on any of the supported data-sets. Obviously, it  would be impossible in this one publication to attempt to benchmark the performance of every such classifier or to evaluate the huge variety of different techniques that can be used with such classifiers to improve adversarial robustness, generalisation performance, or unknown classes rejection. Hence, results are only reported for a few, select, models.

The first set of experimental results (\cref{sec-results_validate_benchmark}) were produced using self-trained ResNet18 models \citep{He_etal16}. These models were  trained on CIFAR10 and CIFAR100 using a variety of training-time data augmentations that have previously been proposed to improve robustness or generalisation. Augmentation methods that produce different levels of robustness to different test data types were chosen, 
so as to be able to ensure that this variation in performance is captured by the proposed benchmark.
These networks have been trained using standard procedures full details of which are given in \cref{app-methods}.

The second set of results (\cref{sec-results_robustness}) focus of benchmarking the performance of state-of-the-art (SOTA) robust models. These models were taken from the RobustBench leaderboards. 
RobustBench \citep{Croce_etal21} is an existing benchmark for assessing the robustness of DNNs. Unlike the proposed benchmark, RobustBench evaluates performance of a model using only a single type of data: either adversarial (produced using either $l_\infty$ or $l_2$ constrained AA) or common corruptions. 
From each of these RobustBench leaderboards a selection of the top performing models were evaluated using the proposed benchmark. The selected models included the highest ranked model from each leaderboard for which pre-trained weights were available from the RobustBench repository\footnote{\url{https://github.com/RobustBench/robustbench}} on November 15$^{th}$ 2024. Other tested models were selected from the top-10 with a bias towards the highest ranked models, but lower ranked models were sometimes selected to increase the diversity  of network architectures and robustification methods evaluated.

\subsection{Experiments to Validate the Benchmark}
\label{sec-results_validate_benchmark}

This paper advocates for (1) more comprehensive evaluation of DNN performance (\cref{sec-practice}), and (2) using a single, consistent, metric to assess performance on all types of testing data (\cref{sec-metrics}).
\cref{tab-ResNet18_CIFAR10_MSP_traditional,tab-ResNet18_CIFAR10_MSP_traditional,tab-ResNet18_CIFAR10_MSP_traditional,tab-ResNet18_CIFAR100_MSP_traditional} shows performance for many different test data sets, measured using existing metrics. Hence, these results address aim (1) but not aim (2). As discussed in \cref{sec-practice} such comprehensive assessment is rarely performed. However, it provides valuable insights into both the strengths and weaknesses of different classifiers. For example, it can be seen that
the networks trained using adversarial training (\AT) with \PGD are most resistant to  adversarial attacks with AA, but that such networks have reduced performance on unknown class rejection (and are particularly vulnerable to making incorrect classifications when presented with images containing scrambled pixels). The former is expected, but the latter is ignored in the adversarial robustness literature as evaluations with such data are never performed. Similarly, it is to be expected given previous research that 
\pixmix produces excellent performance on unknown class rejection, however, the vulnerability of such networks to adversarial attack has been mostly undocumented.

\begin{table}[tp]
  \begin{center}
    \caption{Comparison of results produced using (a) exiting, standard, metrics, and (b) the proposed benchmark, when used to evaluate  ResNet18 networks trained on CIFAR10 using a variety of different training-time data augmentation methods (see \cref{app-augmentations} for details). 
      (c) Results equivalent to those shown in (b) 
      but for different rejection methods. (d) Results equivalent to those shown in (b) 
      but for a different rejection threshold.
      The best results for each column in each block are highlighted in bold.}
    \label{tab-ResNet18_CIFAR10_MSP}
        \subfigure[Existing Metrics]{
      \label{tab-ResNet18_CIFAR10_MSP_traditional}
      \begin{tabular}{l|*{4}{S[table-format=2.2,table-column-width=\eqcolwidthA]}|*{7}{S[table-format=2.2,table-column-width=\eqcolwidthB]}} \hline
        Training & {Clean} & {Corrupt} & {AA$_{l_\infty}$} & {AA$_{l_2}$} & {Textures} & {SVHN} & {CIFAR100} & {Phase} & {Scramble} & {Blobs} & {Uniform}\\
        Method & \multicolumn{4}{c|}{Accuracy (\%)} & \multicolumn{7}{c}{AUROC (\%) when MSP used as prediction confidence score} \\ 
        \hline
        \none        &  94.46& 	74.59   &	0.00   &  0.02	  &91.60 &	95.99 &	89.05 &	95.54 &	92.64 &	94.86 &	87.57\\
        \noise       &  88.73&	82.29   &	4.10   &  38.01	  &88.45 &	89.67 &	83.90 &	88.67 &	75.51 &	91.25 &	98.86\\
        \AT          &  83.91&	75.68   &\B 48.80&\B 59.35&83.82 &	88.38 &	77.36 &	77.72 &	59.16 &	71.46 &	83.02\\
        \pixmix      &  94.10&\B 88.32 &   0.00   &  1.26	  &\B 98.89&\B 98.48&\B 91.74&\B 99.38&\B 99.82&\B 99.96&\B 99.85\\
        \regmixup    &\B94.67&   80.43 &  0.00	& 0.00 &92.05	&94.26	&88.81	&95.32	&98.33	&95.21	&99.78\\
       \hline
      \end{tabular}
    }
    \subfigure[Proposed Comprehensive Assessment Benchmark (DAR (\%) calculated using MSP as the prediction confidence score and a rejection threshold set at 95\% clean data acceptance)]{
      \label{tab-ResNet18_CIFAR10_MSP_95}
      \begin{tabular}{l|*{5}{S[table-format=2.2,table-column-width=\eqcolwidthC]}|*{1}{S[table-format=2.2(2),table-column-width=2.7cm]}} \hline
        Training Method & {Clean} &	{Corrupt}	& {Adversarial} &	{Novel}	& {Unrecog.} & {Mean $\pm$ Std}\\
        \hline
        \none        & 93.20&	80.92&	3.57 &	74.61&	62.62&	62.98	(3.30)\\
        \noise       & 89.24&	84.94&	29.60&	51.93&	56.28&	62.40	(0.30)\\
        \AT          & 85.50&	80.26&\B68.55&	49.76&	13.31&	59.48	(1.28)\\
        \pixmix      & 92.91&\B	85.36&	5.59 &\B87.98&\B99.73&\B74.31	(0.13)\\
        \regmixup    &\B 93.26&	82.16&	14.77&	69.41&	89.99&	69.92	(0.90)\\
        \hline
      \end{tabular}
      }
    \subfigure[DAR  (\%) calculated using different prediction confidence scores]{
      \label{tab-ResNet18_CIFAR10_other_95}
      \begin{tabular}{ll|*{5}{S[table-format=2.2,table-column-width=\eqcolwidthC]}|*{1}{S[table-format=2.2(2),table-column-width=2.7cm]}} \hline
        &Training Method & {Clean} &	{Corrupt}	& {Adversarial} &	{Novel}	& {Unrecog.} & {Mean $\pm$ Std}\\
        \hline
        \multirow{5}{*}{\rotatebox[origin=c]{90}{MLS}} 
        &\none        & 92.50&	79.65&	2.88 &	81.35&	77.74&	66.82	(4.33)\\
        &\noise       & 88.41&	84.01&	27.40&	56.13&	66.05&	64.40	(0.59)\\
        &\AT          & 84.36&	79.01&\B65.87&	52.56&	14.95&	59.35	(2.31)\\
        &\pixmix      & 92.02&	82.35&	4.10 &\B90.26&\B99.97&\B73.74	(0.07)\\
        &\regmixup    &\B 92.58&	80.02&	4.19 &	67.67&	93.77&	67.65	(1.83)\\
        \hline
        \multirow{5}{*}{\rotatebox[origin=c]{90}{Energy}} 
        &\none        &\B92.34&	79.23&	2.73 &	81.01&	78.25&	66.71	(4.59)\\
        &\noise       & 87.96&\B	83.29&	26.96&	52.84&	67.49&	63.71	(0.72)\\
        &\AT          & 83.33&	77.15&\B63.44&	44.40&	12.54&	56.17	(1.87)\\
        &\pixmix      & 91.79&	81.71&	3.80 &\B 90.09&\B99.98&\B73.47	(0.08)\\
        &\regmixup    & 92.09&	78.78&	3.27 &	59.88&	91.92&	65.19	(2.20)\\
        \hline
        \multirow{5}{*}{\rotatebox[origin=c]{90}{GEN}} 
        &\none        &\B92.55&	79.82&	2.93&	81.64&	77.60&	66.91	(4.33)\\
        &\noise       & 88.13&	83.63&	27.08&	54.65&	66.87&	64.07	(0.65)\\
        &\AT          & 83.49&	77.45&\B63.80&	46.67&	12.56&	56.79	(1.80)\\
        &\pixmix      & 91.93&	82.08&	3.94&\B	90.19&\B99.97&\B73.62	(0.08)\\
        &\regmixup    & 92.26&	79.33&	3.73&	64.71&	93.09&	66.62	(2.15)\\
        \hline

      \end{tabular}
    }
    \subfigure[DAR  (\%) calculated using   rejection threshold set at 99\% clean data acceptance]{
    \label{tab-ResNet18_CIFAR10_MSP_99}
      \begin{tabular}{l|*{5}{S[table-format=2.2,table-column-width=\eqcolwidthC]}|*{1}{S[table-format=2.2(2),table-column-width=2.7cm]}} \hline
        Training Method & {Clean} &	{Corrupt}	& {Adversarial} &	{Novel}	& {Unrecog.} & {Mean $\pm$ Std} \\
        \hline
        \none        & 94.66&	78.85&	 1.18&	38.97&	26.63&	48.06	(0.77)\\
        \noise       & 89.50&	84.12&	23.16&	20.08&	37.68&	50.91	(0.99)\\
        \AT          & 84.80&	77.84&\B58.29&	17.35&	2.22 &  48.10	(0.17)\\
        \pixmix      & 94.35&  \B88.97&	2.14 &\B73.33&\B97.23&\B71.20	(0.13)\\
        \regmixup    & \B94.95&	84.15&	1.92 &	32.94&	67.61&	56.31	(0.87)\\
        \hline
      \end{tabular}
      }
  \end{center}
\end{table}

\begin{table}[tp]
  \begin{center}
    \caption{Results equivalent to those shown in \cref{tab-ResNet18_CIFAR10_MSP} but for networks trained on CIFAR100.}
    \label{tab-ResNet18_CIFAR100_MSP}
    \subfigure[Existing Metrics]{
      \label{tab-ResNet18_CIFAR100_MSP_traditional}
      \begin{tabular}{l|*{4}{S[table-format=2.2,table-column-width=\eqcolwidthA]}|*{7}{S[table-format=2.2,table-column-width=\eqcolwidthB]}} \hline
        Training Method & {Clean} & {Corrupt} & {AA$_{l_\infty}$} & {AA$_{l_2}$} & {Textures} & {SVHN} & {CIFAR10} & {Phase} & {Scramble} & {Blobs} & {Uniform}\\
        & \multicolumn{4}{c|}{Accuracy (\%)} & \multicolumn{7}{c}{AUROC (\%) when MSP used as prediction confidence score} \\ 
        \hline
        \none        &\B76.66 &49.68 &    0.00	&0.04	&77.08	&79.11	&76.20	&79.00	&70.82	&82.17	&93.60\\
        \noise       &65.19 &57.31 &    2.70	&20.97	&66.63	&80.79	&70.64	&81.67	&44.70	&81.83	&94.41\\
        \AT          &57.45 &47.33 &    \B25.40	&\B34.77	&65.69	&78.69	&67.22	&62.67	&27.47	&58.69	&16.27\\  
        \pixmix      &73.70 &\B62.88 &    0.00	&0.56	&\B88.31	&\B88.02	&74.08	&\B93.42	&\B99.53	&\B99.66	&\B99.94\\
        \regmixup    & 76.27&	53.51& 0.00&	0.07& 77.11&	81.26&	77.61&	83.56&	65.63&	79.25&	73.74\\
        \hline
      \end{tabular}
    }
   \subfigure[Proposed Comprehensive Assessment Benchmark (DAR (\%) calculated using MSP as the prediction confidence score and a rejection threshold set at 95\% clean data acceptance)]{
      \label{tab-ResNet18_CIFAR100_MSP_95}
      \begin{tabular}{l|*{5}{S[table-format=2.2,table-column-width=\eqcolwidthC]}|*{1}{S[table-format=2.2(2),table-column-width=2.7cm]}} \hline
        Training Method & {Clean} &	{Corrupt}	& {Adversarial} &	{Novel}	& {Unrecog.} & {Mean $\pm$ Std}\\
        \hline
        \none        &\B81.69&	67.57&	9.23 &	42.43&	49.76&	50.14	(1.76)\\
        \noise       & 74.06&	70.90&	26.52&	46.74&	53.47&	54.34	(1.09)\\
        \AT          & 66.90&	62.45&\B48.94&	39.00&	4.31 &  44.32	(0.38)\\
        \pixmix      & 79.87&\B  75.20&	11.28&\B60.99&\B95.67&\B64.60	(1.19)\\
        \regmixup    & 81.25&	69.53&	10.09&	43.43&	28.98&	46.66	(0.97)\\
       \hline
      \end{tabular}
    }    

    \subfigure[DAR  (\%) calculated using different prediction confidence scores]{
\label{tab-ResNet18_CIFAR100_other_95}
      \begin{tabular}{ll|*{5}{S[table-format=2.2,table-column-width=\eqcolwidthC]}|*{1}{S[table-format=2.2(2),table-column-width=2.7cm]}} \hline
        &Training Method & {Clean} &	{Corrupt}	& {Adversarial} &	{Novel}	& {Unrecog.} & {Mean $\pm$ Std}\\
        \hline
        \multirow{5}{*}{\rotatebox[origin=c]{90}{MLS}} 
        &\none        &\B 80.00&	67.31&	7.59 &  55.63&	54.70&	53.05	(2.19)\\
        &\noise       & 72.74&	70.20&	24.91&	52.15&	55.03&	55.00	(0.95)\\
        &\AT          & 66.00&	62.34&\B47.55&	41.10&	3.11 &  44.02	(0.21)\\
        &\pixmix      & 78.49&	73.72&	9.96 &\B68.39&\B97.66&\B65.64	(0.86)\\
        &\regmixup    & 80.42&	68.21&	9.01 &	37.02&	24.93&	43.92	(0.34)\\
        \hline
        \multirow{5}{*}{\rotatebox[origin=c]{90}{Energy}} 
        &\none        &\B 79.33&	66.16&	6.96 &	56.80&	53.34&	52.52	(1.95)\\
        &\noise       & 71.45&	68.72&	23.50&	50.58&	51.98&	53.25	(1.16)\\
        &\AT          & 63.50&	60.24&\B43.70&	40.32&	2.81 &  42.11	(0.32)\\
        &\pixmix      & 77.24&	72.12&	8.73 &\B68.28&\B97.87&\B64.85	(0.78)\\
        &\regmixup    & 78.56&	64.23&	6.96 &	23.60&	19.32&	38.54	(2.12)\\
        \hline
        \multirow{5}{*}{\rotatebox[origin=c]{90}{GEN}} 
        &\none        &\B79.47&	66.45&	7.09&	56.59&	53.52&	52.63	(2.14)\\
        &\noise       & 71.60&	68.96&	23.65&	50.88&	52.38&	53.49	(1.05)\\
        &\AT          & 63.66&	60.35&\B43.94&	40.24&	2.73&	42.18	(0.36)\\
        &\pixmix      & 77.38&	72.29&	8.87&\B 68.19&\B97.77&\B64.90	(0.87)\\
        &\regmixup    & 78.76&	64.80&	7.19&	25.80&	19.92&	39.30	(1.76)\\
        \hline
\end{tabular}
}

      \subfigure[DAR  (\%) calculated using   rejection threshold set at 99\% clean data acceptance]{
    \label{tab-ResNet18_CIFAR100_MSP_99}
      \begin{tabular}{l|*{5}{S[table-format=2.2,table-column-width=\eqcolwidthC]}|*{1}{S[table-format=2.2(2),table-column-width=2.7cm]}} \hline
        Training Method & {Clean} &	{Corrupt}	& {Adversarial} &	{Novel}	& {Unrecog.} & {Mean $\pm$ Std}\\
        \hline
        \none        &\B 78.63&	57.34&	2.90 &	18.47&	19.52&	35.37	(2.73)\\
        \noise       & 68.09&	62.46&	15.98&	17.83&	18.29&	36.53	(1.45)\\
        \AT          & 60.88&	52.98&\B36.14&	14.89&	0.57 &  33.09	(0.49)\\
        \pixmix      & 75.91&\B	69.01&	3.93 &\B32.09&\B88.17&\B53.82	(1.65)\\
        \regmixup    & 78.51&	59.84&	3.42 &	14.57&	6.76 &	32.62	(0.28)\\
        \hline
      \end{tabular}
      }
  \end{center}
\end{table}

\cref{tab-ResNet18_CIFAR10_MSP_95,tab-ResNet18_CIFAR100_MSP_95} show the assessment of the same networks produced using the proposed benchmark. Hence, these results address both concerns (1) and (2) described in \cref{sec-practice,sec-metrics}. The new evaluation metric successfully captures the same pattern of relative performance that can be seen when using the standard metrics. For example, \AT produces the highest DAR for adversarial samples, but this improvement comes at the cost of reduced performance (compared to the baseline) in clean accuracy and novel and unrecognisable image rejection. Of the tested augmentation methods, \pixmix produces the highest DAR on corrupt data and the two types of unknown class data, which is consistent with the results produced using the standard metrics. \regmixup produces the highest performance on clean test data, a result which is captured by both the proposed and standard metrics.

The proposed metrics differ from the standard ones in some respects. For example, 
for all the tested data augmentation techniques, performance  on unknown class data is far worse when measured using DAR than the standard metrics would imply. This is because the same rejection threshold has been used for all data types, rather than being tuned to optimise performance on unknown class rejection. This illustrates that current metrics for unknown class rejection grossly over-estimate likely real-world performance (as discussed in \cref{sec-metrics}). In contrast, the performance of all the tested models on known class data (clean, corrupt and adversarial) is generally higher than would be expected from the standard metrics. This is because networks can reject samples for which prediction confidence is low which include many samples for which the predicted class is wrong.

An advantage of the proposed benchmark is the use of consistent units to measure performance on each type of test data. This facilitates the comparison of performance of different data types. Using a consistent performance metric for each test data type also allows an average to be calculated. In contrast, it would be meaningless to average the standard metrics given that they are reported in different units.
The average performance over all data types (shown in the last columns of \cref{tab-ResNet18_CIFAR10_MSP_95,tab-ResNet18_CIFAR100_MSP_95}) suggests that, of the tested data augmentation methods, \pixmix produces the best performance overall. 
While the average performance helps to identify the most promising methods of improving robustness, the individual results for each test data type help identify remaining vulnerabilities that it would most useful to address in future research.

\subsubsection{Choice of Unknown Class Rejection Method}
\label{sec-results_validate_benchmark_rejection}
\label{sec-ResNet18_CIFAR10_other}

The proposed benchmark uses MSP as the confidence score used to decided if samples are accepted or rejected.
The same networks evaluated in the previous section were re-evaluated using alternative rejection methods: MLS, Energy score and GEN (see \cref{sec-rejection}). The results for the networks trained on CIFAR10 are shown in \cref{tab-ResNet18_CIFAR10_other_95}, and for CIFAR100 in \cref{tab-ResNet18_CIFAR100_other_95}. Comparing the results in \cref{tab-ResNet18_CIFAR10_other_95} with the results obtained using MSP shown in \cref{tab-ResNet18_CIFAR10_MSP_95} it can be seen that while individual accuracy values can vary, a similar overall performance is produced using each method of rejection, and as a result a similar ranking of the overall performance of each training method is obtained. The same can be said for the CIFAR100 results shown in \cref{tab-ResNet18_CIFAR100_other_95} and \cref{tab-ResNet18_CIFAR100_MSP_95}. The benchmark's results, therefore, do not seem to be specific to the particular method of rejection used. Furthermore, the choice of MSP can be justified as the worse performance on any data type is better for MSP than for any of the other three methods, and hence, MSP seems to have a slight advantage in improving robustness.

\subsubsection{Choice of Unknown Class Rejection Threshold}
\label{sec-results_validate_benchmark_threshold}

The proposed benchmark sets the rejection threshold such that 95\% of correctly classified clean examples are accepted (see \cref{sec-rejection}).
\cref{tab-ResNet18_CIFAR10_MSP_99,tab-ResNet18_CIFAR100_MSP_99} shows results for the same networks tested earlier but using a threshold set so that 99\% of correctly classified clean examples are accepted.
The results produced with both thresholds are consistent showing similar trade-offs in performance on the different test data types, and producing similar rankings of different training methods. The benchmark's results, therefore, do not seem to be specific to the particular rejection threshold used.

In general setting the threshold so that  95\% of correctly classified clean examples are accepted produces  more favourable evaluation of each classifier. This is to be expected, as lowering the percentage of correctly classified clean examples that are accepted,
increases the rejection threshold and allows more samples associated with low confidence predictions to be rejected. Such low confidence prediction are generally more likely to be wrong, so rejecting them is beneficial.

\subsection{Experiments to Evaluate Model Robustness}
\label{sec-results_robustness}
\label{sec-results_robustness_robustbench}
\label{sec-robustbench}

The results in this section aim to determine how well existing DNNs perform when evaluated using the more comprehensive assessment methods advocated in this paper. 
Classifiers with SOTA robustness, as determined by RobustBench \citep{Croce_etal21}, are evaluated and the results are shown in \cref{tab-RobustBench}.
The comprehensive assessment provides insights into these models that are not unavailable with the narrower analysis provided by RobustBench. For example, 
the CIFAR10 and CIFAR100 models that have been trained to produce good performance on the common corruptions data have surprisingly good performance at unknown class rejection (\cref{tab-RobustBench_CIFAR10_MSP_95,tab-RobustBench_CIFAR100_MSP_95}, upper rows). Hence, despite their relative susceptibility to adversarial attack, their overall performance is competitive with CIFAR10/100 models that have been trained for adversarial robustness.

The CIFAR10 and CIFAR100 models ranked highly  by RobustBench for  robustness to $l_\infty$ constrained adversarial attacks are generally very poor at rejecting unrecognisable images (\cref{tab-RobustBench_CIFAR10_MSP_95,tab-RobustBench_CIFAR100_MSP_95}, lower rows). As a specific example, RobustBench's top performing model on CIFAR100 $l_\infty$ adversarial robustness \citep{Wang_etal23} produces the wrong predictions around 94\% of the time on such images.
Hence, these models lack the robustness that conventional methods of evaluation suggest they have. On the contrary, the proposed comprehensive evaluation shows that SOTA methods of improving performance against adversarial attack fail to provide security as an attacker can use cheaply and easily generated random images to fool the classifier with a high chance of success.

The CIFAR10 models developed to have robustness to $l_2$ constrained adversarial attacks, compared to their counterparts trained for $l_\infty$ robustness, are much better at dealing correctly with samples from unknown classes  (\cref{tab-RobustBench_CIFAR10_MSP_95}, middle rows). This, perhaps, suggests that diversifying the methods used to produce training-time adversarial images might help reduce this particular vulnerability. 
Another approach to improving performance is also suggested by the performance of the XCiT-L12 model \citep{Debenedetti_etal23} (\cref{tab-RobustBench_CIFAR100_MSP_95}, last row).
This is the only transformer architecture to make it into the top-10 of any of the CIFAR10/100 RobustBench leaderboards (at ninth place for CIFAR100 $l_\infty$ adversarial robustness). 
This model has far better performance at rejecting unrecognisable images, suggesting that architectural changes may be helpful. However, this improvement in robustness to unrecognisable images comes at the cost of increased vulnerability to novel images. As a result, an attacker could present this model with random images of objects with classes that differ from those in the training data-set and the model would fail 66\% of the time. Overall, the proposed comprehensive assessment benchmarks ranks this model as the most robust of the CIFAR100 models tested: a very different ranking than that produced by the narrow testing performed in RobustBench.

It is remarkable how little improvement in overall, average, performance is shown by  
 the SOTA CIFAR10 and CIFAR100 models (\cref{tab-RobustBench_CIFAR10_MSP_95,tab-RobustBench_CIFAR100_MSP_95}) compared with the corresponding results in \cref{tab-ResNet18_CIFAR10_MSP_95,tab-ResNet18_CIFAR100_MSP_95} for (generally) much simpler models made robust with (generally) older methods. Indeed, the best performing  CIFAR100 RobustBench model has worse performance, on average, than the best performing model in \cref{tab-ResNet18_CIFAR100_MSP_95}. This is due to the SOTA models failing to make progress on improving performance on all the different types of test data. For example, if we compare the models trained to be adversarially robust (bottom three rows of \cref{tab-RobustBench_CIFAR100_MSP_95} and the row labelled \AT in \cref{tab-ResNet18_CIFAR100_MSP_95}) it can be seen that the RobustBench models have substantially improved performance on adversarial and clean images, but (as noted above) they remain poor at correctly rejecting samples from unknown classes. 
Massive amounts of time and resources have gone into developing techniques to enhance adversarial robustness beyond that produced by \AT with \PGD \citep{Madry_etal18}. The current results suggest that much of this effort may have been wasted, as these models remain vulnerable 
 as an attacker can easily and effectively induce a classifier to generate the wrong predictions by presenting images from unknown classes.

The performance of ImageNet1k trained models (\cref{tab-RobustBench_ImageNet_MSP_95}) compared with that of the CIFAR10 models (\cref{tab-RobustBench_CIFAR10_MSP_95}) suggests that 
the benefits of an increased training data-set size is off-set by the increase in task difficulty, so that the performance of the ImageNet1k models is no better than that of the CIFAR10 ones. 
The ImageNet1k results, and all the others that have been presented here,
  show that current methods of training deep networks produce classifiers that lack robustness. 
In the real world a model might naturally encounter all the diverse types of data that are used in the proposed comprehensive benchmark and would need to deal correctly with all of them. However, 
the best performing models would only make accurate decisions for just over 77\% of such data.
Furthermore, for each of the models evaluated here 
there is at least one type of test data on which the model performs poorly, which could be exploited by a malicious actor to reduce performance even further.

\newcolumntype{C}[1]{>{\centering\let\newline\\\arraybackslash\hspace{0pt}}m{#1}}
\begin{table}[tp]
\begin{center}
  \caption{Performance measured using the proposed comprehensive benchmark of models ranked highly on the RobustBench leaderboards. The column ``Robustness'' indicates which type of data the model has been trained to generalise to. Note that RobustBench only assesses CIFAR10 models  against $l_2$ constrained adversarial attacks. The best results for each column in each block are highlighted in bold.}
\label{tab-RobustBench}
      \setlength{\tabcolsep}{2.1pt}
      \subfigure[Models trained on CIFAR10]{
    \label{tab-RobustBench_CIFAR10_MSP_95}
      \begin{tabular}{lll|C{15mm}C{15mm}C{15mm}C{15mm}C{15mm}|C{15mm}}
        \hline
  Model & Params. &Robustness & {Clean} &	{Corrupt}	& {Adv.} &	{Novel}	& {Unrecog.} & {Mean}\\
  \hline
  WideResNet18-2 &\multirow{2}{*}{268M} & \multirow{2}{*}{corruptions}&
  \multirow{2}{*}{\B94.16}&\multirow{2}{*}{86.63}&\multirow{2}{*}{40.76}&\multirow{2}{*}{71.32}&\multirow{2}{*}{\B 68.59}&\multirow{2}{*}{72.29}\\
  \citet{Diffenderfer_etal21}&&&&&&& \\
  ResNet18 &\multirow{2}{*}{11M}              & \multirow{2}{*}{corruptions}    &
  \multirow{2}{*}{93.31}&\multirow{2}{*}{\B	89.69}&\multirow{2}{*}{19.73}&\multirow{2}{*}{56.40}&\multirow{2}{*}{54.56}&\multirow{2}{*}{62.74}\\
  \citet{Kireev_etal22}&&&&&&&\\
  WideResNet70-16 &\multirow{2}{*}{267M}        &\multirow{2}{*}{adv. ($l_2$)}    &
  \multirow{2}{*}{93.53}&\multirow{2}{*}{88.70}&\multirow{2}{*}{82.80}&\multirow{2}{*}{73.49}&\multirow{2}{*}{48.53}&\multirow{2}{*}{\B77.41}\\
            \citet{Wang_etal23}&&&&&&& \\
  WideResNet70-16 &\multirow{2}{*}{267M}      &\multirow{2}{*}{adv. ($l_2$)}    &
  \multirow{2}{*}{93.80}&\multirow{2}{*}{88.15}&\multirow{2}{*}{79.15}&\multirow{2}{*}{\B75.82}&\multirow{2}{*}{39.12}&\multirow{2}{*}{75.21}\\
            \citet{Rebuffi_etal21}&&&&&&&\\
  WideResNet94-16 &\multirow{2}{*}{366M} & \multirow{2}{*}{adv. ($l_\infty$)} &
  \multirow{2}{*}{92.04}&\multirow{2}{*}{86.72}&\multirow{2}{*}{86.36}&\multirow{2}{*}{59.26}&\multirow{2}{*}{21.21}&\multirow{2}{*}{69.12}\\
  \citet{Bartoldson_etal24}&&&&&&& \\
  MeanSparse RaWRN70-16 &\multirow{2}{*}{267M} &\multirow{2}{*}{adv. ($l_\infty$)} &
  \multirow{2}{*}{92.06}&\multirow{2}{*}{86.71}&\multirow{2}{*}{\B86.43}&\multirow{2}{*}{60.10}&\multirow{2}{*}{22.42} &\multirow{2}{*}{69.54}\\
  \citet{Amini_etal24}&&&&&&&\\
\hline
      \end{tabular}
      }

      \subfigure[Models trained on CIFAR100]{
\label{tab-RobustBench_CIFAR100_MSP_95}
      \begin{tabular}{lll|C{15mm}C{15mm}C{15mm}C{15mm}C{15mm}|C{15mm}}
          \hline
  Model & Params. &Robustness & {Clean} &	{Corrupt}	& {Adv.} &	{Novel}	& {Unrecog.} & {Mean}\\
  \hline
  WideResNet18-2 &\multirow{2}{*}{268M}&\multirow{2}{*}{corruptions}    &
  \multirow{2}{*}{\B84.72}&\multirow{2}{*}{\B80.56}&\multirow{2}{*}{20.14}&\multirow{2}{*}{50.63}&\multirow{2}{*}{50.70}&\multirow{2}{*}{57.35}\\
  \citet{Diffenderfer_etal21} &&&&&&&\\
  ResNet18 &\multirow{2}{*}{11M}&\multirow{2}{*}{corruptions}    &
  \multirow{2}{*}{81.99}&\multirow{2}{*}{78.02}&\multirow{2}{*}{12.41}&\multirow{2}{*}{\B52.05}&\multirow{2}{*}{\B72.63}&\multirow{2}{*}{59.42}\\
  \citet{Modas_etal22} &&&&&&&\\       
  ResNeXt29 32x4d &\multirow{2}{*}{6.90M}&\multirow{2}{*}{corruptions}    &
  \multirow{2}{*}{83.71}&\multirow{2}{*}{74.80}&\multirow{2}{*}{9.04}&\multirow{2}{*}{39.68}&\multirow{2}{*}{28.56}&\multirow{2}{*}{47.16}\\
  \citet{Hendrycks_etal20} &&&&&&&\\   
  WideResNet70-16 &\multirow{2}{*}{267M}&\multirow{2}{*}{ adv. ($l_\infty$)}&
  \multirow{2}{*}{80.38}&\multirow{2}{*}{72.74}&\multirow{2}{*}{\B63.99}&\multirow{2}{*}{42.86}&\multirow{2}{*}{5.94}&\multirow{2}{*}{53.18}\\
  \citet{Wang_etal23} &&&&&&&\\        
  WideResNet28-10 &\multirow{2}{*}{36M}&\multirow{2}{*}{ adv. ($l_\infty$)}&
  \multirow{2}{*}{80.12}&\multirow{2}{*}{73.62}&\multirow{2}{*}{62.32}&\multirow{2}{*}{45.58}&\multirow{2}{*}{7.96}&\multirow{2}{*}{53.92}\\
  \citet{Cui_etal23}   &&&&&&&\\       
  XCiT-L12 &\multirow{2}{*}{103M}&\multirow{2}{*}{ adv. ($l_\infty$)}&
  \multirow{2}{*}{75.95}&\multirow{2}{*}{72.25}&\multirow{2}{*}{58.85}&\multirow{2}{*}{34.18}&\multirow{2}{*}{60.11}&\multirow{2}{*}{\B60.27}\\
  \citet{Debenedetti_etal23}  &&&&&&&\\
\hline
      \end{tabular}
      }
      \subfigure[Models trained on ImageNet1k]{
      \label{tab-RobustBench_ImageNet_MSP_95}
      \begin{tabular}{lll|C{15mm}C{15mm}C{15mm}C{15mm}C{15mm}|C{15mm}}
          \hline
  Model & Params. & Robustness & {Clean} &	{Corrupt}	& {Adv.} &	{Novel}	& {Unrecog.} & {Mean}\\
  \hline
  DeiT Base     &\multirow{2}{*}{86.6M}& \multirow{2}{*}{corruptions} &
	\multirow{2}{*}{\B85.75}&	\multirow{2}{*}{74.69}&	\multirow{2}{*}{32.10}&	\multirow{2}{*}{46.50}&	\multirow{2}{*}{\B99.61}& \multirow{2}{*}{67.73}
  \\ \citet{Tian_etal22} &&&&&&&\\
    ResNet50      &\multirow{2}{*}{25.6M}& \multirow{2}{*}{corruptions} &
    \multirow{2}{*}{80.90}&	\multirow{2}{*}{74.77}&	\multirow{2}{*}{15.75}&	\multirow{2}{*}{37.83}&	\multirow{2}{*}{58.85}&	\multirow{2}{*}{53.62}
  \\\citet{Hendrycks_etal21} &&&&&&&\\
  Swin-L                &\multirow{2}{*}{197M}&\multirow{2}{*}{ adv. ($l_\infty$)}&
	\multirow{2}{*}{80.40}&\multirow{2}{*}{78.86}&\multirow{2}{*}{\B76.90}&\multirow{2}{*}{45.60}&\multirow{2}{*}{95.71}&\multirow{2}{*}{75.49}
  \\\citet{Liu_etal25}&&&&&&&\\
  ConvNeXt-L + ConvStem &\multirow{2}{*}{198M}&\multirow{2}{*}{adv. ($l_\infty$)}&
  \multirow{2}{*}{79.80}&	\multirow{2}{*}{\B79.26}&	\multirow{2}{*}{76.10}&	\multirow{2}{*}{\B46.70}&	\multirow{2}{*}{98.98}&	\multirow{2}{*}{\B76.17}
  \\\citet{Singh_etal23} &&&&&&&\\
\hline
      \end{tabular}}
      
\end{center}
\end{table}

\section{Discussion}
\label{sec-discussion}
\begin{figure}[tp]
\begin{center}
\subfigure[]{\fbox{\includegraphics[width=0.32\textwidth,clip,angle=90]{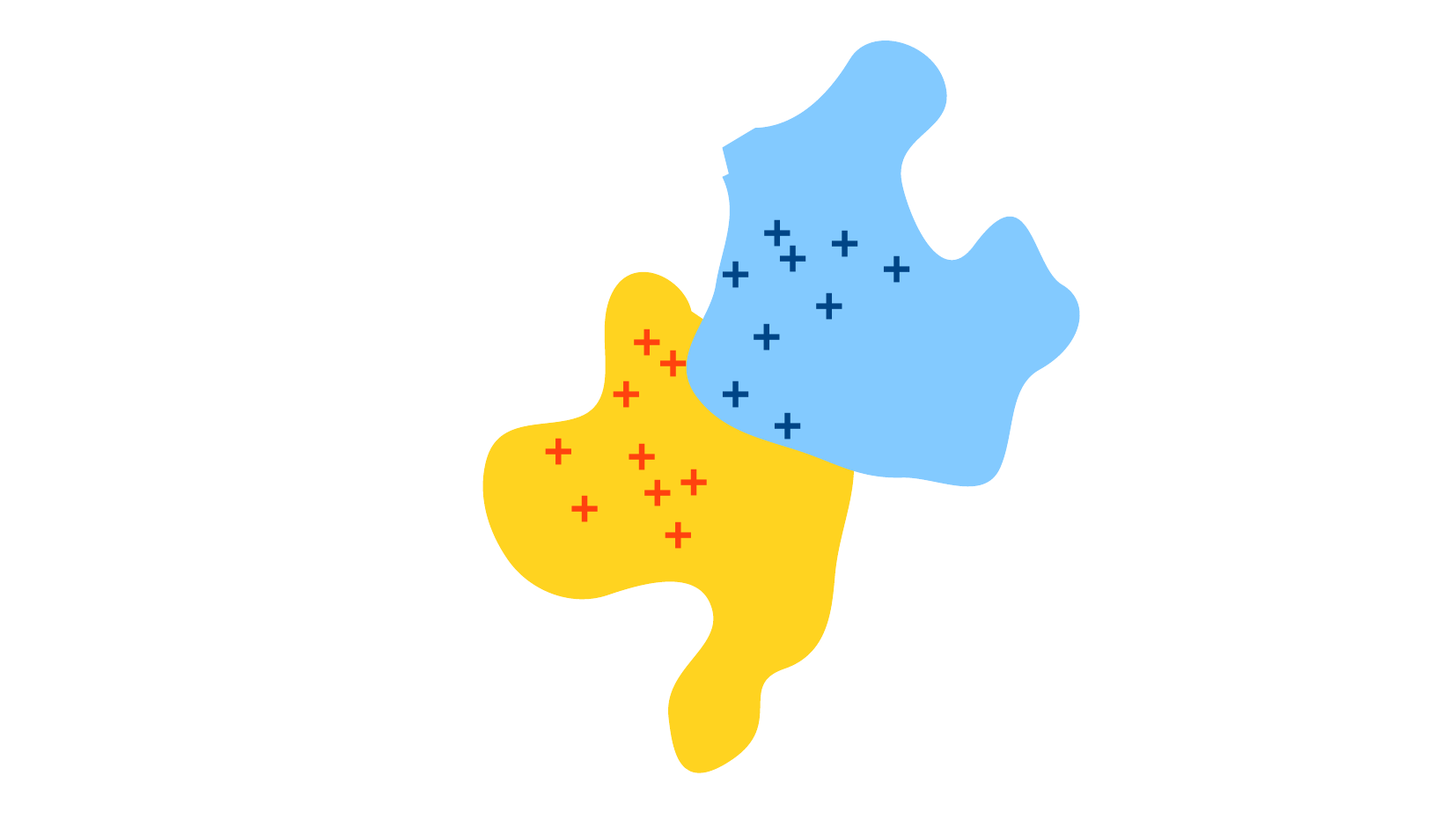}}}\hfill
\subfigure[]{\fbox{\includegraphics[width=0.32\textwidth,angle=90]{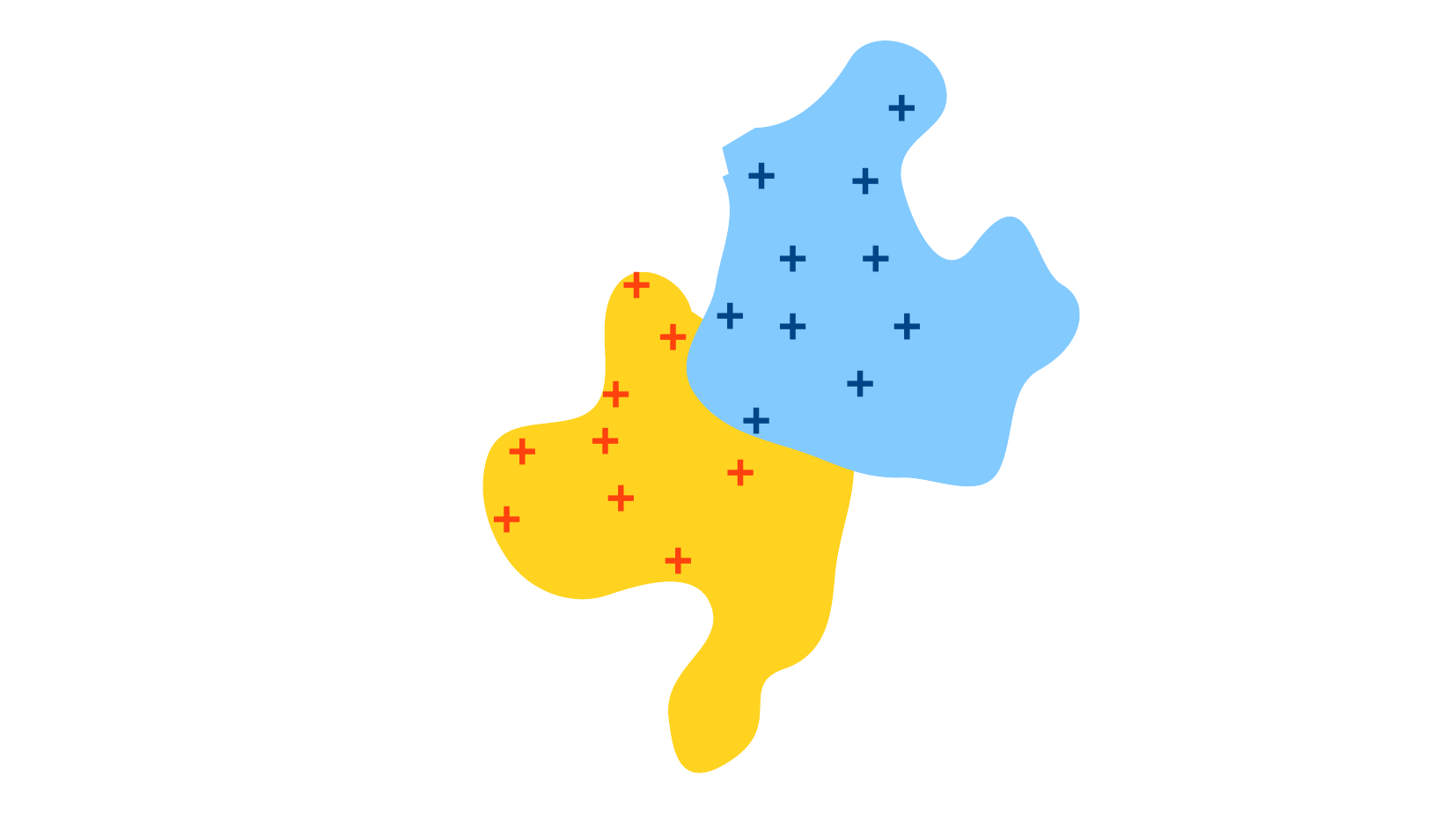}}}\hfill
\subfigure[]{\fbox{\includegraphics[width=0.32\textwidth,angle=90]{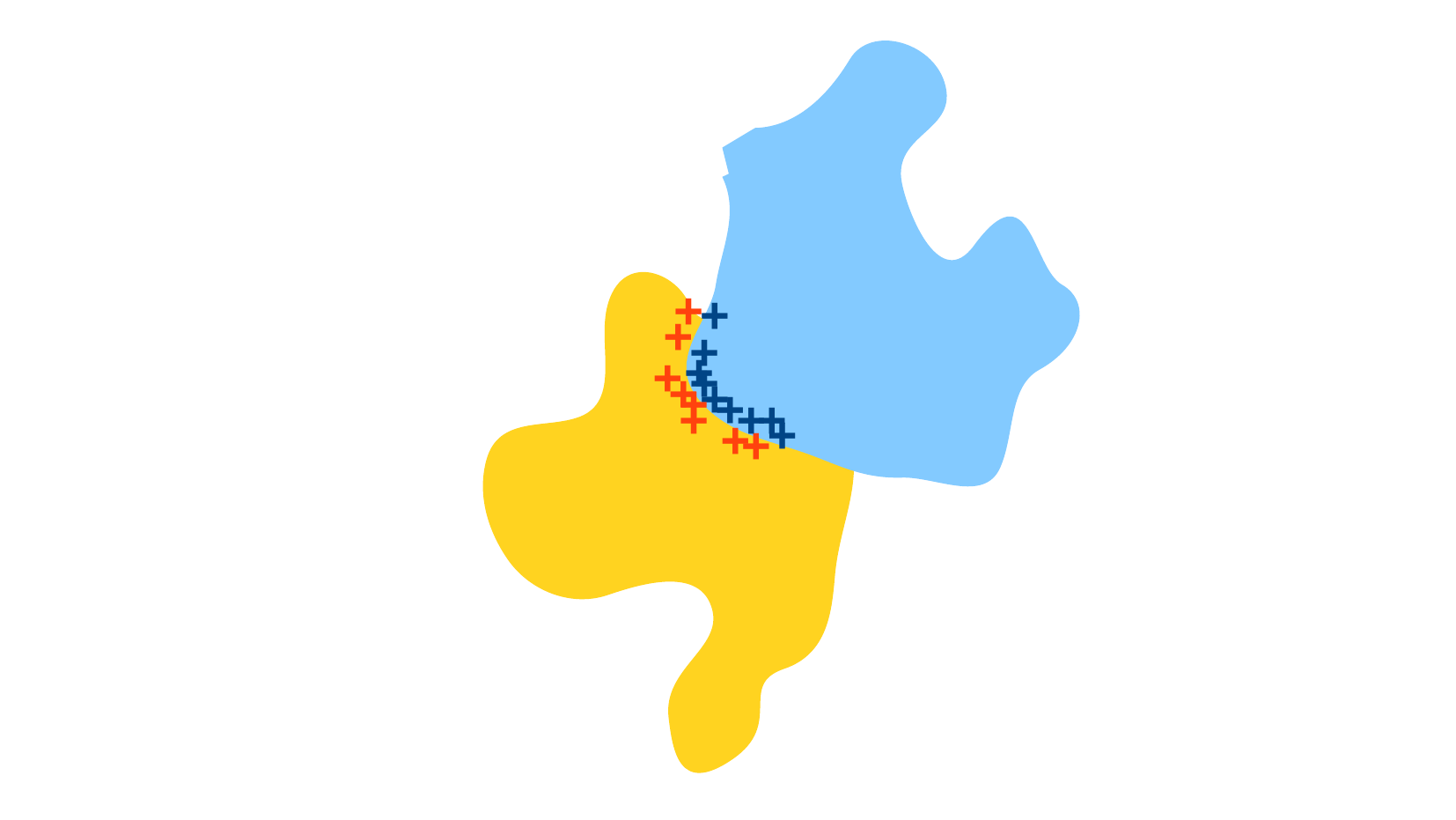}}}\hfill
\subfigure[]{\fbox{\includegraphics[width=0.32\textwidth,angle=90]{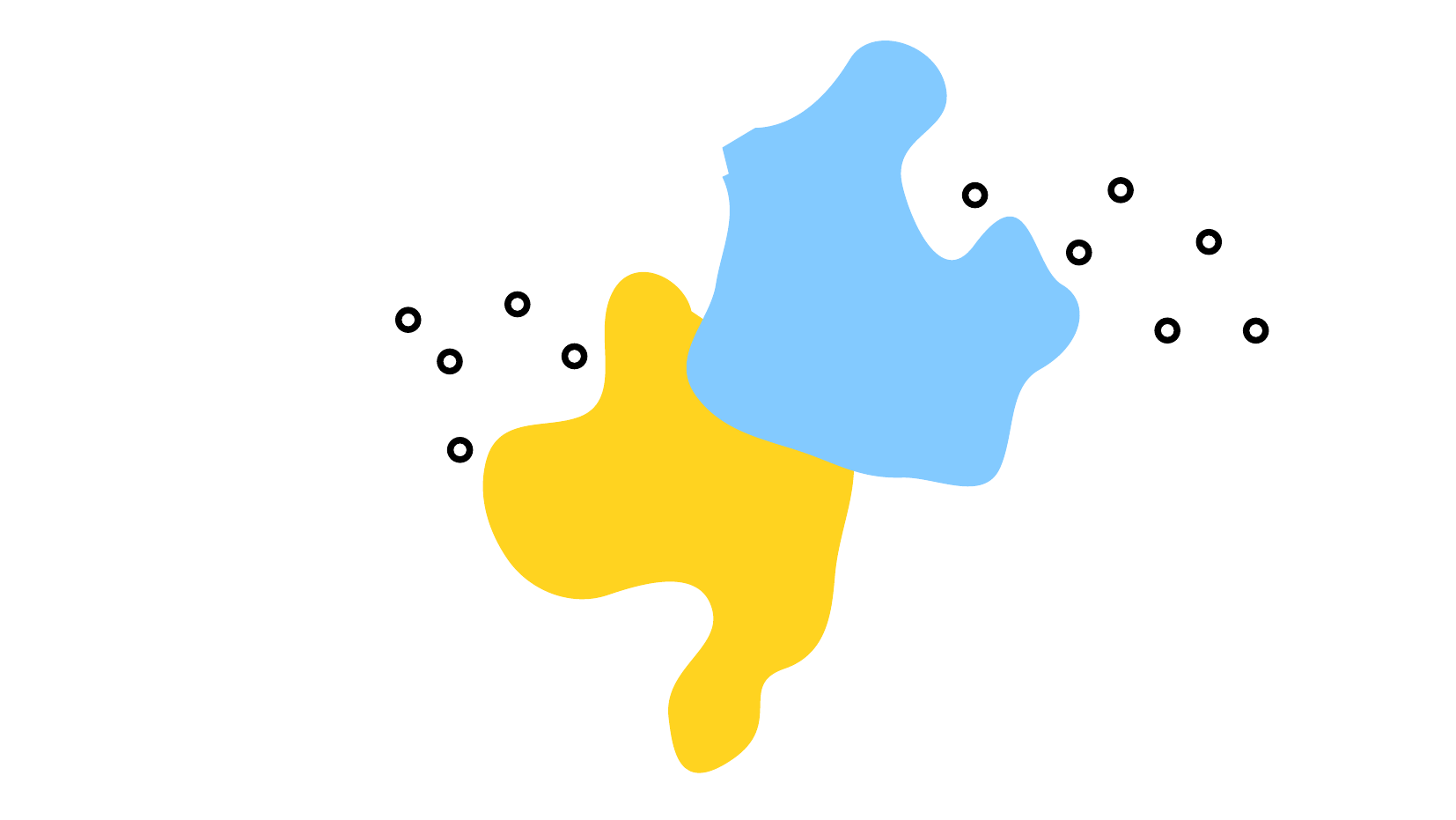}}}\hfill
\subfigure[]{\fbox{\includegraphics[width=0.32\textwidth,angle=90]{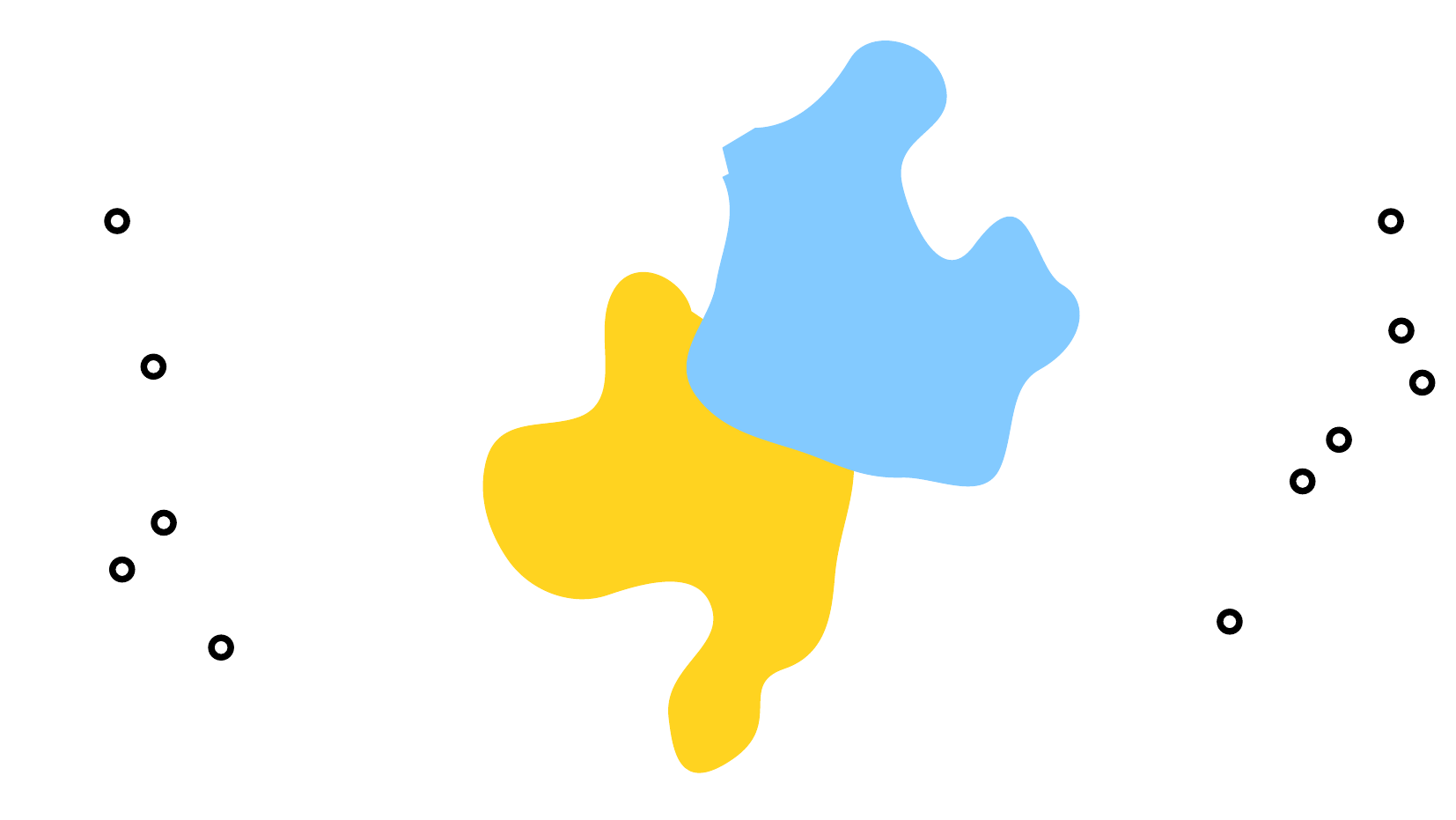}}}
\caption{An illustration of the regions of feature-space explored by different methods of testing  the accuracy of the classifier. The shaded regions in each sub-figure show  regions of a hypothetical feature-space truly occupied by two categories on which a classifier has been trained. (a) Samples from the clean test data-set probe sub-regions within each known category. (b) Samples used to assess generalisation probe larger or different regions within each known category. (c) Adversarially perturbed samples probe along the decision boundary between known classes. (d) Samples from novel classes and (e) unrecognisable samples probe regions of feature-space outside the regions occupied by the known classes.}
\label{fig-decision_boundaries}
\end{center}
\end{figure}

  The aim of testing a classifier is to ensure that the decision boundaries that it defines have been positioned correctly. However, as illustrated in \cref{fig-decision_boundaries}, 
each of the types of test data illustrated in \cref{fig-DL_failures}, on its own, is incapable of comprehensively evaluating the accuracy with which the classifier has positioned the decision boundaries for each class. For simplicity, the figure shows only two known classes, but the same observations hold true for multi-class classification problems.
The clean and corrupt test data-set (due to their finite size) can not test that the classifier responds correctly to every possible exemplar of each known class. Furthermore, they can not ensure that the classifier fails to assign class labels to samples that are outside the decision boundaries of any known class. For the example shown in the figure, it would be possible for a classifier to achieve perfect performance on the clean and corrupt data by defining a straight decision boundary roughly dividing the feature-space into two halves, and entirely ignore all other boundaries around the regions occupied by the known classes. The adversarial samples evaluate in fine detail the placement of the decision boundaries between known classes, but fails to ensure that other boundaries are defined appropriately. The unknown test data (due to its finite size) can not test that the classifier responds correctly to every possible novel or unrecognisable sample. Furthermore, this type of test data can not ensure that the classifier assigns correct class labels to samples from known classes.
This article advocates testing classifiers with all these types of data in order to make steps towards a more complete and comprehensive evaluation of a classifier's positioning of the decision boundaries, and hence, its ability to make accurate predictions on unseen data.

Despite these limitations, current practice is to test classifiers
using only one (typically the clean test data), or two (clean and adversarial, or clean and novel) types of test
data.  While others have
previously advocated using more types of test data (see
\cref{sec-practice})  they have proposed using different metrics for
different data-sets (\cref{sec-metrics}), which fails to adequately test true
performance. This article promotes an existing metric
\citep{Zhu_etal22} that is capable of being used with different types
of data to provide a consistent and comprehensive evaluation of
performance. The proposed benchmark is likely to still be imperfect, and is likely to be biased: like all assessment methods it will favour certain models over others, those that are good at processing 
the specific data-sets that has been chosen for performing the evaluation, but may be poorer at dealing with other data. However, the proposed benchmark is less biased than those that make no attempt to evaluate robustness to certain types of data, and (as discussed in \cref{sec-data}) the data sets used for assessment in the proposed benchmark can easily be extended.

Using the proposed benchmark it is found that existing methods of training DNNs, including methods that are claimed to produce state-of-the-art robustness,
fail to perform as well as previous, less comprehensive, assessments would suggest. Furthermore, it appears that there are trade-offs that mean that training methods that have good performance on one type of data (\ie improve the placement of the decision boundary in one region of feature-space) suffer from poor performance on other data (\ie place the decision boundary less accurately elsewhere in feature-space). This means that all methods of training DNNs that have been tested here are highly susceptible to malicious attacks and, by appropriate choice of data type, can be fooled into producing the wrong predictions with high probability. Improved performance might be possible by combining current methods: \eg using one model to perform unknown class rejection, and an adversarially robust model to classify the unrejected samples. However, appropriate evaluation protocols, such as the one advocated here, will still be needed to test such ensemble methods.

The lack of robustness in current DNNs is an inconvenient truth that has been, and is likely to continue to be, ignored by many as it gets in the way of claiming success on a particular task and publishing results that can be said to be state-of-the-art on the very specific data-set that has been used for evaluation. Changing this culture is likely to be only brought about by reviewers and editors insisting on more comprehensive evaluation of machine learning methods. Just as it is unacceptable to test a model on the \emph{training} data, so it should become unacceptable to test a model on a single, or a couple of, types of test data.

Without such a change in evaluation practices it will continue to be the case that the first time DNNs are rigorously assessed is after deployment in the real-world. Real-world performance will be far lower than expected, and the field will continue to be rightly criticised for making exaggerated claims. Worse leaving rigorous testing until production can have dire consequences, that the research field should feel a moral duty to at least attempt to avoid.
Without the advocated change in evaluation practices the field is likely to continue to fail to make advances in improving the security, reliability and trustfulness of machine learning models. Worse, lots of time, money, and CO$_2$ \citep{Strubell_etal20,Thompson_etal20} will continue to be expended creating methods that give a false sense of progress and security, but 
when tested thoroughly do not produce substantially improved robustness.

\paragraph{Acknowledgements}
The author gratefully acknowledges use of the King's Computational Research, Engineering and Technology Environment (CREATE).
\ifreview
\section*{Declarations}
\paragraph{Funding} No funds, grants, or other support was received to support this work.
\paragraph{Conflict of interest} The author has no competing financial interests or personal relationships that could have appeared to influence the work reported in this article.
\paragraph{Ethics approval} Not applicable.
\paragraph{Consent to participate} Not applicable.
\paragraph{Consent for publication} Not applicable.
\paragraph{Availability of data and material} All datasets used are publically available.
\paragraph{Code availability} Custom written code is available  at: \url{https://codeberg.org/mwspratling/RobustnessEvaluation}.
\paragraph{Authors' contributions} Not applicable.
\fi





\appendix
\clearpage

\section{Appendix: Methods used to Train Models}
\label{app-methods}
\subsection{Training-Time Data Augmentations}
\label{app-augmentations}
A common approach to improving performance on all the different types of test-data described in \cref{sec-types} is to use data augmentation techniques: \ie to  expand the training data-set by creating new samples that are synthetically manipulated versions of samples from the original training data-set. In the results section (\cref{sec-results}) a number of such techniques are evaluated. The selection of the particular methods to test was made so as to include standard baselines and state-of-the-art methods for producing robust models for all the different types of test-data described in \cref{sec-types}.
\begin{description}
\item[\none] Improved performance on the standard test data is often achieved by applying simple geometric data augmentations to the training data. Here 4-pixel random cropping and random horizontal flipping where applied to the CIFAR10, CIFAR100 and TinyIN training data, while no augmentations were applied to the MNIST data. These standard augmentations were used in combination with the additional augmentation methods listed below. Given that these baseline augmentation methods are very commonly used when training models, the results show how standardly trained networks cope under comprehensive evaluation. The results for this method of training also provides a baseline against which to judge the effectiveness of other methods that have been proposed to improve generalisation performance.

\item[\noise] Augmenting training images with random noise has been shown to reduce over-fitting, and hence, improve generalisation \citep{SietsmaDow91,HolmstromKoistinen92,Goodfellow_etal16}, especially to common image corruptions \citep{Rusak_etal20,Ford_etal19}. Various methods of adding noise to the training data have been proposed.
\citet{Rusak_etal20} augmented half the training data with Gaussian noise, and found that this technique produced performance on corrupt test data that was  competitive with the more complex augmentation method, AugMix, and that for the MNIST data-set specifically was just as effective as their proposed method of Adversarial Noise Training.
\citet{Lopes_etal19} advocated adding Gaussian noise to only a randomly chosen patch of the image, which they claimed  produced better robustness to image corruptions on CIFAR10 data than applying random noise to the whole image.
\citet{Ford_etal19} used an alternative method for corrupting training images with Gaussian noise in which the standard deviation of the noise applied to each image was chosen with equal probability from the range $[0,\sigma_u]$. 
The method implemented here is a generalisation for the method proposed by \citet{Ford_etal19}. Specifically, for each image the standard deviation of the noise was chosen from a uniform distribution with range $[\sigma_l,\sigma_u]$. Then each pixel intensity value was independently modified by the addition of a value chosen at random from a Gaussian distribution with mean of zero and the chosen standard deviation. The original images had pixel intensity values in the range [0,1], and after the application of the noise pixel values were clipped at 0 and 1 to remain in this range. Appropriate values of $\sigma_l$ and $\sigma_u$ were found via a grid search using small ConvNets. Using this method $\sigma_l=0$ and $\sigma_u=0.4$ were chosen for use with CIFAR10, and the same hyper-parameters were also used for CIFAR100 and TinyIN. For MNIST $\sigma_l=\sigma_u=0.5$ was used.

\item[\AT] Multi-step adversarial training with Projected Gradient Descent \citep[PGD;][]{Madry_etal18} is a standard, and highly effective, defence against adversarial attack. As a result, this method has become a standard benchmark against which all other methods of adversarial defence are judged. This method was, therefore, also selected as a data augmentation technique to be tested. It was implemented using the torchattacks PyTorch library \citep{Kim21torchattacks}.
  To perform adversarial training, clean images from the current batch of training examples were used to generate non-targeted adversarial examples, that then replaced the original training images \citep[as in][]{Madry_etal18}. Following standard protocols, the maximum allowed perturbation was constrained by the $l_\infty$-norm to be less than the perturbation budget ($\epsilon$), and the training-time adversarial images were generated using 10 steps of PGD (denoted as \PGD for brevity).
  Standard values used in the previous literature were used for $\epsilon$. Specifically, $\epsilon$ was set to $\frac{8}{255}$ for performing \AT with the CIFAR10, CIFAR100 and TinyIN training data, and $\epsilon=0.3$ was used for MNIST.

\item[\pixmix] This is a training data augmentation pipeline that repeatedly mixes (either additively or multiplicatively) an augmented image with an augmented clean image or a fractal image \citep{Hendrycks_etal22}. It is claimed that this augmentation method can improve generalisation to both clean and corrupt data, improve adversarial robustness, and improve unknown class rejection. This method of data augmentation was therefore also tested using the proposed, comprehensive, evaluation method. It was implemented using the official code released by the original authors.
\item[\regmixup] This is a variation on mixup \citep{ZhangH_etal17} that is specifically designed to improve performance on both corrupt data and on unknown class rejection \citep{Pinto_etal23}. It was re-implemented with custom code.
\end{description}

\subsection{Training Setup}
\label{app-training}


The ResNet18 \citep{He_etal16} architecture was chosen as this is very commonly used in the literature. This model has just over 11M trainable parameters. 
Training was performed using a batch size of 128 for 
110 epochs using the SGD optimiser with  weight decay of 5e-4 and an initial learning rate of 0.1. The learning rate was step-wise decayed by a factor of 10 at the start of epochs 100 and 105. 
The set-up was based on that recommended by \citet{Pang_etal21} for performing \AT on the CIFAR data-sets. Hence, these hyper-parameters may not be optimal for training with the other data augmentations used.

For each experimental condition (combination of data-set, network architecture, and training data augmentation method) the network was trained and evaluated three times, each time with a different random weight initialisation and random presentation order of training samples. The reported result for each condition is the mean accuracy over these three trials. The standard-deviation in the mean performance measured using the proposed metric is shown in the tables of results and generally shows a high degree of consistency across trials.
Publicly available checkpoints created by other researchers were only evaluated once.

Open-source code written in Python using the Pytorch library \citep{Pytorch19}, which can be used to train networks and perform all the experiments described in this article, is available for download from: \url{https://codeberg.org/mwspratling/RobustnessEvaluation}.


\begin{thebibliography}{}

\bibitem[Aboitiz et~al., 2023]{Aboitiz_etal23}
Aboitiz, F. J.~K., Legenstein, R., and Özdenizci, O. (2023).
\newblock Interaction of generalization and out-of-distribution detection
  capabilities in deep neural networks.
\newblock In {\em Proceedings of the International Conference on Artificial
  Neural Networks}, volume 14263 of {\em Lecture Notes in Computer Science},
  pages 248--59.
\newblock \doi{10.1007/978-3-031-44204-9_21}.

\bibitem[Akhtar and Mian, 2018]{AkhtarMian18}
Akhtar, N. and Mian, A. (2018).
\newblock Threat of adversarial attacks on deep learning in computer vision:
  {A} survey.
\newblock {\em IEEE Access}, 6:14410--30.
\newblock \doi{10.1109/ACCESS.2018.2807385}.

\bibitem[Amini et~al., 2024]{Amini_etal24}
Amini, S., Teymoorianfard, M., Ma, S., and Houmansadr, A. (2024).
\newblock Meansparse: Post-training robustness enhancement through
  mean-centered feature sparsification.
\newblock \arXiv{2406.05927}.

\bibitem[Amodei et~al., 2016]{Amodei_etal16}
Amodei, D., Olah, C., Steinhardt, J., Christiano, P., Schulman, J., and
  Man{\'e}, D. (2016).
\newblock Concrete problems in {AI} safety.
\newblock \arXiv{1606.06565}.

\bibitem[Averly and Chao, 2023]{AverlyChao23}
Averly, R. and Chao, W.-L. (2023).
\newblock Unified out-of-distribution detection: {A} model-specific
  perspective.
\newblock In {\em Proceedings of the International Conference on Computer
  Vision}.
\newblock \arXiv{2304.06813}.

\bibitem[Bar, 2004]{Bar04}
Bar, M. (2004).
\newblock Visual objects in context.
\newblock {\em Nature Reviews Neuroscience}, 5:617--29.

\bibitem[Bartoldson et~al., 2024]{Bartoldson_etal24}
Bartoldson, B.~R., Diffenderfer, J., Parasyris, K., and Kailkhura, B. (2024).
\newblock Adversarial robustness limits via scaling-law and human-alignment
  studies.
\newblock In {\em Proceedings of the International Conference on Machine
  Learning}.
\newblock \arXiv{2404.09349}.

\bibitem[Biggio and Roli, 2018]{BiggioRoli18}
Biggio, B. and Roli, F. (2018).
\newblock Wild patterns: Ten years after the rise of adversarial machine
  learning.
\newblock {\em Pattern Recognition}, 84:317--31.
\newblock \doi{10.1016/j.patcog.2018.07.023}.

\bibitem[Bitterwolf et~al., 2022]{Bitterwolf_etal22}
Bitterwolf, J., Meinke, A., Augustin, M., and Hein, M. (2022).
\newblock Breaking down out-of-distribution detection: Many methods based on
  {OOD} training data estimate a combination of the same core quantities.
\newblock In {\em Proceedings of the International Conference on Machine
  Learning}, volume 162 of {\em Proceedings of Machine Learning Research},
  pages 2041--74.
\newblock \arXiv{2206.09880}.

\bibitem[Bowers et~al., 2023]{Bowers_etal23}
Bowers, J.~S., Malhotra, G., Dujmovi{\'c}, M., Montero, M.~L., Tsvetkov, C.,
  Biscione, V., Puebla, G., Adolfi, F., Hummel, J.~E., Heaton, R.~F., Evans,
  B.~D., Mitchell, J., and Blything, R. (2023).
\newblock Deep problems with neural network models of human vision.
\newblock {\em Behavioral and Brain Sciences}, 46:e385.
\newblock \doi{10.1017/S0140525X22002813}.

\bibitem[Chen et~al., 2022]{Chen_etal22}
Chen, J., Raghuram, J., Choi, J., Wu, X., Liang, Y., and Jha, S. (2022).
\newblock Revisiting adversarial robustness of classifiers with a reject
  option.
\newblock In {\em Proceedings of the AAAI Conference on Artificial
  Intelligence}, Workshop on Adversarial Machine Learning and Beyond.
\newblock \url{https://openreview.net/forum?id=UiF3RTES7pU}.

\bibitem[Chen et~al., 2023]{Chen_etal23}
Chen, Y., Lin, Y., Xu, R., and Vela, P.~A. (2023).
\newblock {WDiscOOD}: Out-of-distribution detection via whitened linear
  discriminant analysis.
\newblock In {\em Proceedings of the International Conference on Computer
  Vision}.
\newblock \arXiv{2303.07543}.

\bibitem[Cheng et~al., 2023]{Cheng_etal23}
Cheng, Z., Zhu, F., Zhang, X.-Y., and Liu, C.-L. (2023).
\newblock Average of pruning: Improving performance and stability of
  out-of-distribution detection.
\newblock \arXiv{2303.01201}.

\bibitem[Cimpoi et~al., 2014]{Cimpoi_etal14}
Cimpoi, M., Maji, S., Kokkinos, I., Mohamed, S., , and Vedaldi, A. (2014).
\newblock Describing textures in the wild.
\newblock In {\em Proceedings of the IEEE Computer Society Conference on
  Computer Vision and Pattern Recognition}.

\bibitem[Clanuwat et~al., 2018]{Clanuwat_etal18}
Clanuwat, T., Bober-Irizar, M., Kitamoto, A., Lamb, A., Yamamoto, K., and Ha,
  D. (2018).
\newblock Deep learning for classical japanese literature.
\newblock In {\em Proceedings of the Conference on Advances in Neural
  Information Processing Systems}, Workshop on Machine Learning for Creativity
  and Design.
\newblock \arXiv{1812.01718}.

\bibitem[Croce et~al., 2021]{Croce_etal21}
Croce, F., Andriushchenko, M., Sehwag, V., Debenedetti, E., Flammarion, N.,
  Chiang, M., Mittal, P., and Hein, M. (2021).
\newblock Robustbench: a standardized adversarial robustness benchmark.
\newblock In {\em Proceedings of the Conference on Advances in Neural
  Information Processing Systems}.
\newblock \url{https://openreview.net/forum?id=SSKZPJCt7B}.

\bibitem[Croce and Hein, 2020]{CroceHein20}
Croce, F. and Hein, M. (2020).
\newblock Reliable evaluation of adversarial robustness with an ensemble of
  diverse parameter-free attacks.
\newblock In {\em Proceedings of the International Conference on Machine
  Learning}, volume 119 of {\em Proceedings of Machine Learning Research},
  pages 2206--16.
\newblock \arXiv{2003.01690}.

\bibitem[Cui et~al., 2024]{Cui_etal23}
Cui, J., Tian, Z., Zhong, Z., Qi, X., Yu, B., and Zhang, H. (2024).
\newblock Decoupled kullback-leibler divergence loss.
\newblock In Globerson, A., Mackey, L., Belgrave, D., Fan, A., Paquet, U.,
  Tomczak, J., and Zhang, C., editors, {\em Proceedings of the Conference on
  Advances in Neural Information Processing Systems}, pages 74461--86. Curran
  Associates, Inc.
\newblock \url{https://openreview.net/forum?id=bnZZedw9CM}.
\newblock \arXiv{2305.13948}.

\bibitem[Debenedetti et~al., 2023]{Debenedetti_etal23}
Debenedetti, E., Sehwag, V., and Mittal, P. (2023).
\newblock A light recipe to train robust vision transformers.
\newblock In {\em First IEEE Conference on Secure and Trustworthy Machine
  Learning}.
\newblock \arXiv{2209.07399}.

\bibitem[Diffenderfer et~al., 2021]{Diffenderfer_etal21}
Diffenderfer, J., Bartoldson, B.~R., Chaganti, S., Zhang, J., and Kailkhura, B.
  (2021).
\newblock A winning hand: Compressing deep networks can improve
  out-of-distribution robustness.
\newblock \arXiv{2106.09129}.

\bibitem[Eykholt et~al., 2018]{Eykholt_etal17}
Eykholt, K., Evtimov, I., Fernandes, E., Li, B., Rahmati, A., Xiao, C.,
  Prakash, A., Kohno, T., and Song, D. (2018).
\newblock Robust physical-world attacks on deep learning models.
\newblock In {\em Proceedings of the IEEE Computer Society Conference on
  Computer Vision and Pattern Recognition}.
\newblock \arXiv{1707.08945}.

\bibitem[Feng et~al., 2022]{Feng_etal22}
Feng, Y., Jun, D., Ng, X., and Easwaran, A. (2022).
\newblock A unified perspective on adversarial and out-of-distribution
  detection in the open world.
\newblock In {\em Proceedings of the AAAI Conference on Artificial
  Intelligence}, Workshop on Engineering Dependable and Secure Machine Learning
  Systems.

\bibitem[Geirhos et~al., 2020]{Geirhos_etal20}
Geirhos, R., Jacobsen, J.-H., Michaelis, C., Zemel, R., Brendel, W., Bethge,
  M., and Wichmann, F.~A. (2020).
\newblock Shortcut learning in deep neural networks.
\newblock {\em Nature Machine Intelligence}, 2(11):665--73.
\newblock \doi{10.1038/s42256-020-00257-z}.
\newblock \arXiv{2004.07780}.

\bibitem[Geirhos et~al., 2019]{Geirhos_etal18b}
Geirhos, R., Rubisch, P., Michaelis, C., Bethge, M., Wichmann, F.~A., and
  Brendel, W. (2019).
\newblock {ImageNet}-trained {CNNs} are biased towards texture; increasing
  shape bias improves accuracy and robustness.
\newblock In {\em Proceedings of the International Conference on Learning
  Representations}.
\newblock \arXiv{1811.12231}.

\bibitem[Geirhos et~al., 2018]{Geirhos_etal18}
Geirhos, R., Temme, C. R.~M., Rauber, J., Sch{\"u}tt, H.~H., Bethge, M., and
  Wichmann, F.~A. (2018).
\newblock Generalisation in humans and deep neural networks.
\newblock In {\em Proceedings of the Conference on Advances in Neural
  Information Processing Systems}.
\newblock \arXiv{1808.08750}.

\bibitem[Gilmer et~al., 2019]{Ford_etal19}
Gilmer, J., Ford, N., Carlini, N., and Cubuk, E. (2019).
\newblock Adversarial examples are a natural consequence of test error in
  noise.
\newblock In Chaudhuri, K. and Salakhutdinov, R., editors, {\em Proceedings of
  the International Conference on Machine Learning}, volume~97, pages 2280--9.
  PMLR.
\newblock \url{https://proceedings.mlr.press/v97/gilmer19a.html}.

\bibitem[Goodfellow et~al., 2016]{Goodfellow_etal16}
Goodfellow, I., Bengio, Y., and Courville, A. (2016).
\newblock {\em Deep Learning}.
\newblock MIT Press.
\newblock \url{http://www.deeplearningbook.org}.

\bibitem[Goodfellow et~al., 2015]{Goodfellow_etal15}
Goodfellow, I.~J., Shlens, J., and Szegedy, C. (2015).
\newblock Explaining and harnessing adversarial examples.
\newblock In {\em Proceedings of the International Conference on Learning
  Representations}.
\newblock \arXiv{1412.6572}.

\bibitem[Gu{\'e}rin et~al., 2022]{Guerin_etal22}
Gu{\'e}rin, J., Delmas, K., Ferreira, R.~S., and Guiochet, J. (2022).
\newblock Out-of-distribution detection is not all you need.
\newblock In {\em Proceedings of the AAAI Conference on Artificial
  Intelligence}.
\newblock \arXiv{2211.16158}.

\bibitem[Gu{\'e}rin et~al., 2023]{Guerin_etal23}
Gu{\'e}rin, J., Delmas, K., Ferreira, R.~S., and Guiochet, J. (2023).
\newblock Out-of-distribution detection is not all you need.
\newblock In {\em Proceedings of the AAAI Conference on Artificial
  Intelligence}.
\newblock \arXiv{2211.16158}.

\bibitem[Guille-Escuret et~al., 2024]{Guille-Escuret_etal23}
Guille-Escuret, C., No{\"e}l, P.-A., Mitliagkas, I., Vazquez, D., and Monteiro,
  J. (2024).
\newblock Expecting the unexpected: Towards broad out-of-distribution
  detection.
\newblock In {\em Proceedings of the Conference on Advances in Neural
  Information Processing Systems}.
\newblock \arXiv{2308.11480}.

\bibitem[Guo et~al., 2023]{Guo_etal23}
Guo, J., Bao, W., Wang, J., Ma, Y., Gao, X., Xiao, G., Liu, A., Dong, J., Liu,
  X., and Wu, W. (2023).
\newblock A comprehensive evaluation framework for deep model robustness.
\newblock {\em Pattern Recognition}, 137:109308.
\newblock \doi{10.1016/j.patcog.2023.109308}.

\bibitem[Gwon and Yoo, 2023]{KyungpilJoonhyuk23}
Gwon, K. and Yoo, J. (2023).
\newblock Out-of-distribution ({OOD}) detection and generalization improved by
  augmenting adversarial mixup samples.
\newblock {\em Electronics}, 12(6).
\newblock \doi{10.3390/electronics12061421}.

\bibitem[He et~al., 2016]{He_etal16}
He, K., Zhang, X., Ren, S., and Sun, J. (2016).
\newblock Deep residual learning for image recognition.
\newblock In {\em Proceedings of the IEEE Computer Society Conference on
  Computer Vision and Pattern Recognition}, pages 770--8.
\newblock \arXiv{1512.03385}.

\bibitem[Heaven, 2019]{Heaven19}
Heaven, D. (2019).
\newblock Why deep-learning {AI}s are so easy to fool.
\newblock {\em Nature}, 574:163--6.

\bibitem[Hendrycks et~al., 2021a]{Hendrycks_etal21}
Hendrycks, D., Basart, S., Mu, N., Kadavath, S., Wang, F., Dorundo, E., Desai,
  R., Zhu, T., Parajuli, S., Guo, M., Song, D., Steinhardt, J., and Gilmer, J.
  (2021a).
\newblock The many faces of robustness: {A} critical analysis of
  out-of-distribution generalization.
\newblock In {\em Proceedings of the International Conference on Computer
  Vision}, pages 8320--9.
\newblock \doi{10.1109/ICCV48922.2021.00823}.
\newblock \arXiv{2006.16241}.

\bibitem[Hendrycks and Dietterich, 2019]{HendrycksDietterich19}
Hendrycks, D. and Dietterich, T.~G. (2019).
\newblock Benchmarking neural network robustness to common corruptions and
  perturbations.
\newblock In {\em Proceedings of the International Conference on Learning
  Representations}.
\newblock \arXiv{1903.12261}.

\bibitem[Hendrycks and Gimpel, 2017]{HendrycksGimpel17}
Hendrycks, D. and Gimpel, K. (2017).
\newblock A baseline for detecting misclassified and out-of-distribution
  examples in neural networks.
\newblock In {\em Proceedings of the International Conference on Learning
  Representations}.
\newblock \url{https://openreview.net/forum?id=Hkg4TI9xl}.
\newblock \arXiv{1610.02136}.

\bibitem[Hendrycks et~al., 2019a]{Hendrycks_etal19}
Hendrycks, D., Mazeika, M., and Dietterich, T. (2019a).
\newblock Deep anomaly detection with outlier exposure.
\newblock In {\em Proceedings of the International Conference on Learning
  Representations}.
\newblock \arXiv{1812.04606}.

\bibitem[Hendrycks et~al., 2019b]{Hendrycks_etal19b}
Hendrycks, D., Mazeika, M., Kadavath, S., and Song, D. (2019b).
\newblock Using self-supervised learning can improve model robustness and
  uncertainty.
\newblock In {\em Proceedings of the Conference on Advances in Neural
  Information Processing Systems}.
\newblock \arXiv{1906.12340}.

\bibitem[Hendrycks et~al., 2020]{Hendrycks_etal20}
Hendrycks, D., Mu, N., Cubuk, E.~D., Zoph, B., Gilmer, J., and
  Lakshminarayanan, B. (2020).
\newblock Augmix: {A} simple data processing method to improve robustness and
  uncertainty.
\newblock In {\em Proceedings of the International Conference on Learning
  Representations}.
\newblock \arXiv{1912.02781}.

\bibitem[Hendrycks et~al., 2021b]{Hendrycks_etal21b}
Hendrycks, D., Zhao, K., Basart, S., Steinhardt, J., and Song, D. (2021b).
\newblock Natural adversarial examples.
\newblock In {\em Proceedings of the IEEE Computer Society Conference on
  Computer Vision and Pattern Recognition}.
\newblock \arXiv{1907.07174}.

\bibitem[Hendrycks et~al., 2021c]{Hendrycks_etal19c}
Hendrycks, D., Zhao, K., Basart, S., Steinhardt, J., and Song, D. (2021c).
\newblock Natural adversarial examples.
\newblock In {\em Proceedings of the IEEE Computer Society Conference on
  Computer Vision and Pattern Recognition}.
\newblock \arXiv{1907.07174}.

\bibitem[Hendrycks et~al., 2022]{Hendrycks_etal22}
Hendrycks, D., Zou, A., Mazeika, M., Tang, L., Li, B., Song, D., and
  Steinhardt, J. (2022).
\newblock Pixmix: Dreamlike pictures comprehensively improve safety measures.
\newblock In {\em Proceedings of the IEEE Computer Society Conference on
  Computer Vision and Pattern Recognition}.
\newblock \arXiv{2112.05135}.

\bibitem[Holmstrom and Koistinen, 1992]{HolmstromKoistinen92}
Holmstrom, L. and Koistinen, P. (1992).
\newblock Using additive noise in back-propagation training.
\newblock {\em IEEE Transactions on Neural Networks}, 3(1):24--38.
\newblock \doi{10.1109/72.105415}.

\bibitem[Ilyas et~al., 2019]{Ilyas_etal19}
Ilyas, A., Santurkar, S., Tsipras, D., Engstrom, L., Tran, B., and Madry, A.
  (2019).
\newblock Adversarial examples are not bugs, they are features.
\newblock In Wallach, H., Larochelle, H., Beygelzimer, A., d\textquotesingle
  Alch\'{e}-Buc, F., Fox, E., and Garnett, R., editors, {\em Proceedings of the
  Conference on Advances in Neural Information Processing Systems}, volume~32.
  Curran Associates, Inc.
\newblock \arXiv{1905.02175}.

\bibitem[Jansson and Lindeberg, 2020]{JanssonLindeberg20}
Jansson, Y. and Lindeberg, T. (2020).
\newblock Exploring the ability of {CNNs} to generalise to previously unseen
  scales over wide scale ranges.
\newblock In {\em Proceedings of the International Conference on Pattern
  Recognition}, pages 1181--8.
\newblock \arXiv{2004.01536}.

\bibitem[Kannan et~al., 2018]{Kannan_etal18}
Kannan, H., Kurakin, A., and Goodfellow, I. (2018).
\newblock Adversarial logit pairing.
\newblock \arXiv{1803.06373}.

\bibitem[Khazaie et~al., 2022]{Khazaie_etal22}
Khazaie, V.~R., Wong, A., and Sabokrou, M. (2022).
\newblock Are out-of-distribution detection methods reliable?
\newblock \arXiv{2211.10892}.

\bibitem[Kim, 2021]{Kim21torchattacks}
Kim, H. (2021).
\newblock Torchattacks: a pytorch repository for adversarial attacks.
\newblock \arXiv{2010.01950}.

\bibitem[Kirchheim et~al., 2022]{Kirchheim_etal22}
Kirchheim, K., Filax, M., and Ortmeier, F. (2022).
\newblock Pytorch-{OOD}: {A} library for out-of-distribution detection based on
  pytorch.
\newblock In {\em Proceedings of the IEEE Computer Society Conference on
  Computer Vision and Pattern Recognition}, Workshops, pages 4351--60.
\newblock \doi{10.1109/CVPRW56347.2022.00481}.

\bibitem[Kireev et~al., 2022]{Kireev_etal22}
Kireev, K., Andriushchenko, M., and Flammarion, N. (2022).
\newblock On the effectiveness of adversarial training against common
  corruptions.
\newblock \arXiv{2103.02325}.

\bibitem[Krizhevsky, 2009]{Krizhevsky09}
Krizhevsky, A. (2009).
\newblock Learning multiple layers of features from tiny images.
\newblock Technical report, University of Toronto.

\bibitem[Kumano et~al., 2022]{Kumano_etal22}
Kumano, S., Kera, H., and Yamasaki, T. (2022).
\newblock Are {DNNs} fooled by extremely unrecognizable images?
\newblock \arXiv{2012.03843}.

\bibitem[Kurakin et~al., 2017]{Kurakin_etal17}
Kurakin, A., Goodfellow, I., and Bengio, S. (2017).
\newblock Adversarial examples in the physical world.
\newblock In {\em Proceedings of the International Conference on Learning
  Representations}.
\newblock \arXiv{1607.02533}.

\bibitem[Lake et~al., 2015]{Lake_etal15}
Lake, B.~M., Salakhutdinov, R., and Tenenbaum, J.~B. (2015).
\newblock Human-level concept learning through probabilistic program induction.
\newblock {\em Science}, 350(6266):1332--8.
\newblock \doi{10.1126/science.aab3050}.

\bibitem[Lapuschkin et~al., 2019]{Lapuschkin_etal19}
Lapuschkin, S., W{\"a}ldchen, S., Binder, A., Montavon, G., Samek, W., and
  M{\"u}ller, K.-R. (2019).
\newblock Unmasking clever hans predictors and assessing what machines really
  learn.
\newblock {\em Nature Communications}, 10(1096).
\newblock \doi{10.1038/s41467-019-08987-4}.

\bibitem[LeCun et~al., 1998]{LeCun_etal98}
LeCun, Y., Bottou, L., Bengio, Y., and Haffner, P. (1998).
\newblock Gradient-based learning applied to document recognition.
\newblock {\em Proceedings of the IEEE}, 86(11):2278--324.
\newblock \doi{10.1109/5.726791}.

\bibitem[Lee et~al., 2022]{Lee_etal22}
Lee, J., Prabhushankar, M., and AlRegib, G. (2022).
\newblock Gradient-based adversarial and out-of-distribution detection.
\newblock In {\em Proceedings of the International Conference on Machine
  Learning}, Workshop on New Frontiers in Adversarial Machine Learning.
\newblock \arXiv{2206.08255}.

\bibitem[Li et~al., 2017]{Li_etal17}
Li, D., Yang, Y., Song, Y.-Z., and Hospedales, T.~M. (2017).
\newblock Deeper, broader and artier domain generalization.
\newblock In {\em Proceedings of the International Conference on Computer
  Vision}, pages 5543--51.

\bibitem[Li et~al., 2024]{Li_etal24}
Li, L., Wang, Y., Sitawarin, C., and Spratling, M.~W. (2024).
\newblock {OODRobustBench}: a benchmark and large-scale analysis of adversarial
  robustness under distribution shift.
\newblock In {\em Proceedings of the International Conference on Machine
  Learning}.
\newblock \arXiv{2310.12793}.

\bibitem[Liang et~al., 2022]{Liang_etal22}
Liang, H., He, E., Zhao, Y., Jia, Z., and Li, H. (2022).
\newblock Adversarial attack and defense: {A} survey.
\newblock {\em Electronics}, 11(8).
\newblock \doi{10.3390/electronics11081283}.

\bibitem[Liu et~al., 2025]{Liu_etal25}
Liu, C., Dong, Y., Xiang, W., Yang, X., Su, H., Zhu, J., Chen, Y., He, Y., Xue,
  H., and Zheng, S. (2025).
\newblock A comprehensive study on robustness of image classification models:
  Benchmarking and rethinking.
\newblock {\em International Journal of Computer Vision}, 133:567--89.
\newblock \arXiv{2302.14301}.

\bibitem[Liu et~al., 2020]{Liu_etal20}
Liu, W., Wang, X., Owens, J., and Li, Y. (2020).
\newblock Energy-based out-of-distribution detection.
\newblock In Larochelle, H., Ranzato, M., Hadsell, R., Balcan, M.~F., and Lin,
  H., editors, {\em Proceedings of the Conference on Advances in Neural
  Information Processing Systems}, volume~33, pages 21464--75. Curran
  Associates, Inc.
\newblock
  \url{https://proceedings.neurips.cc/paper_files/paper/2020/file/f5496252609c43eb8a3d147ab9b9c006-Paper.pdf}.

\bibitem[Liu et~al., 2023]{Liu_etal23}
Liu, X., Lochman, Y., and Zach, C. (2023).
\newblock {GEN}: Pushing the limits of softmax-based out-of-distribution
  detection.
\newblock In {\em Proceedings of the IEEE Computer Society Conference on
  Computer Vision and Pattern Recognition}, pages 23946--55.
\newblock \doi{10.1109/CVPR52729.2023.02293}.

\bibitem[Lopes et~al., 2019]{Lopes_etal19}
Lopes, R.~G., Yin, D., Poole, B., Gilmer, J., and Cubuk, E.~D. (2019).
\newblock Improving robustness without sacrificing accuracy with patch gaussian
  augmentation.
\newblock \arXiv{1906.02611}.

\bibitem[Madry et~al., 2018]{Madry_etal18}
Madry, A., Makelov, A., Schmidt, L., Tsipras, D., and Vladu, A. (2018).
\newblock Towards deep learning models resistant to adversarial attacks.
\newblock In {\em Proceedings of the International Conference on Learning
  Representations}.
\newblock \arXiv{1706.06083}.

\bibitem[Malhotra et~al., 2020]{Malhotra_etal20}
Malhotra, G., Evans, B.~D., and Bowers, J.~S. (2020).
\newblock Hiding a plane with a pixel: examining shape-bias in {CNNs} and the
  benefit of building in biological constraints.
\newblock {\em Vision Research}, 174:57--68.
\newblock \doi{10.1016/j.visres.2020.04.013}.

\bibitem[Marcus, 2020]{Marcus20}
Marcus, G. (2020).
\newblock The next decade in {AI}: Four steps towards robust artificial
  intelligence.
\newblock \arXiv{2002.06177}.

\bibitem[Michaelis et~al., 2019]{Michaelis_etal19}
Michaelis, C., Mitzkus, B., Geirhos, R., Rusak, E., Bringmann, O., Ecker,
  A.~S., Bethge, M., and Brendel, W. (2019).
\newblock Benchmarking robustness in object detection: Autonomous driving when
  winter is coming.
\newblock \arXiv{1907.07484}.

\bibitem[Modas et~al., 2022]{Modas_etal22}
Modas, A., Rade, R., {Ortiz-Jim\'enez}, G., {Moosavi-Dezfooli}, S.-M., and
  Frossard, P. (2022).
\newblock {PRIME}: a few primitives can boost robustness to common corruptions.
\newblock In {\em Proceedings of the European Conference on Computer Vision}.
\newblock \arXiv{2112.13547}.

\bibitem[Mohseni et~al., 2020]{Mohseni_etal20}
Mohseni, S., Pitale, M., Yadawa, J., and Wang, Z. (2020).
\newblock Self-supervised learning for generalizable out-of-distribution
  detection.
\newblock In {\em Proceedings of the AAAI Conference on Artificial
  Intelligence}, volume~34, pages 5216--23.
\newblock \doi{10.1609/aaai.v34i04.5966}.

\bibitem[Mu and Gilmer, 2019]{MuGilmer19}
Mu, N. and Gilmer, J. (2019).
\newblock {MNIST-C:} {A} robustness benchmark for computer vision.
\newblock \arXiv{1906.02337}.

\bibitem[Netzer et~al., 2011]{Netzer_etal11}
Netzer, Y., Wang, T., Coates, A., Bissacco, A., Wu, B., and Ng, A.~Y. (2011).
\newblock Reading digits in natural images with unsupervised feature learning.
\newblock In {\em Proceedings of the Conference on Advances in Neural
  Information Processing Systems}, Workshop on Deep Learning and Unsupervised
  Feature Learning.

\bibitem[Nguyen et~al., 2015]{Nguyen_etal15}
Nguyen, A., Yosinski, J., and Clune, J. (2015).
\newblock Deep neural networks are easily fooled: High confidence predictions
  for unrecognizable images.
\newblock In {\em Proceedings of the IEEE Computer Society Conference on
  Computer Vision and Pattern Recognition}.
\newblock \arXiv{1412.1897}.

\bibitem[Pang et~al., 2021]{Pang_etal21}
Pang, T., Yang, X., Dong, Y., Su, H., and Zhu, J. (2021).
\newblock Bag of tricks for adversarial training.
\newblock In {\em Proceedings of the International Conference on Learning
  Representations}.
\newblock \arXiv{2010.00467}.

\bibitem[Pang et~al., 2022]{Pang_etal22}
Pang, T., Zhang, H., He, D., Dong, Y., Su, H., Chen, W., Zhu, J., and Liu,
  T.-Y. (2022).
\newblock Two coupled rejection metrics can tell adversarial examples apart.
\newblock In {\em Proceedings of the IEEE Computer Society Conference on
  Computer Vision and Pattern Recognition}.
\newblock \arXiv{2105.14785}.

\bibitem[Papernot et~al., 2016]{Papernot_etal16}
Papernot, N., McDaniel, P., Jha, S., Fredrikson, M., Celik, Z.~B., and Swami,
  A. (2016).
\newblock The limitations of deep learning in adversarial settings.
\newblock In {\em {IEEE} European Symposium on Security and Privacy}.
\newblock \arXiv{1511.07528}.

\bibitem[Paszke et~al., 2019]{Pytorch19}
Paszke, A., Gross, S., Massa, F., Lerer, A., Bradbury, J., Chanan, G., Killeen,
  T., Lin, Z., Gimelshein, N., Antiga, L., Desmaison, A., Kopf, A., Yang, E.,
  DeVito, Z., Raison, M., Tejani, A., Chilamkurthy, S., Steiner, B., Fang, L.,
  Bai, J., and Chintala, S. (2019).
\newblock Pytorch: An imperative style, high-performance deep learning library.
\newblock In {\em Proceedings of the Conference on Advances in Neural
  Information Processing Systems}, volume~32.
\newblock \arXiv{1912.01703}.

\bibitem[Pinto et~al., 2023]{Pinto_etal23}
Pinto, F., Yang, H., Lim, S.-N., Torr, P. H.~S., and Dokania, P.~K. (2023).
\newblock Regmixup: Mixup as a regularizer can surprisingly improve accuracy
  and out distribution robustness.
\newblock \arXiv{2206.14502}.

\bibitem[Rauter et~al., 2023]{Rauter_etal23}
Rauter, R., Nocker, M., Merkle, F., and Sch{\"o}ttle, P. (2023).
\newblock On the effect of adversarial training against invariance-based
  adversarial examples.
\newblock In {\em Proceedings of the 2023 8th International Conference on
  Machine Learning Technologies}, page 54–60, New York, NY, USA. Association
  for Computing Machinery.
\newblock \doi{10.1145/3589883.3589891}.

\bibitem[Rebuffi et~al., 2021]{Rebuffi_etal21}
Rebuffi, S.-A., Gowal, S., Calian, D.~A., Stimberg, F., Wiles, O., and Mann, T.
  (2021).
\newblock Data augmentation can improve robustness.
\newblock In {\em Proceedings of the Conference on Advances in Neural
  Information Processing Systems}.
\newblock \url{https://openreview.net/pdf?id=kgVJBBThdSZ}.
\newblock \arXiv{2111.05328}.

\bibitem[Recht et~al., 2018]{Recht_etal18}
Recht, B., Roelofs, R., Schmidt, L., and Shankar, V. (2018).
\newblock Do {CIFAR}-10 classifiers generalize to {CIFAR}-10?
\newblock \arXiv{1806.00451}.

\bibitem[Recht et~al., 2019]{Recht_etal19}
Recht, B., Roelofs, R., Schmidt, L., and Shankar, V. (2019).
\newblock Do {ImageNet} classifiers generalize to {ImageNet}?
\newblock In Chaudhuri, K. and Salakhutdinov, R., editors, {\em Proceedings of
  the International Conference on Machine Learning}, volume~97 of {\em
  Proceedings of Machine Learning Research}, pages 5389--400.
\newblock \url{https://proceedings.mlr.press/v97/recht19a.html}.
\newblock \arXiv{1902.10811}.

\bibitem[Ren et~al., 2020]{Ren_etal20}
Ren, K., Zheng, T., Qin, Z., and Liu, X. (2020).
\newblock Adversarial attacks and defenses in deep learning.
\newblock {\em Engineering}, 6(3):346--60.
\newblock \doi{10.1016/j.eng.2019.12.012}.

\bibitem[Ren et~al., 2022]{Ren_etal22}
Ren, M., Wang, Y.~L., and He, Z.~F. (2022).
\newblock Towards interpretable defense against adversarial attacks via causal
  inference.
\newblock {\em Machine Intelligence Research}, 19(3):209--26.
\newblock \doi{10.1007/s11633-022-1330-7}.

\bibitem[Riaz and Smeaton, 2023]{RiazSmeaton23}
Riaz, H. and Smeaton, A.~F. (2023).
\newblock Vision based machine learning algorithms for out-of-distribution
  generalisation.
\newblock In {\em Computing Conference}.
\newblock \arXiv{2301.06975}.

\bibitem[Rusak et~al., 2020]{Rusak_etal20}
Rusak, E., Schott, L., Zimmermann, R.~S., Bitterwolf, J., Bringmann, O.,
  Bethge, M., and Brendel, W. (2020).
\newblock A simple way to make neural networks robust against diverse image
  corruptions.
\newblock In {\em Proceedings of the European Conference on Computer Vision}.
\newblock \arXiv{2001.06057}.

\bibitem[Russakovsky et~al., 2015]{Russakovsky_etal15}
Russakovsky, O., Deng, J., Su, H., Krause, J., Satheesh, S., Ma, S., Huang, Z.,
  Karpathy, A., Khosla, A., Bernstein, M., Berg, A.~C., and Fei-Fei, L. (2015).
\newblock {ImageNet Large Scale Visual Recognition Challenge}.
\newblock {\em International Journal of Computer Vision}, 115(3):211--52.
\newblock \doi{10.1007/s11263-015-0816-y}.

\bibitem[Sa-Couto and Wichert, 2021]{Sa-CoutoWichert21}
Sa-Couto, L. and Wichert, A. (2021).
\newblock {Simple Convolutional-Based Models: Are They Learning the Task or the
  Data?}
\newblock {\em Neural Computation}, 33(12):3334--50.
\newblock \doi{10.1162/neco\_a\_01446}.

\bibitem[Serre, 2019]{Serre19}
Serre, T. (2019).
\newblock Deep learning: The good, the bad, and the ugly.
\newblock {\em Annual Review of Vision Science}, 5(1):399--426.
\newblock \doi{10.1146/annurev-vision-091718-014951}.

\bibitem[Shafaei et~al., 2019]{Shafaei_etal18}
Shafaei, A., Schmidt, M., and Little, J.~J. (2019).
\newblock A less biased evaluation of out-of-distribution sample detectors.
\newblock In {\em Proceedings of the British Machine Vision Conference}.
\newblock \arXiv{1809.04729}.

\bibitem[Shao et~al., 2020]{Shao_etal20}
Shao, R., Perera, P., Yuen, P.~C., and Patel, V.~M. (2020).
\newblock Open-set adversarial defense.
\newblock In {\em Proceedings of the European Conference on Computer Vision}.
\newblock \arXiv{2009.00814}.

\bibitem[Sietsma and Dow, 1991]{SietsmaDow91}
Sietsma, J. and Dow, R. J.~F. (1991).
\newblock Creating artificial neural networks that generalize.
\newblock {\em Neural Networks}, 4:67--79.

\bibitem[Singh et~al., 2023]{Singh_etal23}
Singh, N.~D., Croce, F., and Hein, M. (2023).
\newblock Revisiting adversarial training for imagenet: Architectures, training
  and generalization across threat models.
\newblock In Oh, A., Naumann, T., Globerson, A., Saenko, K., Hardt, M., and
  Levine, S., editors, {\em Proceedings of the Conference on Advances in Neural
  Information Processing Systems}, volume~36, pages 13931--55. Curran
  Associates, Inc.
\newblock \arXiv{2303.01870}.

\bibitem[Song et~al., 2020]{Song_etal20}
Song, L., Sehwag, V., Bhagoji, A.~N., and Mittal, P. (2020).
\newblock A critical evaluation of open-world machine learning.
\newblock In {\em Proceedings of the International Conference on Machine
  Learning}, Workshop on Uncertainty and Robustness in Deep Learning.
\newblock \arXiv{2007.04391}.

\bibitem[Strubell et~al., 2020]{Strubell_etal20}
Strubell, E., Ganesh, A., and McCallum, A. (2020).
\newblock Energy and policy considerations for modern deep learning research.
\newblock In {\em Proceedings of the AAAI Conference on Artificial
  Intelligence}, volume~34, pages 13693--6.
\newblock \doi{10.1609/aaai.v34i09.7123}.

\bibitem[Stutz et~al., 2020]{Stutz_etal20}
Stutz, D., Hein, M., and Schiele, B. (2020).
\newblock Confidence-calibrated adversarial training: Generalizing to unseen
  attacks.
\newblock In {\em Proceedings of the International Conference on Machine
  Learning}, volume 119 of {\em Proceedings of Machine Learning Research},
  pages 9155--66.
\newblock \arXiv{1910.06259}.

\bibitem[Sun et~al., 2022]{Sun_etal21}
Sun, J., Mehra, A., Kailkhura, B., Chen, P.-Y., Hendrycks, D., Hamm, J., and
  Mao, Z.~M. (2022).
\newblock Certified adversarial defenses meet out-of-distribution corruptions:
  Benchmarking robustness and simple baselines.
\newblock In {\em Proceedings of the European Conference on Computer Vision}.
\newblock \arXiv{2112.00659}.

\bibitem[Szegedy et~al., 2014]{Szegedy_etal14}
Szegedy, C., Zaremba, W., Sutskever, I., Bruna, J., Erhan, D., Goodfellow,
  I.~J., and Fergus, R. (2014).
\newblock Intriguing properties of neural networks.
\newblock In {\em Proceedings of the International Conference on Learning
  Representations}.
\newblock \arXiv{1312.6199}.

\bibitem[Szyc et~al., 2023]{Szyc_etal23}
Szyc, K., Walkowiak, T., and Maciejewski, H. (2023).
\newblock Why out-of-distribution detection experiments are not reliable -
  subtle experimental details muddle the {OOD} detector rankings.
\newblock In Evans, R.~J. and Shpitser, I., editors, {\em Proceedings of the
  Thirty-Ninth Conference on Uncertainty in Artificial Intelligence}, volume
  216 of {\em Proceedings of Machine Learning Research}, pages 2078--88.
\newblock \url{https://proceedings.mlr.press/v216/szyc23a.html}.

\bibitem[Tajwar et~al., 2021]{Tajwar_etal21}
Tajwar, F., Kumar, A., Xie, S.~M., and Liang, P. (2021).
\newblock No true state-of-the-art? {OOD} detection methods are inconsistent
  across datasets.
\newblock In {\em Proceedings of the International Conference on Machine
  Learning}, Workshop on Uncertainty and Robustness in Deep Learning.
\newblock \arXiv{2109.05554}.

\bibitem[Thompson et~al., 2020]{Thompson_etal20}
Thompson, N.~C., Greenewald, K., Lee, K., and Manso, G.~F. (2020).
\newblock The computational limits of deep learning.
\newblock \arXiv{2007.05558}.

\bibitem[Tian et~al., 2022]{Tian_etal22}
Tian, R., Wu, Z., Dai, Q., Hu, H., and Jiang, Y.-G. (2022).
\newblock Deeper insights into the robustness of vits towards common
  corruptions.
\newblock \arXiv{2204.12143}.

\bibitem[Tram{\`e}r, 2021]{Tramer21}
Tram{\`e}r, F. (2021).
\newblock Detecting adversarial examples is (nearly) as hard as classifying
  them.
\newblock In {\em Proceedings of the International Conference on Machine
  Learning}, volume 162 of {\em Proceedings of Machine Learning Research},
  pages 21692--702.
\newblock \arXiv{2107.11630}.

\bibitem[Tram{\`e}r et~al., 2020]{Tramer_etal20}
Tram{\`e}r, F., Behrmann, J., Carlini, N., Papernot, N., and Jacobsen, J.-H.
  (2020).
\newblock Fundamental tradeoffs between invariance and sensitivity to
  adversarial perturbations.
\newblock In {\em Proceedings of the International Conference on Machine
  Learning}.
\newblock \arXiv{2002.04599}.

\bibitem[Tsipras et~al., 2019]{Tsipras_etal19}
Tsipras, D., Santurkar, S., Engstrom, L., Turner, A., and Madry, A. (2019).
\newblock Robustness may be at odds with accuracy.
\newblock In {\em Proceedings of the International Conference on Learning
  Representations}.
\newblock \arXiv{1805.12152}.

\bibitem[{Van Horn} et~al., 2018]{VanHorn_etal18}
{Van Horn}, G., Aodha, O.~M., Song, Y., Cui, Y., Sun, C., Shepard, A., Adam,
  H., Perona, P., and Belongie, S. (2018).
\newblock The inaturalist species classification and detection dataset.
\newblock In {\em Proceedings of the IEEE Computer Society Conference on
  Computer Vision and Pattern Recognition}.
\newblock \arXiv{1707.06642}.

\bibitem[Vasconcelos et~al., 2021]{Vasconcelos_etal21}
Vasconcelos, C., Larochelle, H., Dumoulin, V., Romijnders, R., Roux, N.~L., and
  Goroshin, R. (2021).
\newblock Impact of aliasing on generalization in deep convolutional networks.
\newblock In {\em Proceedings of the International Conference on Computer
  Vision}, pages 10509--18.
\newblock \doi{10.1109/ICCV48922.2021.01036}.

\bibitem[Vaze et~al., 2022]{Vaze_etal22}
Vaze, S., Han, K., Vedaldi, A., and Zisserman, A. (2022).
\newblock Open-set recognition: a good closed-set classifier is all you need?
\newblock In {\em Proceedings of the International Conference on Learning
  Representations}.
\newblock \url{https://openreview.net/forum?id=5hLP5JY9S2d}.
\newblock \arXiv{2110.06207}.

\bibitem[Vojir et~al., 2023]{Vojir_etal23}
Vojir, T., Sochman, J., Aljundi, R., and Matas, J. (2023).
\newblock Calibrated out-of-distribution detection with a generic
  representation.
\newblock In {\em Proceedings of the International Conference on Computer
  Vision}, Workshop on Uncertainty Quantification for Computer Vision.
\newblock \arXiv{2303.13148}.

\bibitem[Wang et~al., 2022]{Wang_etal22}
Wang, Y., Guo, J., Guo, S., Zhang, W., and Zhang, J. (2022).
\newblock Exploring optimal substructure for out-of-distribution generalization
  via feature-targeted model pruning.
\newblock \arXiv{2212.09458}.

\bibitem[Wang et~al., 2023]{Wang_etal23}
Wang, Z., Pang, T., Du, C., Lin, M., Liu, W., and Yan, S. (2023).
\newblock Better diffusion models further improve adversarial training.
\newblock In {\em Proceedings of the International Conference on Machine
  Learning}.
\newblock \arXiv{2302.04638}.

\bibitem[Weiss and Tonella, 2022]{WeissTonella22}
Weiss, M. and Tonella, P. (2022).
\newblock Simple techniques work surprisingly well for neural network test
  prioritization and active learning.
\newblock In {\em Proceedings of the 31th ACM SIGSOFT International Symposium
  on Software Testing and Analysis}.

\bibitem[Xiao et~al., 2017]{Xiao_etal17}
Xiao, H., Rasul, K., and Vollgraf, R. (2017).
\newblock Fashion-{MNIST}: a novel image dataset for benchmarking machine
  learning algorithms.
\newblock \arXiv{1708.07747}.

\bibitem[Xu et~al., 2020]{Xu_etal20}
Xu, H., Ma, Y., Liu, H.-C., Deb, D., Liu, H., Tang, J.-L., and Jain, A.~K.
  (2020).
\newblock Adversarial attacks and defenses in images, graphs and text: {A}
  review.
\newblock {\em International Journal of Automation and Computing}, 17:151--78.
\newblock \doi{10.1007/s11633-019-1211-x}.

\bibitem[Xu-Darme et~al., 2023]{Xu-Darme_etal23}
Xu-Darme, R., Girard-Satabin, J., Hond, D., Incorvaia, G., and Chihani, Z.
  (2023).
\newblock Interpretable out-of-distribution detection using pattern
  identification.
\newblock \url{https://hal-cea.archives-ouvertes.fr/cea-03951966}.

\bibitem[Yadav and Bottou, 2019]{YadavBottou19}
Yadav, C. and Bottou, L. (2019).
\newblock Cold case: The lost {MNIST} digits.
\newblock In Wallach, H., Larochelle, H., Beygelzimer, A., d\textquotesingle
  Alch\'{e}-Buc, F., Fox, E., and Garnett, R., editors, {\em Proceedings of the
  Conference on Advances in Neural Information Processing Systems}, volume~32.
  Curran Associates, Inc.
\newblock \arXiv{1905.10498}.

\bibitem[Yang et~al., 2022]{Yang_etal22}
Yang, J., Wang, P., Zou, D., Zhou, Z., Ding, K., Peng, W., Wang, H., Chen, G.,
  Li, B., Sun, Y., Du, X., Zhou, K., Zhang, W., Hendrycks, D., Li, Y., and Liu,
  Z. (2022).
\newblock {OpenOOD}: Benchmarking generalized out-of-distribution detection.
\newblock In {\em Proceedings of the Conference on Advances in Neural
  Information Processing Systems}.
\newblock \arXiv{2210.07242}.

\bibitem[Yang et~al., 2024]{Yang_etal21}
Yang, J., Zhou, K., Li, Y., and Liu, Z. (2024).
\newblock Generalized out-of-distribution detection: {A} survey.
\newblock {\em International Journal of Computer Vision}, 132:5635--62.
\newblock \doi{10.1007/s11263-024-02117-4}.
\newblock \arXiv{2110.11334}.

\bibitem[Yang et~al., 2023a]{Yang_etal23b}
Yang, J., Zhou, K., and Liu, Z. (2023a).
\newblock Full-spectrum out-of-distribution detection.
\newblock {\em International Journal of Computer Vision}, 131:2607--622.
\newblock \doi{10.1007/s11263-023-01811-z}.

\bibitem[Yang et~al., 2023b]{Yang_etal23}
Yang, T., Huang, Y., Xie, Y., Liu, J., and Wang, S. (2023b).
\newblock {MixOOD}: Improving out-of-distribution detection with enhanced data
  mixup.
\newblock {\em {ACM} Transactions on Multimedia Computing, Communications, and
  Applications}.
\newblock \doi{10.1145/3578935}.

\bibitem[Yuille and Liu, 2021]{YuilleLiu21}
Yuille, A.~L. and Liu, C. (2021).
\newblock Deep nets: What have they ever done for vision?
\newblock {\em International Journal of Computer Vision}, 129:781--802.
\newblock \doi{10.1007/s11263-020-01405-z}.

\bibitem[Zhang et~al., 2017]{ZhangH_etal17}
Zhang, H., Cisse, M., Dauphin, Y.~N., and Lopez-Paz, D. (2017).
\newblock mixup: Beyond empirical risk minimization.
\newblock In {\em Proceedings of the International Conference on Learning
  Representations}.
\newblock \arXiv{1710.09412}.

\bibitem[Zhang et~al., 2019]{Zhang_etal19}
Zhang, H., Yu, Y., Jiao, J., Xing, E.~P., Ghaoui, L.~E., and Jordan, M.~I.
  (2019).
\newblock Theoretically principled trade-off between robustness and accuracy.
\newblock In {\em Proceedings of the International Conference on Machine
  Learning}.
\newblock \arXiv{1901.08573}.

\bibitem[Zhang et~al., 2023a]{Zhang_etal23b}
Zhang, J., Yang, J., Wang, P., Wang, H., Lin, Y., Zhang, H., Sun, Y., Du, X.,
  Zhou, K., Zhang, W., Li, Y., Liu, Z., Chen, Y., and Li, H. (2023a).
\newblock Open{OOD} v1.5: Enhanced benchmark for out-of-distribution detection.
\newblock In {\em Proceedings of the Conference on Advances in Neural
  Information Processing Systems}, Workshop on Distribution Shifts.
\newblock \arXiv{2306.09301}.

\bibitem[Zhang and Ranganath, 2023]{ZhangRanganath23}
Zhang, L.~H. and Ranganath, R. (2023).
\newblock Robustness to spurious correlations improves semantic
  out-of-distribution detection.
\newblock In {\em Proceedings of the AAAI Conference on Artificial
  Intelligence}.
\newblock \arXiv{2302.04132}.

\bibitem[Zhang et~al., 2023b]{Zhang_etal23}
Zhang, X., He, Y., Wang, T., Qi, J., Yu, H., Wang, Z., Peng, J., Xu, R., Shen,
  Z., Niu, Y., Zhang, H., and Cui, P. (2023b).
\newblock {NICO} challenge: Out-of-distribution generalization for image
  recognition challenges.
\newblock In Karlinsky, L., Michaeli, T., and Nishino, K., editors, {\em
  Proceedings of the European Conference on Computer Vision}, volume 13806 of
  {\em Workshop on Causality in Vision}, pages 433--50. Lecture Notes in
  Computer Science book series.

\bibitem[Zhao et~al., 2024]{Zhao_etal23}
Zhao, B., Wang, J., Ma, W., Jesslen, A., Yang, S., Yu, S., Zendel, O.,
  Theobalt, C., Yuille, A., and Kortylewski, A. (2024).
\newblock {OOD-CV-v2}: An extended benchmark for robustness to
  out-of-distribution shifts of individual nuisances in natural images.
\newblock {\em IEEE Transactions on Pattern Analysis and Machine Intelligence},
  46(12):11104--18.
\newblock \doi{10.1109/TPAMI.2024.3462293}.
\newblock \arXiv{2304.10266}.

\bibitem[Zhu et~al., 2024]{Zhu_etal22}
Zhu, Y., Chen, Y., Li, X., Zhang, R., Xue, H., Tian, X., Jiang, R., Zheng, B.,
  and Chen, Y. (2024).
\newblock Rethinking out-of-distribution detection from a human-centric
  perspective.
\newblock {\em International Journal of Computer Vision}, 132:4633--50.
\newblock \doi{10.1007/s11263-024-02099-3}.
\newblock \arXiv{2211.16778}.

\end{thebibliography}
\end{document}